\documentclass{article} 
\usepackage[final]{colm2026_conference}

\usepackage{microtype}
\usepackage{hyperref}
\usepackage{url}
\usepackage{booktabs}

\newif\ifshowcomments
\showcommentsfalse

\usepackage[colorinlistoftodos]{todonotes}

\usepackage{lineno}
\usepackage[table]{xcolor}
\usepackage{amssymb}
\usepackage{makecell}
\usepackage{etoc}

\definecolor{darkblue}{rgb}{0, 0, 0.5}
\hypersetup{colorlinks=true, citecolor=darkblue, linkcolor=darkblue, urlcolor=darkblue}

\usepackage{minted}
\usepackage{fvextra}
\usepackage[hybrid]{markdown}
\usepackage{upquote}
\usepackage{listings}

\lstdefinelanguage{text}{}
\usepackage{tcolorbox}
\tcbuselibrary{breakable}
\usepackage{ragged2e}

\newcolumntype{L}[1]{>{\raggedright\arraybackslash\hyphenpenalty=10000\exhyphenpenalty=10000}p{#1}}

\usepackage{tikz}
\usepackage{subcaption}  
\usetikzlibrary{arrows.meta, positioning, fit, backgrounds, calc, decorations.pathreplacing}

\usepackage{xcolor}
\usepackage{colortbl}
\usepackage{array}

\usepackage{enumitem} 

\newenvironment{tightitemize}
  {\begin{itemize}[noitemsep, topsep=0pt, leftmargin=*]}
  {\end{itemize}}

\usepackage{subfiles} 

\usepackage[capitalise]{cleveref}
\crefname{section}{\S}{\S\S}
\Crefname{section}{\S}{\S\S}
\crefname{subsection}{\S}{\S\S}
\Crefname{subsection}{\S}{\S\S}

\title{A case study of evaluating AI agents on a neuroscience\\data-to-discovery pipeline}

\author{Kai A. Horstmann$^1$, Ethan Lin$^1$, Alice A. Robie$^2$, Jennifer J. Sun$^1$, Kristin Branson$^2$\\
$^1$Cornell University\quad$^2$HHMI Janelia Research Campus\\
}

\newcommand{\mrred}{\textcolor{red!80!black}{$\times$}}
\newcommand{\mryellow}{\textcolor{yellow!80!black}{$\sim$}}
\newcommand{\mrgreen}{\textcolor{green!70!black}{\checkmark}}

\begin{document}

\ifcolmsubmission
\linenumbers
\fi

\maketitle

\begin{abstract}
Agentic AI offers a promising path to automating software development bottlenecks in scientific research pipelines, particularly for stages that take domain experts days to months to build and where correctness and robustness matter more than implementation details. We present an empirical study of general-purpose coding agents on a fly optogenetics data-to-discovery pipeline. We assess agents on tasks and datasets substantially larger than existing benchmarks and evaluation criteria grounded in domain expert standards. We show that agents can solve several pipeline stages, suggesting stage-level automation is tractable. By analyzing agents' code iterations, we show they struggle most without a pre-defined criterion, when they must instead use their scientific judgment to assess their current solution. Mirroring scientists, they sometimes attempt visual inspection of intermediate outputs for self-evaluation, but largely fail to interpret what they see or act on it appropriately. Solving the end-to-end pipeline requires stringing together successes across all stages, which is beyond agents' current abilities. We identify challenges largely absent from existing benchmarks, including computational resource management and generalization to held-out data. Finally, we distill principles for constructing scientific tasks and rigorous evaluation criteria for open-ended problems.
\end{abstract}

\begingroup
\renewcommand{\thefootnote}{}
\footnotetext{Code available at: \href{https://github.com/kaihorstmann/neuro-d2d-eval}{https://github.com/kaihorstmann/neuro-d2d-eval}}
\addtocounter{footnote}{-1}
\endgroup

\section{Introduction}

Science is inherently iterative: scientists break complex phenomena into manageable pieces, understand each with as much rigor as possible, and build on those who came before.
In behavioral neuroscience, modern experiments \citep{Schretter2025, FlyDisco2024, AllenInstitute2021VisualBehavior2P, IBL2022BrainwideMapWhitepaper} rely on tools and methodologies \citep{JAABAInteractiveMachinekabra2013, ctrax, FlyTracker, pachitariu2016suite2p, Pachitariu2023.01.07.523036, mathis2018deeplabcut} developed over decades. These tools function as trusted modules: they solve specific problems (e.g., tracking or signal processing) well enough for the community to treat them as reliable foundations, allowing researchers to move on to harder questions. These modules represent months to years of effort by scientists with deep, narrow expertise.

\begin{figure}[t]
  \centering
  \resizebox{\linewidth}{!}{%
    \input{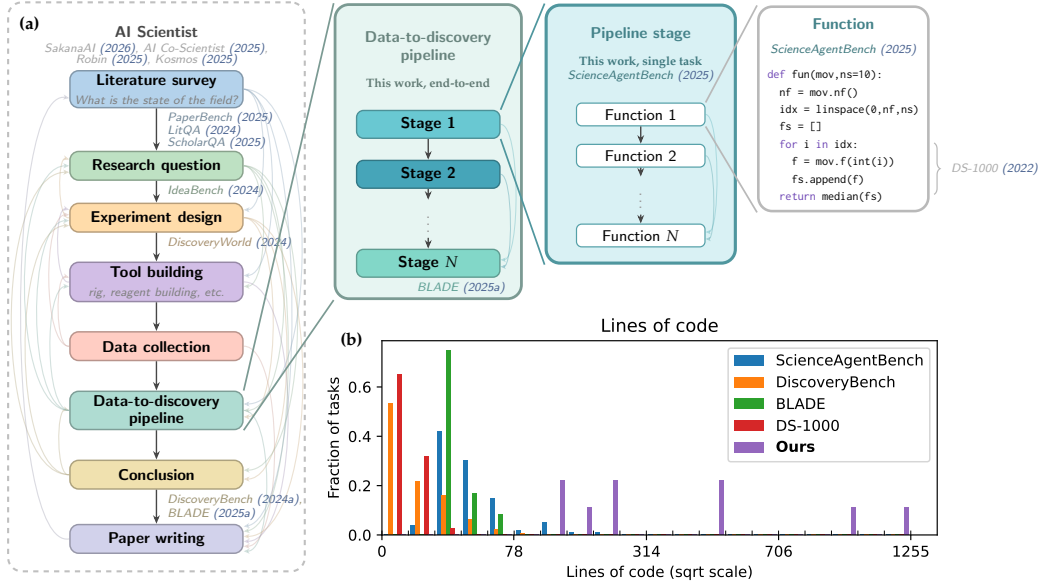}%
  }
  \caption{\small
(a) Taxonomy of AI in science benchmarks, mapping existing work to stages of the scientific workflow from literature survey to paper writing. (b) Distribution of task sizes (lines of code) across benchmarks on a square-root scale. Our pipeline stages are substantially larger, reflecting the scale of tasks that would represent significant time savings for an expert scientist if successfully automated.} 
  \label{fig:related_work_fig}
  \vspace{-0.115cm}
\end{figure}

This reductive approach---decomposing complex workflows into discrete, trusted modules---presents unique challenges and opportunities for AI in science. Because scientific validity relies on the integrity of every step in the chain, any AI-produced solution must be as valid as that produced by expert scientists. This high bar has made scientists hesitant to adopt fully automated ``AI scientist'' or self-driving lab approaches \citep{Sakana2026,AICoScientist2025,Robin2025} for which we lack not only evidence of human-level performance, but even the metrics to assess capabilities like creativity, diversity, and scientific reasoning. Instead, a more tractable near-term opportunity lies in automating the software development problems embedded in research pipelines, verifiable stages for which scientists care about correctness and robustness, but not about implementation details. 
Building these stages is increasingly feasible with rapid progress in agentic coding, but they are substantially more complex than prior AI-for-science benchmarks (\cref{fig:related_work_fig}): they can take a developer days to months to build, and must generalize to unseen, possibly out-of-distribution data.

\begin{figure}
    \centering
    \includegraphics[width=\linewidth]{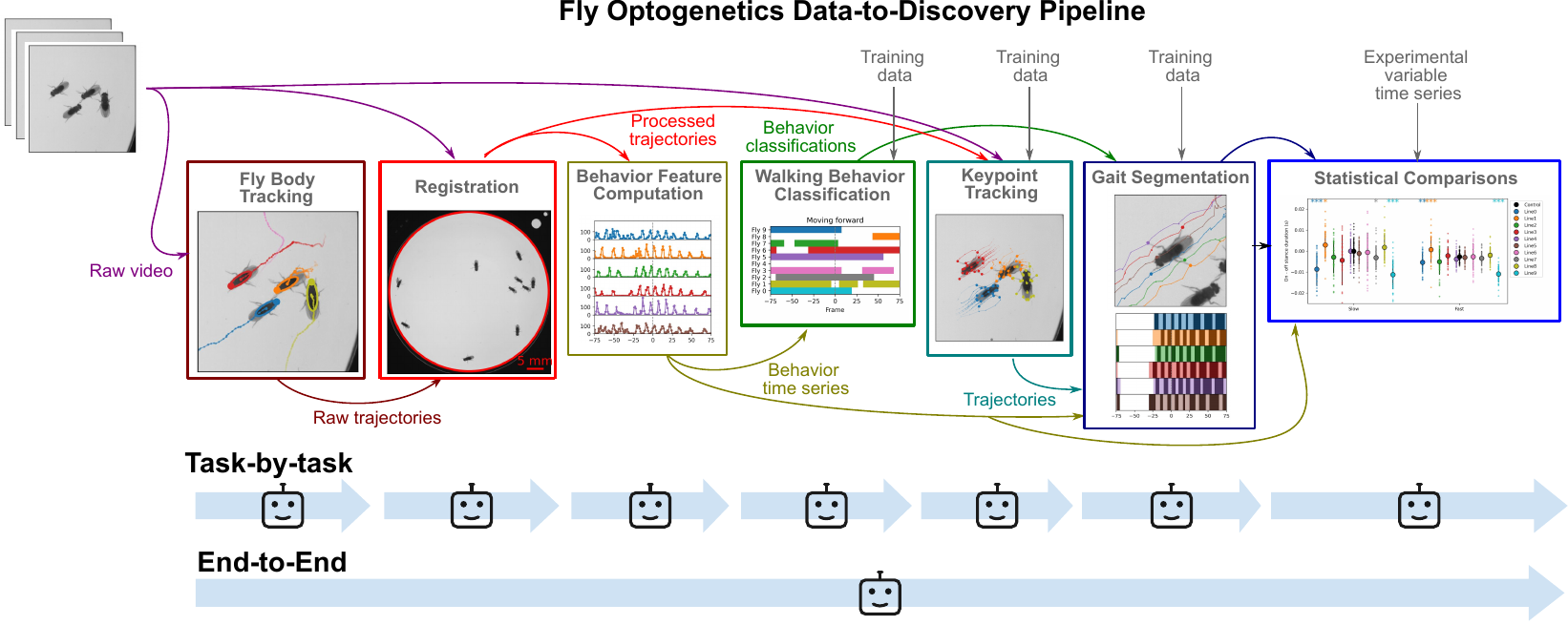}
    \caption{\small Outline of our analysis of a 7-stage fly optogenetics data-to-discovery pipeline. We evaluate and analyze agents at the task level as well as end-to-end across the entire pipeline.}
    \label{fig:pipeline_overview}
    \vspace{-0.218cm}
\end{figure}

In this work, we present an empirical study of general-purpose coding agents on a representative end-to-end neuroscience research pipeline: fly optogenetics (\cref{fig:pipeline_overview}). 
We decompose this pipeline into 7 tasks spanning the full analysis, from bespoke data handling and computer vision tracking to feature computation, behavior classification, and statistical interpretation, and evaluate agents against both expert human annotations and trusted legacy codebases that define the current scientific standard.
We focus on general-purpose models to assess immediate, off-the-shelf automation potential and to systematically measure the impact of specific design choices on performance. Beyond quantifying success rates, we manually analyze the agents' code iterations and solutions to identify major failure modes and surprising success modes, identifying key directions for future development of agentic systems for science and benchmarking. We also distill principles of task design, with particular focus on constructing quantitative evaluation criteria for open-ended problems, a challenge existing benchmarks acknowledge but rarely address concretely. 

\begin{figure}[t]
    \begin{minipage}[m]{.3\linewidth}
        \vspace{0pt}
        \centering
        \includegraphics[width=\linewidth]{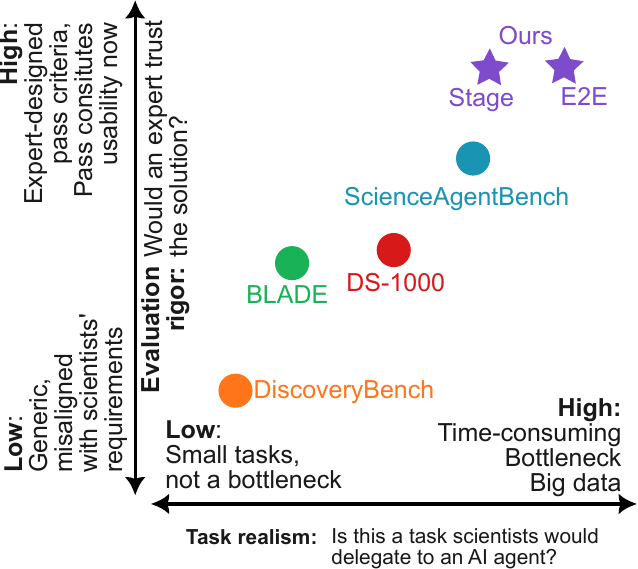}
    \end{minipage}
    \begin{minipage}[m]{.7\linewidth}
    \vspace{0pt}
    \centering
    \definecolor{greenheader}{RGB}{180,220,180}
\definecolor{greendark}{RGB}{200,228,200}
\definecolor{greenlight}{RGB}{225,242,225}
\definecolor{blueheader}{RGB}{180,210,235}
\definecolor{bluedark}{RGB}{195,215,235}
\definecolor{bluelight}{RGB}{220,232,245}

\begin{tiny}
\renewcommand{\arraystretch}{1}
\setlength{\tabcolsep}{0.5pt}
\begin{tabular}{|>{\centering\arraybackslash}p{0.2\linewidth-2\tabcolsep-2\arrayrulewidth}|
                  >{\centering\arraybackslash}p{0.27\linewidth-2\tabcolsep-2\arrayrulewidth}|
                  >{\centering\arraybackslash}p{0.13\linewidth-2\tabcolsep-2\arrayrulewidth}|
                  >{\centering\arraybackslash}p{0.2\linewidth-2\tabcolsep-2\arrayrulewidth}|
                  >{\centering\arraybackslash}p{0.2\linewidth-2\tabcolsep-2\arrayrulewidth}|}
\hline
 & \multicolumn{2}{c|}{\cellcolor{greenlight}\bf Task realism} & \multicolumn{2}{c|}{\cellcolor{bluelight}\bf Evaluation rigor} \\
\hline
\cellcolor{gray!20}\textbf{Stage} &
\cellcolor{greenheader}\textbf{Input modalities} &
\cellcolor{greenheader}\textbf{Scientist effort} &
\cellcolor{blueheader}\textbf{Scientists} &
\cellcolor{blueheader}\textbf{This study} \\
\hline
\cellcolor{gray!10}Fly Body Tracking &
\cellcolor{greenlight}Video &
\cellcolor{greenlight}Days-months &
\cellcolor{bluelight}Manual, heuristics &
\cellcolor{bluelight}MOTA $>$ scientist \\
\hline
\cellcolor{gray!20}Keypoint Tracking &
\cellcolor{greendark}Training data &
\cellcolor{greendark}Days-months &
\cellcolor{bluedark}Manual, comparison to GT &
\cellcolor{bluedark}\%ile error $\leq$ $(1+\epsilon){\cdot}$ scientist\\
\hline
\cellcolor{gray!10}Registration &
\cellcolor{greenlight}Trajectories &
\cellcolor{greenlight}Hours-days &
\cellcolor{bluelight}Manual&
\cellcolor{bluelight}Error vs scientist $< \epsilon$ \\
\hline
\cellcolor{gray!20}Behavior Feature Computation &
\cellcolor{greendark}Trajectories &
\cellcolor{greendark}Hours-days &
\cellcolor{bluedark}Matches specs &
\cellcolor{bluedark}Error vs GT $< \epsilon$ \\
\hline
\cellcolor{gray!10}Walking Behavior Classification &
\cellcolor{greenlight}Behavior time series, training data &
\cellcolor{greenlight}Days-months &
\cellcolor{bluelight}Manual, comparison to GT &
\cellcolor{bluelight}Accuracy $\geq (1-\epsilon){\cdot}$scientist \\
\hline
\cellcolor{gray!20}Gait Segmentation &
\cellcolor{greendark}Trajectories, behavior classes, training data &
\cellcolor{greendark}Days &
\cellcolor{bluedark}Manual, comparison to GT &
\cellcolor{bluedark}Bout-level error vs scientist $< \epsilon$ \\
\hline
\cellcolor{gray!10}Statistical Comparisons &
\cellcolor{greenlight}Behavior classes, exp.~var., behavior time series &
\cellcolor{greenlight}Hours &
\cellcolor{bluelight}Matches specs &
\cellcolor{bluelight}Significance tests match scientist\\
\hline
\end{tabular}
\end{tiny}
    \end{minipage}\hspace{.1cm}%
    \caption{\small \emph{Left}: Comparison of our case study to existing benchmarks. \emph{Task realism} refers to how well tasks mirror what scientists would like to delegate to an AI agent. \emph{Evaluation rigor} refers to whether criteria are sufficient to convince domain experts that a solution is trustworthy. \emph{Right}: Properties of each pipeline stage, organized by task realism and evaluation rigor. For each stage, we describe the input modalities and estimated expert effort required to implement it, the methods scientists use to validate solutions in practice, and the corresponding quantitative criteria used in this study.}
    \label{fig:task_table}
    \vspace{-0.398cm}
\end{figure}

Our contributions are:
\begin{tightitemize}
\item \emph{Data-to-discovery pipeline tasks}. We develop a comprehensive, end-to-end evaluation framework based on a fly optogenetics data-to-discovery pipeline, designed to evaluate agents for scientific validity on 7 unique tasks individually and combined. 
\item \emph{Task and data size}. Our tasks are substantially larger than those in existing benchmarks in terms of lines of code (a proxy for human effort), and closer to what scientists actually want agentic AI to do: solve time-consuming bottlenecks (\cref{fig:task_table}). Datasets are correspondingly larger, requiring computationally efficient solutions---a constraint small-scale benchmarks do not impose or measure.
\item \emph{Generalization}. Real scientific data is messy and variable; solutions must generalize to unseen, potentially out-of-distribution inputs. We examine how agents handle this challenge and where brittleness to data variability manifests.
\item \emph{Stages vs.\ pipelines}. We evaluate agents both on individual stages and on the full end-to-end pipeline, revealing a gap between stage-level and pipeline-level feasibility.
\item \emph{Breadth vs.\ depth}. Rather than sampling broadly across domains, we take a depth-first approach on a single pipeline, enabling detailed characterization of task design, evaluation criteria, and agents' success and failure modes. We manually examine agents' solutions and trajectories, quantifying how often iteration leads to improvement and auditing their use of data visualization as a self-evaluation mechanism. Tasks range from rigid, fully specified problems to open-ended steps with a space of acceptable solutions, allowing us to compare agents and task properties quantitatively.
\end{tightitemize}

\section{Related work}

\textbf{LLMs and AI agents in science.} There has been rapidly growing interest in leveraging LLMs and AI agents to accelerate scientific discovery \citep{amodei2024machines,gridach2025agenticaiscientificdiscovery}. General-purpose agents with access to diverse scientific tools \citep{tusoai, googleagent} have demonstrated promising capabilities in biomedicine \citep{biomni}, chemistry \citep{chemcrow}, and biomedical image analysis \citep{simpleagents}, while more specialized agents have been designed for tasks like gene perturbation \citep{biodiscoveryagent} and editing \citep{crispragent}. Recent work has pushed further toward fully autonomous AI scientists \citep{Sakana2026, kosmos, AICoScientist2025, Robin2025}. Our work complements these efforts by rigorously characterizing where general-purpose agents succeed and fail within a complex, multi-stage scientific pipeline.

\textbf{Science benchmarks.} Benchmarking AI agents in science remains challenging, with open questions about how to subdivide the problem space, automatically assess performance on open-ended problems, and construct representative datasets to support reliable conclusions.

Most closely related to our work is ScienceAgentBench \citep{ScienceAgentBench2025}, which evaluates agents on software-development tasks across scientific domains and parts of the data-to-discovery pipeline. However, its tasks are on average 10x smaller (41 vs 447 lines of code, a proxy for expert effort) than the pipeline stages we consider, thus less representative of the time-consuming bottlenecks scientists actually want to automate. In addition, datasets are small, on average just 54 MB, over 250x smaller than those in our tasks. While evaluation criteria are task-specific, the reasoning behind thresholds is not documented. For example, the cutoff for a cell-counting task was an MAE of 30 cells, despite images containing as few as 9 cells, a threshold that may reflect insufficient depth of domain expertise, and differs from the evaluation criterion reported in the source paper for that dataset~\citep{Carpenter2006}. Given the benchmark's breadth, this may be an unavoidable tradeoff. 

DiscoveryBench \citep{discoverybench} and BLADE \citep{blade} focus on hypothesis generation and testing through retrieval from small tabular datasets already abstracted to concepts.
While these tasks require scientific reasoning skills, they would require little scientist effort (on average 24 and 3 lines of code), and they are not representative of bottlenecks to scientists' progress. 
Both approaches rely on LLMs as a judge to measure success, a measure whose reliability for assessing scientific validity has not been established~\citep{JudgeBench2025}. BLADE is exceptional in that it attempts to characterize the full solution space rather than a single ground truth, but is restricted to problems where that solution space is enumerable, limiting its applicability to more open-ended or computationally intensive stages. 
Neither benchmark addresses the kind of multi-stage, high-precision computational pipelines we study, and neither validates with the rigor required for scientific validity.

DS-1000 \citep{DS1000_2022} focuses on data science problems drawn from StackOverflow, most requiring a few lines of code with no data input, useful for measuring general programming competence but not representative of the complexity of scientific analysis stages. 

\textbf{Data-to-discovery pipelines in neuroscience.} The majority of experiments in modern neuroscience necessitate complex sequences of analyses that have evolved alongside advancements in data acquisition \citep{FlyDisco2024,AllenInstitute2021VisualBehavior2P,IBL2022BrainwideMapWhitepaper,zheng2018complete, 10.7554/eLife.80660,winnubst2019reconstruction}. These workflows typically require the integration of multiple data modalities, such as microscopy, neural dynamics, and behavior, where processing any single modality involves a distinct, multi-stage pipeline. Data-to-discovery pipelines transform raw experimental data into abstracted representations that scientists can reason about. This transformation involves two interleaved processes: \emph{quantification}: extracting measurable signals from raw data, such as tracking positions from video or counting cells from images---and \emph{conceptualization}: iterative decisions about what to measure, how to define it, and what level of abstraction is appropriate for the scientific question at hand.

Here, we focus on behavioral analysis in fly optogenetics \citep{Schretter2025, rubin2025networks, FlyDisco2024}, a modality requiring pipelines that chain together tracking, pose estimation, feature extraction, behavior classification, and statistical analyses. Ultimately, many modern experiments aim to synthesize these diverse streams of data into a unified interpretation~\citep{international2025brain}, which also significantly increases the complexity of the underlying computational workflows.

\vspace{-0.1cm}
\section{Task design}
\vspace{-0.1cm}
\label{sec:task_design}

Designing this case study's prompts and evaluation criteria was an iterative process: prompts must be precise without being overly prescriptive, and evaluation must be robust to valid variation while reflecting the high bar for correctness in science---a standard that is challenging to automate, given the unreliability of LLM-as-a-judge \citep{JudgingTheJudges, NeurodataWithoutBoredom}. \cref{sec:appendix_task_design} details the shared structure of prompts and evaluations across tasks. Below, we distill observations from this process that may inform future scientific benchmarks. Each reflects a specific revision we made, which we detail in the Appendix.

\begin{itemize}[noitemsep,leftmargin=*]
	\item \textbf{Avoid over-prescription in prompts.} When prompts describe a specific algorithm rather than the problem itself, the task reduces to rote implementation. Early prompts designed with domain experts often included the scientist's algorithmic choices, which constrained agents to particular methods (\cref{sec:appendix_gait}). Including extraneous implementation details can distract agents from the core task (\cref{sec:appendix_body_tracking}).
	\item \textbf{Avoid under-specification in prompts.} Ambiguous problem definitions admit multiple valid interpretations, making evaluation difficult. Under-specified tasks meant solutions could differ from the reference for arguably valid reasons, requiring iteration on both prompts and evaluation criteria based on results (\cref{sec:appendix_perframe}, \cref{sec:appendix_stat_comp}).
	\item \textbf{Isolate evaluation from scientist-specific implementation choices.} Scientist solutions may contain incidental choices that are not intrinsic to correctness. Evaluation should not penalize agents for omitting unstated steps that the scientist happened to include, nor should they assume agents will make the same implementation decisions. Evaluation should admit any solution satisfying the core definition (\cref{sec:appendix_stat_comp}).
	\item \textbf{Robust evaluation instead of prompt constraints.} Rather than adding prompt constraints to simplify evaluation, invest effort in making evaluation robust to valid variation. This keeps tasks realistic (\cref{sec:appendix_registration}, \cref{sec:appendix_perframe}) and avoids over-indexing to evaluation convenience.
	\item \textbf{Specify evaluation criteria in the prompt.} Agents benefit from knowing how they will be evaluated. When prompts lacked success criteria, agent solutions tended to not meet the high bar needed for downstream tasks. Adding explicit metrics and target thresholds enabled agents to monitor their own performance against the provided data (\cref{sec:appendix_keypoint}).
	\item \textbf{Validate prompts and scientist solutions.} Multiple reviewers including domain experts should verify that prompts contain all necessary information and no extraneous details. Domain expertise is particularly valuable for identifying implicit assumptions that may be obvious to specialists but absent from the prompt. This process uncovered bugs in scientists' reference implementations that had survived peer review (\cref{sec:appendix_registration}, \cref{sec:appendix_perframe}).
\end{itemize}

\vspace{-.2cm}
\subsection{Evaluation criteria design}
\vspace{-.15cm}
\label{sec:evaluation_design}
A central challenge in evaluating agent solutions is defining what constitutes ``correct.'' In scientific workflows, correctness is rarely binary, and ground truth labels are often unavailable: many tasks require scientists to assess correctness through qualitative inspection and iterative refinement. Our evaluation criteria aim to answer: \emph{would a domain scientist trust and use the agent's solution in place of their own?} We operationalize this by comparing agents' solutions to a scientist reference solution derived from published methods (\cref{sec:appendix_reference_sources}). We employ two primary evaluation strategies depending on whether ground truth labels are available. %
\begin{tightitemize}
\item \textbf{Ground truth available.} For tasks where labeled ground truth exists (Body Tracking, Keypoint Tracking, Walking Behavior Classification), we compare both the agent's and scientist's solutions against the ground truth. The agent passes if its score is within a tolerance of the scientist's:
$\operatorname{score}(\text{agent}, \text{GT}) \geq (1 - \epsilon) \cdot \operatorname{score}(\text{scientist}, \text{GT})$.
\item \textbf{Ground truth unavailable.} For tasks where ground truth labels do not exist or are impractical to obtain (Behavior Feature Computation, Registration, Statistical Comparisons), we treat the scientist's solution as the reference and measure deviation from it. The agent passes if all outputs $i$ are sufficiently close:
$\max_i |\text{agent}_i - \text{scientist}_i| \leq \epsilon$.
\end{tightitemize}
Here, $\epsilon$ is a task-specific tolerance set in consultation with domain scientists to reflect variability between valid solutions and the sensitivity of downstream analyses to errors. A tolerance-sensitivity analysis (\cref{sec:sensitvity_analysis}) shows that results are robust to reasonable variation in $\epsilon$.

\textbf{Alternative approaches.} A third strategy we did not employ but merits consideration is measuring inter-scientist agreement: $\operatorname{dist}(\text{scientist}_1, \text{scientist}_2)$. This would establish a baseline for acceptable variation by quantifying how much two domain experts differ on the same task, providing a principled threshold for agent evaluation.

\vspace{-0.1cm}
\subsection{Task descriptions}
\label{sec:task_descriptions}

We evaluate agents under two settings: \emph{single-stage} tasks, which isolate individual pipeline components, and \emph{end-to-end} tasks, which require executing the full workflow (\cref{fig:pipeline_overview}). Evaluation criteria are detailed in \cref{sec:appendix_eval}, and the scientific abilities tested in \cref{sec:appendix_all_tasks_table}.

\begin{tightitemize}
    \item \textbf{Body Tracking.}
    Tracks positions and identities of multiple interacting flies over time, a long-standing computer vision problem. Agents are given no ground truth or optimizable criterion, and so must devise their own evaluation criteria.
    \item \textbf{Registration.} Scales and translates trajectories into a common coordinate system, and cleans frames where tracking is missing. This is a comparatively straightforward step.
    \item \textbf{Behavior Feature Computation.} From trajectories, computes interpretable behavior features. We provide equations defining features, but not implementation details.
    \item \textbf{Walking Behavior Classification.} Classifies whether each fly is walking forward in each frame, given a labeled training dataset.
    \item \textbf{Keypoint Tracking.} Estimates positions of 21 keypoints per fly, with pixel-level precision needed for downstream analyses. The criterion is well-defined, but the approach is not.
    \item \textbf{Gait Segmentation.} Segments each leg's motion into swing (lifted off the ground) and stance (touching the ground) phases, a classical locomotion analysis typically solved with thresholding heuristics. We provide a small set of labeled examples for tuning.
    \item \textbf{Statistical Comparisons.} Statistically tests whether optogenetic perturbation affects stance bout duration for 10 GAL4 lines compared to a genetic control group.
    \item \textbf{End-to-End Minimal Prompt.} Given the raw data and a description of the final scientific goal, produces the full analysis. No explicit stage decomposition is given, thus success can only be measured by the correctness of the final output.
    \item \textbf{End-to-End Maximal Prompt.} Given the raw data plus {\em all} stage-level prompts, produces the full analysis. This tests whether agents can manage a long-context, multi-step task. Intermediate stages are evaluated independently, in addition to the final result.
\end{tightitemize}

\section{Results}
\label{sec:results}

\subsection{Quantitative results}
\label{sec:quantiative_results}

\cref{table:results} summarizes agent performance across single-stage and end-to-end tasks. We report pass (\mrgreen) and fail (\mryellow/\mrred) outcomes for three trials per agent-task pair, along with a task-specific score (higher is better). Trial-to-trial variance is low, with 73\% of agent-task pairs yielding unanimous pass/fail verdicts. Definitions of pass/fail and the task scores are in \cref{sec:appendix_eval}, with details on per-trial token counts and wall-clock runtimes in \cref{sec:appendix_usage}.

\textbf{Single-stage tasks.} Agents achieve reliable performance on several individual pipeline stages. Feature Computation and Behavior Classification are solved consistently across all agents, with nearly all trials passing in the single-task setting. Keypoint Tracking and Registration are similarly reliable for most agent configurations, with the exception of \texttt{terminus-2/gpt-5.4}, which passes zero trials. Two stages prove difficult for all agents. Body Tracking yields zero passes across all agent configurations, suggesting that this task, which requires sustained visual reasoning and identity maintenance through occlusions, remains beyond current capabilities. Gait Segmentation is similarly challenging, with only \texttt{terminus-2/claude-opus-4-6} achieving any passes.

\textbf{End-to-end tasks.} Performance degrades markedly in the end-to-end setting. Under the maximal prompt, stages that agents solve reliably in isolation often fail when composed into a pipeline. Registration and Keypoint Tracking, which both agents largely pass as single-stage tasks, see reduced pass rates end-to-end, and Behavior Classification degrades similarly, though to a lesser degree. Notably, all three stages are evaluated independently of upstream outputs (e.g., on generalization to held-out test sets) and thus reflect genuine difficulty in the composed setting rather than propagated errors. Statistical Comparison, which Opus-4.6-backed agents solve in the single-stage setting, yields zero passes end-to-end for all configurations; unlike the stages above, it depends on accumulated outputs from all prior stages, and its failure reflects both compositional difficulty and prior stage degradation. The minimal prompt is similarly challenging: neither agent configuration produces a passing solution for Statistical Comparison, indicating that, for the scale of tasks studied here, current agents struggle to construct and execute multi-stage workflows without explicit decomposition. Together, these results suggest that managing long-context instructions and sustaining attention over extended tasks pose challenges beyond solving the individual subproblems.

\begin{table*}[t]
\centering
\small
\setlength{\tabcolsep}{2.4pt}
\renewcommand{\arraystretch}{0.92}
\setlength{\aboverulesep}{2pt}
\setlength{\belowrulesep}{2pt}
\newcommand{\mrna}{\textcolor{gray}{--}}
\newcommand{\mrscore}[1]{\\[-3pt]\rule[-2pt]{0pt}{9pt}\scriptsize #1}
\newcommand{\mrablation}[1]{\hspace{1.2cm}{\scriptsize + #1}}
\newcommand{\mrablcell}[1]{\cellcolor{gray!4}#1}
\newcommand{\mrablmakecell}[1]{\cellcolor{gray!4}\begingroup\setlength{\fboxsep}{0pt}\colorbox{gray!4}{\makecell{#1}}\endgroup}
\resizebox{\textwidth}{!}{%
\begin{tabular}{m{4.6cm}ccccccc}
\toprule
\textbf{Agent} & \makecell{\textbf{Body}\\\textbf{Tracking}} & \makecell{\textbf{Registration}} & \makecell{\textbf{Keypoint}\\\textbf{Tracking}} & \makecell{\textbf{Feature}\\\textbf{Computation}} & \makecell{\textbf{Behavior}\\\textbf{Classifier}} & \makecell{\textbf{Gait}\\\textbf{Segmentation}} & \makecell{\textbf{Statistical}\\\textbf{Comparison}} \\
\midrule
\rowcolor{gray!12}\multicolumn{8}{l}{\textbf{Single-stage tasks}} \\
\addlinespace[1pt]
\texttt{\small claude-code/claude-opus-4-6} & \makecell{\mrred\ \mrred\ \mrred\mrscore{0.017$\pm$0.003}} & \makecell{\mrred\ \mrgreen\ \mrgreen\mrscore{1.928$\pm$0.125}} & \makecell{\mrgreen\ \mrgreen\ \mrgreen\mrscore{\textbf{1.376$\pm$0.055}}} & \makecell{\mrgreen\ \mrgreen\ \mrgreen\mrscore{\textbf{1.855$\pm$0.000}}} & \makecell{\mrgreen\ \mrgreen\ \mrgreen\mrscore{2.281$\pm$0.108}} & \makecell{\mrred\ \mrred\ \mrred\mrscore{0.451$\pm$0.387}} & \makecell{\mrgreen\ \mrgreen\ \mrgreen\mrscore{\textbf{1.000$\pm$0.000}}} \\
\texttt{\small codex/gpt-5.4} & \makecell{\mrred\ \mrred\ \mrred\mrscore{0.020$\pm$0.006}} & \makecell{\mrgreen\ \mrgreen\ \mrgreen\mrscore{1.999$\pm$0.002}} & \makecell{\mrgreen\ \mrgreen\ \mrgreen\mrscore{1.355$\pm$0.286}} & \makecell{\mrgreen\ \mrgreen\ \mrgreen\mrscore{\textbf{1.855$\pm$0.000}}} & \makecell{\mrgreen\ \mrred\ \mrgreen\mrscore{1.167$\pm$0.396}} & \makecell{\mrred\ \mrred\ \mrred\mrscore{0.929$\pm$0.039}} & \makecell{\mryellow\ \mrred\ \mrgreen\mrscore{0.733$\pm$0.379}} \\
\texttt{\small terminus-2/claude-opus-4-6} & \makecell{\mrred\ \mrred\ \mrred\mrscore{\textbf{0.044$\pm$0.022}}} & \makecell{\mrgreen\ \mrgreen\ \mrgreen\mrscore{\textbf{2.000$\pm$0.000}}} & \makecell{\mrgreen\ \mrgreen\ \mrgreen\mrscore{1.371$\pm$0.046}} & \makecell{\mrgreen\ \mryellow\ \mrgreen\mrscore{1.851$\pm$0.006}} & \makecell{\mrgreen\ \mrgreen\ \mrgreen\mrscore{\textbf{3.354$\pm$1.019}}} & \makecell{\mrgreen\ \mrred\ \mrgreen\mrscore{\textbf{1.201$\pm$0.225}}} & \makecell{\mrgreen\ \mrgreen\ \mrgreen\mrscore{\textbf{1.000$\pm$0.000}}} \\
\texttt{\small terminus-2/gpt-5.4} & \makecell{\mrred\ \mrred\ \mrred\mrscore{0.005$\pm$0.008}} & \makecell{\mryellow\ \mryellow\ \mrred\mrscore{1.892$\pm$0.090}} & \makecell{\mrred\ \mrred\ \mrred\mrscore{0.464$\pm$0.367}} & \makecell{\mrgreen\ \mrgreen\ \mrred\mrscore{1.662$\pm$0.334}} & \makecell{\mrgreen\ \mrgreen\ \mrgreen\mrscore{1.260$\pm$0.233}} & \makecell{\mrred\ \mrred\ \mrred\mrscore{0.682$\pm$0.127}} & \makecell{\mryellow\ \mryellow\ \mrgreen\mrscore{0.933$\pm$0.058}} \\
\midrule
\rowcolor{gray!12}\multicolumn{8}{l}{\textbf{E2E Maximal}} \\
\addlinespace[1pt]
\texttt{\small claude-code/claude-opus-4-6} & \makecell{\mrred\ \mrred\ \mrred\mrscore{0.020$\pm$0.002}} & \makecell{\mryellow\ \mrred\ \mryellow\mrscore{1.918$\pm$0.052}} & \makecell{\mrred\ \mrred\ \mrgreen\mrscore{\textbf{0.702$\pm$0.640}}} & \mrna & \makecell{\mrred\ \mrgreen\ \mrgreen\mrscore{\textbf{1.139$\pm$0.986}}} & \mrna & \makecell{\mrred\ \mrred\ \mrred\mrscore{\textbf{0.467$\pm$0.058}}} \\
\mrablcell{\mrablation{gold body tracking}} & \mrablcell{\mrna} & \mrablmakecell{\mryellow\ \mrred\ \mryellow\mrscore{1.918$\pm$0.052}} & \mrablmakecell{\mrred\ \mrred\ \mrgreen\mrscore{0.702$\pm$0.640}} & \mrablmakecell{\mrred\ \mrred\ \mrred\mrscore{1.622$\pm$0.086}} & \mrablmakecell{\mrred\ \mrgreen\ \mrgreen\mrscore{1.139$\pm$0.986}} & \mrablmakecell{\mrred\ \mrred\ \mrred\mrscore{0.247$\pm$0.200}} & \mrablmakecell{\mrred\ \mrred\ \mrred\mrscore{0.533$\pm$0.252}} \\
\mrablcell{\mrablation{all gold upstream}} & \mrablcell{\mrna} & \mrablcell{\mrna} & \mrablcell{\mrna} & \mrablcell{\mrna} & \mrablcell{\mrna} & \mrablcell{\mrna} & \mrablmakecell{\mrred\ \mrgreen\ \mrgreen\mrscore{0.933$\pm$0.115}} \\
\addlinespace[1pt]
\texttt{\small codex/gpt-5.4} & \makecell{\mrred\ \mrred\ \mrred\mrscore{\textbf{0.039$\pm$0.038}}} & \makecell{\mrgreen\ \mryellow\ \mrred\mrscore{\textbf{1.948$\pm$0.063}}} & \makecell{\mrred\ \mrred\ \mrred\mrscore{0.247$\pm$0.427}} & \mrna & \makecell{\mrred\ \mrgreen\ \mrred\mrscore{1.123$\pm$1.946}} & \mrna & \makecell{\mrred\ \mrred\ \mrred\mrscore{0.300$\pm$0.100}} \\
\mrablcell{\mrablation{gold body tracking}} & \mrablcell{\mrna} & \mrablmakecell{\mrgreen\ \mryellow\ \mrred\mrscore{1.948$\pm$0.063}} & \mrablmakecell{\mrred\ \mrred\ \mrred\mrscore{0.247$\pm$0.427}} & \mrablmakecell{\mrred\ \mrgreen\ \mrred\mrscore{1.554$\pm$0.472}} & \mrablmakecell{\mrred\ \mrgreen\ \mrred\mrscore{1.123$\pm$1.946}} & \mrablmakecell{\mrred\ \mrred\ \mrred\mrscore{0.300$\pm$0.033}} & \mrablmakecell{\mrred\ \mrred\ \mrred\mrscore{0.333$\pm$0.115}} \\
\mrablcell{\mrablation{all gold upstream}} & \mrablcell{\mrna} & \mrablcell{\mrna} & \mrablcell{\mrna} & \mrablcell{\mrna} & \mrablcell{\mrna} & \mrablcell{\mrna} & \mrablmakecell{\mrred\ \mrgreen\ \mrred\mrscore{\textbf{0.833$\pm$0.153}}} \\
\midrule
\rowcolor{gray!12}\multicolumn{8}{l}{\textbf{E2E Minimal}} \\
\addlinespace[1pt]
\texttt{\small claude-code/claude-opus-4-6} & \mrna & \mrna & \mrna & \mrna & \mrna & \mrna & \makecell{\mrred\ \mrred\ \mrred\mrscore{\textbf{0.467$\pm$0.058}}} \\
\texttt{\small codex/gpt-5.4} & \mrna & \mrna & \mrna & \mrna & \mrna & \mrna & \makecell{\mrred\ \mrred\ \mrred\mrscore{0.233$\pm$0.153}} \\
\bottomrule
\end{tabular}}
\caption{\small Agents' performance on single-stage and end-to-end tasks, including oracle ablations for E2E Maximal. \mrgreen: passed stringent criteria for nearly all test videos/lines; \mryellow: passed on average; \mrred: failed. Numbers are continuous scores (mean $\pm$ standard deviation over three trials), 1.0 = passing. Dashes mark stages not separately evaluable, either because evaluation required correct upstream Body Tracking or the agent was not required to follow the reference stages. Indented rows inject gold Body Tracking inputs, or gold inputs at all stages upstream of Statistical Comparison.}
\label{table:results}
\vspace{-0.317cm}
\end{table*}

\textbf{End-to-end oracle ablations.} To localize end-to-end failure across stages, we injected ground truth inputs at two points in the agent-produced E2E Maximal pipelines (\cref{table:results}). We injected gold Body Tracking trajectories to isolate the effect of that stage's low performance, with no improvement: the final stage continues to fail in all trials. This shows that the final stage's failure is not driven by upstream tracking quality alone. It also allows us to evaluate the two intermediate stages which could not be compared to the reference unless the tracked targets match. Both are degraded relative to single-stage performance, consistent with other stages. Next, to further isolate failures in the final stage, we injected gold outputs from all upstream stages. Scores improve substantially relative to E2E for both agents, but remain below single-stage for Claude Code, reflecting both upstream degradation and weaknesses in the stage implementation itself. For Codex, gold-input and single-stage scores are comparable within noise, implicating upstream degradation alone.

\vspace{-0.1cm}
\subsection{Failure modes}\label{ssec:qualitative_failure}
\vspace{-0.1cm}

To diagnose common failure modes, we manually reviewed the agents' code and trajectories to trace their stated reasoning as they iterated on their solutions. Below, we describe observed failure modes, along with the hypothesized task properties that cause them.

\begin{figure}[t]
    \centering
    \includegraphics[width=\linewidth]{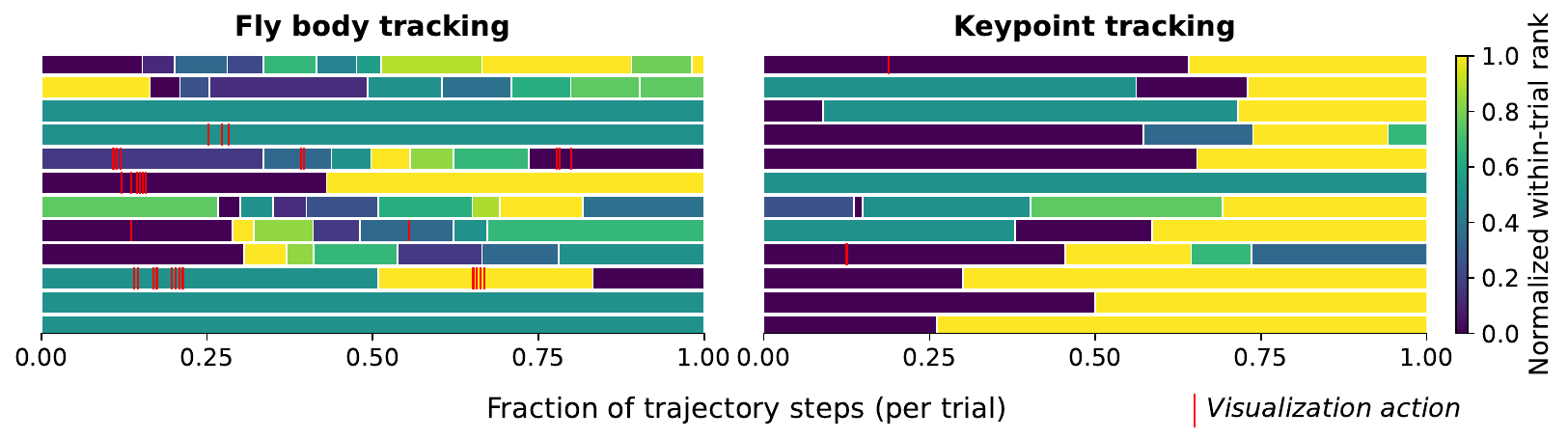}\vspace{-0.2cm}
    \caption{\small Per-trial iteration trajectories for Body Tracking and Keypoint Tracking. The two tasks span contrasting evaluation regimes: Body Tracking requires scientific judgment to assess output quality, while Keypoint Tracking provides a concrete, pre-defined ML metric. Each bar is one trial, segmented by iteration in chronological order, with segment width proportional to trajectory steps. Segment color encodes each iteration's within-trial rank (dark=worst, bright=best), scored from a code snapshot at that iteration. Red ticks mark steps where agents visualized an image. Trials are sorted by final-iteration value per task in descending order. Full seven-task view in \cref{fig:appendix_all-trajectories}.}
    \label{fig:subset-trajectories}
    \vspace{-0.447cm}
\end{figure}

\begin{figure}[t]
    \centering
    \includegraphics[width=\linewidth]{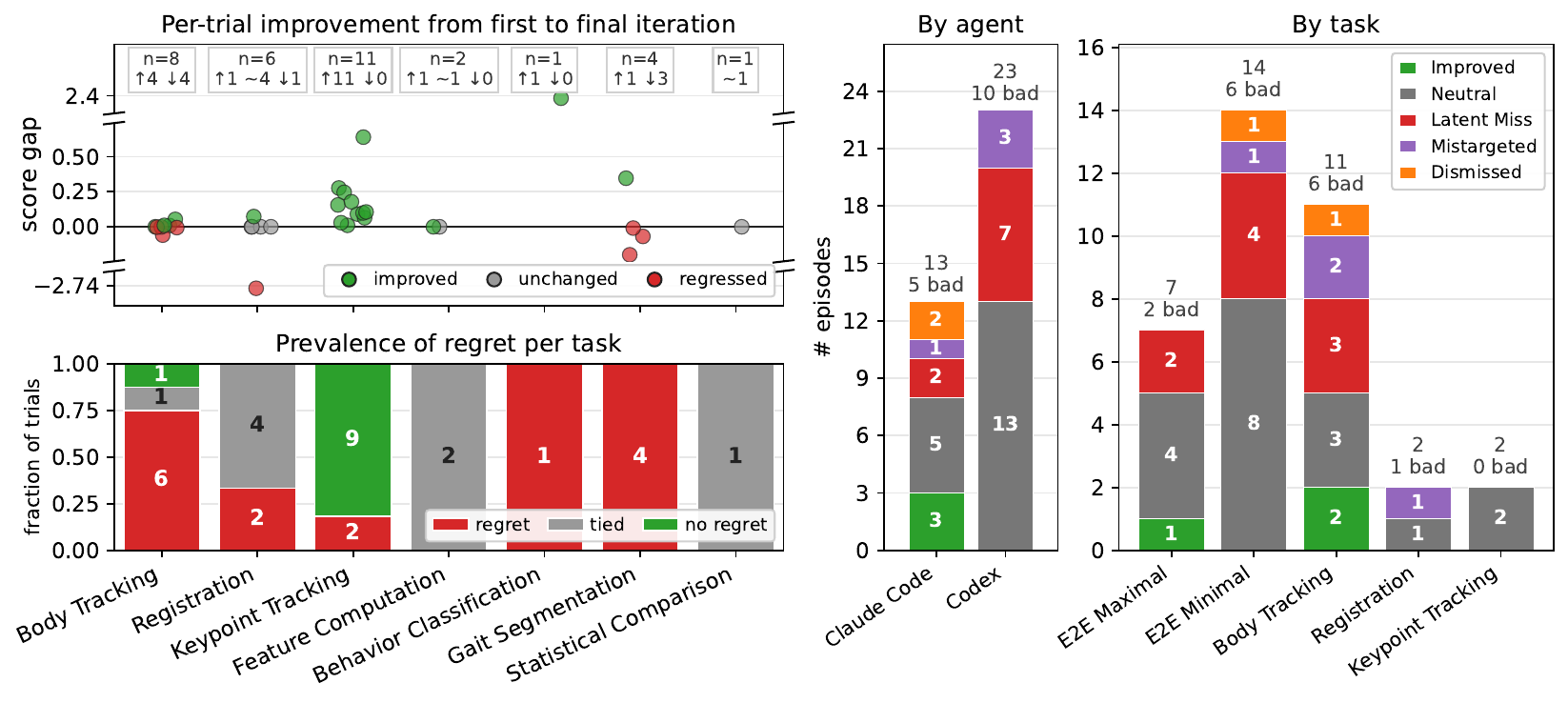} \vspace{-.45cm}
    \caption{\small Iteration quality and image-view episode outcomes. \emph{Left, top}: per-trial change in score (\cref{sec:appendix_eval}) from the agent's first to its submitted solution, for trials with $\geq$2 iterations; boxes give counts of improved ($\uparrow$), regressed ($\downarrow$), and unchanged ($\sim$) trials. \emph{Left, bottom}: per-task prevalence of \textit{regret}: submitting a solution worse than the best reached at an earlier iteration. Bars give the fraction of trials with regret (final $<$ best non-final), tied, or no regret. \emph{Right}: image-view episode outcomes by agent and by task, classified as Improved, Neutral, Mistargeted, Latent Miss, or Dismissed (\cref{ssec:qualitative_failure}). Terminus-2 trials make no image-view calls and are omitted, as are tasks with zero episodes.}
    \label{fig:iterations-imageviews}
    \vspace{-0.066cm}
\end{figure}

\textbf{Agents fail to iterate productively when self-evaluation requires scientific judgment.} Tasks that do not provide a clear evaluation metric leave the agent to devise its own criterion; if unsuccessful, iterations will not improve. We see this in Body Tracking, where agents are not provided ground truth to compare to (\cref{table:expanded_tasks}). Scientists doing this task visualize intermediate outputs and devise metrics for assessing different types of mistakes. In \cref{fig:subset-trajectories}, agents perform many iterations and even visualize intermediate results, yet scores do not reliably improve. The iteration-quality gap analysis (Fig.~\ref{fig:iterations-imageviews}) supports this: 6/8 trials exhibited \textit{regret} (the agent did not submit its best solution), and 4/8 submitted solutions worse than their first. Fig.~\ref{fig:appendix_ex-poor-it} visualizes a submitted solution with major tracking errors that humans would not ignore: all true detections offset from flies and frequent false detections. The pattern extends across other tasks lacking a scalar metric (Registration, Feature Computation, and Statistical Comparison). This stands in contrast to ML-style tasks (Keypoint Tracking, Behavior Classification), which iteratively improve (\cref{ssec:qualitative_success}).

\textbf{Agents fail to translate visual evidence into corrective action.} Visualizing intermediate outputs is a key part of scientific analysis. Agents mimic this pattern with image-viewing tool calls, but these rarely yield a sensible update. To study this systematically, we grouped contiguous runs of image-view calls answering one investigative question into distinct \textit{episodes}, and manually classified them based on the agent's code updates and stated reasoning:
\begin{tightitemize}
\item Improved: the visualization exposed an implementation issue in the agent's solution, the agent correctly diagnosed it, and proceeded to fix it.
\item Neutral: the visualization was used to confirm correctness or characterize data, with no underlying issue to act on.
\item Mistargeted: the visualization was uninformative for the agent's posed question.
\item Latent Miss: the visualization surfaced a real implementation issue but the agent misinterpreted it or failed to perceive it.
\item Dismissed: the agent perceived the issue but rationalized it as benign. 
\end{tightitemize}
These failure modes isolate failures of visual perception (Latent Miss) from failures of reasoning both upstream (Mistargeted) and downstream (Dismissed) of the visualization itself, all of which cause underlying issues to persist across iterations. In Claude Code and Codex episodes (Fig.~\ref{fig:iterations-imageviews}), the two agent frameworks that support image viewing, all three failure modes are common, and failures outnumber improvements across every task with observed image-view calls. Consistent with this, Body Tracking trials in \cref{fig:subset-trajectories} accumulate the densest image-view coverage while solutions do not improve monotonically, with some visualizations driving active regression. We present a representative Latent Miss episode per agent (\cref{sssec:appendix_ex-latent-miss-claude}, \cref{sssec:appendix_ex-latent-miss-codex}), where a visualization exposing a real implementation issue was misinterpreted. These failures on seemingly straightforward visual reasoning suggest that vision-based scientific reasoning is a current weakness and an area for future development.

\textbf{Agents simplify pipeline decompositions.} Agents consistently use fewer sub-tasks in the E2E Minimal task: only the first and last stages survive in every trial, with the rest fused or replaced by simpler, less accurate implementations.
While the most aggressive stage merging produced the worst score (1/10), closely replicating the canonical decomposition did not guarantee success when a key stage was replaced by a poor shortcut.
Indeed, even when the decomposition is provided under the maximal prompt, agents perform no better on the final stage: across all 12 E2E trials, they identify essentially no fly lines with significant behavioral effects (1/48). This suggests that the bottleneck for minimal-prompt E2E performance is not discovering the decomposition but implementing its stages. These implementations degrade end-to-end relative to the single-stage setting (\cref{table:results}); we attribute this to challenges in long-horizon task management and long-context retrieval.

\textbf{Attention and effort suffer in maximal end-to-end tasks.} The maximal E2E task sees basic formatting errors absent from single-stage tasks, where each workspace contains only one model to submit. Across six Maximal trials, nearly half of submitted keypoint and walking-classifier model files violate the prompt-specified interface and cannot be evaluated, despite the same information being available in both settings. Maximal E2E solutions are also systematically simpler than their single-stage counterparts (Claude Code's keypoint networks train for fewer epochs; Codex's Keypoint Tracking trials drop the data augmentation every Codex single-stage uses; walking classifiers omit rolling-window features every single-stage submission computes). We thus hypothesize that these agents perform worse on larger problems, even when it is trivial to decompose the problem.

\textbf{Agents lack consistent judgment about reasonable runtimes.} Agents sometimes kill training processes on weak or fabricated evidence, particularly in tasks involving long-running scripts. One recurrent failure mode is misinterpreting a process's aggregate CPU time as wall-clock time; in 4 instances, the agent prematurely terminated training after asserting that a script had been running 13× to 42× longer than it had (\cref{sssec:appendix_ex-overestimate}). Agents also reason inconsistently about training budgets within trials. In one Keypoint Tracking trial (\cref{sssec:appendix_ex-premature}), the agent killed a heatmap-based training run after 3 epochs, concluding it had ``plateaued,'' then switched to coordinate regression, which performed worse initially but was allowed to continue. These failures suggest the importance of measuring agent performance on larger tasks, which produce unique types of failure.

\textbf{Agent framework and model choice impact environment handling.} GPT-5.4-based agents frequently avoid installing new dependencies, at times reimplementing standard library functionality. They ran \texttt{pip install} in only $\sim$25\% of trials, compared to $\sim$70\% for Opus-based agents. For example, one agent abandoned its initial \texttt{cv2}-based solution for a subpar NumPy reimplementation, leading to a worse-performing iteration (\texttt{zZMfrKG} in Fig.~\ref{fig:appendix_all-trajectories}). For the Terminus-2 framework, 5/42 trials encountered temporarily unresponsive shells, causing agents to repeatedly emit stuck messages and consume substantial numbers of tokens (\cref{sssec:appendix_ex-infinite}). Claude Code is vulnerable to a different framework-level failure: in 2/27 trials, the agent terminated its own session by issuing a \texttt{pkill} command whose pattern matched the agent process itself, ending the trial mid-task (\cref{sssec:appendix_ex-suicide}).

\subsection{Success modes}\label{ssec:qualitative_success}

\textbf{Agents reliably solve ML-style tasks.} Two tasks fit a standard supervised learning setup, wherein labeled training data and a scalar evaluation metric are provided. For these, agents iterate productively toward the metric. Keypoint Tracking exemplifies this: trials show progressive within-trial improvement (\cref{fig:subset-trajectories}), with the best iteration reliably at or near the trajectory end. The iteration-quality analysis (\cref{fig:iterations-imageviews}) agrees: all multi-iteration trials improved from first to final iteration, with 9/11 submitting their best iteration. Notably, several Keypoint Tracking and Behavior Classification submissions matched or exceeded the scientist's solution, while no Body Tracking trials even reached the minimum threshold.

\textbf{Agents successfully leverage domain knowledge not provided in the prompt.} By design, our prompts do not specify details considered to be in-domain knowledge (e.g.\ keypoint names use anatomical terms such as ``right hind-leg tarsus'' as opposed to ``leg tips''). We find that agents consistently understand and apply relevant domain-specific concepts in-context. In the Gait Segmentation task, the ``swing'' and ``stance'' leg phases are defined conceptually (``the leg is in contact with the ground'' / ``lifted from the ground'') with no operational criteria for detecting them from keypoint trajectories; yet, across all trials, agents demonstrated the biomechanical understanding of fruit flies needed to distinguish them: 8/12 trials' reasoning traces articulated the kinematic relationship between tarsus motion and the fly body (\cref{sssec:appendix_ex-kinematics}), and 10/12 final implementations constructed a body-aligned coordinate axis from rigid-segment keypoints to encode the classification (\cref{sssec:appendix_ex-body-aligned}).

\textbf{Agents effectively parallelize sub-tasks within complex workflows.} When the E2E pipeline includes a long-running stage, agents consistently identify other work that can proceed concurrently and pursue it via backgrounded tool calls. Across the 12 E2E trials, 6 show explicit parallelization, where the agent started one script, developed another part of the pipeline in parallel, and then returned to it later. All six are Claude Code runs; the six Codex trials instead execute the same long stages serially in the foreground. The most common parallelization comprises overlapping Keypoint Tracking training with Body Tracking development, Behavior Classifier integration, or pipeline-script construction (\cref{sssec:appendix_ex-parallel}).

\section{Discussion}

We present a depth-first case study of how well general-purpose AI agents solve stages of a data-to-discovery pipeline: a subdivision of scientific analysis natural to practicing scientists and amenable to mixing and matching as workflows evolve. Our findings were directly observed only in a single pipeline; we hypothesize they will generalize to scientific tasks sharing a high bar for precision, large-scale data, long multi-stage workflows, and reliance on data-based or visual validation---structural properties of tasks or agent abilities, rather than of a science domain. Establishing where these patterns hold across additional problems and agents is important future work, supported by our released dataset.

Individual pipeline stages in our benchmark are on average 10x larger than those in the most closely related work \citep{ScienceAgentBench2025}, yet agents solve several of them, suggesting that stage-level tasks of this complexity are tractable for current agents---an encouraging signal for near-term automation of scientific bottlenecks. The end-to-end setting, however, surfaces a qualitatively different failure mode: lapses in sustained attention across the full task. End-to-end automation thus remains an open challenge that scales non-trivially with task length.
Our pipeline also operated on substantially larger data (\textgreater250x bigger than \citet{ScienceAgentBench2025}), introducing a challenge ubiquitous in real workflows but absent from small-scale benchmarks: agents must manage time and compute by estimating runtimes and deciding when to sample, abort, or prioritize. We found mixed evidence of success here, and agentic frameworks will need to address this explicitly as they scale to realistic settings.

Scientific problems pose distinct challenges for agents beyond general ML or coding. In science, many problems exist for which no explicit mathematical criterion has yet been formulated: the scientist intuitively understands what types of errors matter and develops methods for visualizing outcomes and diagnosing failures. For example, a scientist validating a tracking algorithm will overlay tracked positions on raw video frames, or develop heuristics for identifying likely tracking failures. Such exploratory, visually-grounded validation is central to scientific practice. Current agents do attempt this behavior but largely fail to close the loop from inspection to correction. Whether stronger integration of vision models, targeted training, or prompting can address this gap remains an open question.

Finally, we offer detailed guidelines for constructing tasks and evaluation criteria, favoring depth over breadth to achieve the rigor necessary for results that practicing scientists can trust. Breadth matters too, and we hope this work encourages future collaborations between AI researchers and domain experts to expand the number of pipeline stages that can be evaluated with this level of rigor. The framework we describe---grounded in domain standards, attentive to equivalence classes and tolerances, and validated against the scientists who built the original pipelines---is, we believe, broadly applicable beyond the specific domain studied here.

\section*{Acknowledgments}
We thank Adam Taylor for help with the original Fly Disco pipeline code, Goran Ceric for help running the Harbor framework on the Janelia cluster, and the Branson and Sun labs, Ranjit Singh, Atharva Sehgal, and John Bogovic for helpful discussions. This work was supported by the Janelia Visiting Scientists program, the Howard Hughes Medical Institute, and the National Science Foundation (IIS-2505098).

\bibliography{colm2026_conference}

@article{Carpenter2006,
  author  = {Carpenter, Anne E. and Jones, Thouis R. and Lamprecht, Michael R. and Clarke, Colin and Kang, In Han and Friman, Ola and Guertin, David A. and Chang, Joo Han and Lindquist, Robert A. and Moffat, Jason and Golland, Polina and Sabatini, David M.},
  title   = {CellProfiler: image analysis software for identifying and quantifying cell phenotypes},
  journal = {Genome Biology},
  year    = {2006},
  volume  = {7},
  number  = {10},
  pages   = {R100},
  doi     = {10.1186/gb-2006-7-10-r100},
  issn    = {1474-760X},
  url     = {https://doi.org/10.1186/gb-2006-7-10-r100},
}

@misc{ScienceAgentBench2025,
      title={ScienceAgentBench: Toward Rigorous Assessment of Language Agents for Data-Driven Scientific Discovery}, 
      author={Ziru Chen and Shijie Chen and Yuting Ning and Qianheng Zhang and Boshi Wang and Botao Yu and Yifei Li and Zeyi Liao and Chen Wei and Zitong Lu and Vishal Dey and Mingyi Xue and Frazier N. Baker and Benjamin Burns and Daniel Adu-Ampratwum and Xuhui Huang and Xia Ning and Song Gao and Yu Su and Huan Sun},
      year={2025},
      eprint={2410.05080},
      archivePrefix={arXiv},
      primaryClass={cs.CL},
      url={https://arxiv.org/abs/2410.05080}, 
}

@article{DS1000_2022,
  title={DS-1000: A Natural and Reliable Benchmark for Data Science Code Generation},
  author={Lai, Yuhang and Li, Chengxi and Wang, Yiming and Zhang, Tianyi and Zhong, Ruiqi and Zettlemoyer, Luke and Yih, Wen-Tau and Fried, Daniel and Wang, Sida and Yu, Tao},
  journal={ArXiv},
  year={2022},
  volume={abs/2211.11501}
}

@article{Sakana2026,
  author    = {Lu, Chris and Lu, Cong and Lange, Robert Tjarko and Yamada, Yutaro and Hu, Shengran and Foerster, Jakob and Ha, David and Clune, Jeff},
  title     = {Towards end-to-end automation of {AI} research},
  journal   = {Nature},
  year      = {2026},
  volume    = {651},
  number    = {8107},
  pages     = {914--919},
  doi       = {10.1038/s41586-026-10265-5},
  url       = {https://doi.org/10.1038/s41586-026-10265-5},
  issn      = {1476-4687},
}

@misc{AICoScientist2025,
      title={Towards an AI co-scientist}, 
      author={Juraj Gottweis and Wei-Hung Weng and Alexander Daryin and Tao Tu and Anil Palepu and Petar Sirkovic and Artiom Myaskovsky and Felix Weissenberger and Keran Rong and Ryutaro Tanno and Khaled Saab and Dan Popovici and Jacob Blum and Fan Zhang and Katherine Chou and Avinatan Hassidim and Burak Gokturk and Amin Vahdat and Pushmeet Kohli and Yossi Matias and Andrew Carroll and Kavita Kulkarni and Nenad Tomasev and Yuan Guan and Vikram Dhillon and Eeshit Dhaval Vaishnav and Byron Lee and Tiago R D Costa and José R Penadés and Gary Peltz and Yunhan Xu and Annalisa Pawlosky and Alan Karthikesalingam and Vivek Natarajan},
      year={2025},
      eprint={2502.18864},
      archivePrefix={arXiv},
      primaryClass={cs.AI},
      url={https://arxiv.org/abs/2502.18864}, 
}

@misc{Robin2025,
      title={Robin: A multi-agent system for automating scientific discovery}, 
      author={Ali Essam Ghareeb and Benjamin Chang and Ludovico Mitchener and Angela Yiu and Caralyn J. Szostkiewicz and Jon M. Laurent and Muhammed T. Razzak and Andrew D. White and Michaela M. Hinks and Samuel G. Rodriques},
      year={2025},
      eprint={2505.13400},
      archivePrefix={arXiv},
      primaryClass={cs.AI},
      url={https://arxiv.org/abs/2505.13400}, 
}

@article {Schretter2025,
article_type = {journal},
title = {Cell type-specific contributions to a persistent aggressive internal state in female \textit{Drosophila}},
author = {Chiu, Hui and Robie, Alice A and Branson, Kristin and Vippa, Tanvi and Epstein, Samantha and Rubin, Gerald M and Anderson, David J and Schretter, Catherine E},
editor = {Sen, Sonia Q and VijayRaghavan, K},
volume = 12,
year = 2025,
month = {jul},
pub_date = {2025-07-25},
pages = {RP88598},
citation = {eLife 2025;12:RP88598},
doi = {10.7554/eLife.88598},
url = {https://doi.org/10.7554/eLife.88598},
journal = {eLife},
issn = {2050-084X},
publisher = {eLife Sciences Publications, Ltd},
}

@article{zheng2018complete,
  title={A complete electron microscopy volume of the brain of adult Drosophila melanogaster},
  author={Zheng, Zhihao and Lauritzen, J Scott and Perlman, Eric and Robinson, Camenzind G and Nichols, Matthew and Milkie, Daniel and Torrens, Omar and Price, John and Fisher, Corey B and Sharifi, Nadiya and others},
  journal={Cell},
  volume={174},
  number={3},
  pages={730--743},
  year={2018},
  publisher={Elsevier}
}

@article {10.7554/eLife.80660,
article_type = {journal},
title = {A searchable image resource of \textit{Drosophila} GAL4 driver expression patterns with single neuron resolution},
author = {Meissner, Geoffrey W and Nern, Aljoscha and Dorman, Zachary and DePasquale, Gina M and Forster, Kaitlyn and Gibney, Theresa and Hausenfluck, Joanna H and He, Yisheng and Iyer, Nirmala A and Jeter, Jennifer and Johnson, Lauren and Johnston, Rebecca M and Lee, Kelley and Melton, Brian and Yarbrough, Brianna and Zugates, Christopher T and Clements, Jody and Goina, Cristian and Otsuna, Hideo and Rokicki, Konrad and Svirskas, Robert R and Aso, Yoshinori and Card, Gwyneth M and Dickson, Barry J and Ehrhardt, Erica and Goldammer, Jens and Ito, Masayoshi and Kainmueller, Dagmar and Korff, Wyatt and Mais, Lisa and Minegishi, Ryo and Namiki, Shigehiro and Rubin, Gerald M and Sterne, Gabriella R and Wolff, Tanya and Malkesman, Oz and FlyLight Project Team},
editor = {Grunwald Kadow, Ilona C and Desplan, Claude},
volume = 12,
year = 2023,
month = {feb},
pub_date = {2023-02-23},
pages = {e80660},
citation = {eLife 2023;12:e80660},
doi = {10.7554/eLife.80660},
url = {https://doi.org/10.7554/eLife.80660},
journal = {eLife},
issn = {2050-084X},
publisher = {eLife Sciences Publications, Ltd},
}

@article{winnubst2019reconstruction,
  title={Reconstruction of 1,000 projection neurons reveals new cell types and organization of long-range connectivity in the mouse brain},
  author={Winnubst, Johan and Bas, Erhan and Ferreira, Tiago A and Wu, Zhuhao and Economo, Michael N and Edson, Patrick and Arthur, Ben J and Bruns, Christopher and Rokicki, Konrad and Schauder, David and others},
  journal={Cell},
  volume={179},
  number={1},
  pages={268--281},
  year={2019},
  publisher={Elsevier}
}

@article{pachitariu2016suite2p,
  title={Suite2p: beyond 10,000 neurons with standard two-photon microscopy},
  author={Pachitariu, Marius and Stringer, Carsen and Schr{\"o}der, Sylvia and Dipoppa, Mario and Rossi, L Federico and Carandini, Matteo and Harris, Kenneth D},
  journal={BioRxiv},
  pages={061507},
  year={2016},
  publisher={Cold Spring Harbor Laboratory}
}

@article {Pachitariu2023.01.07.523036,
	author = {Pachitariu, Marius and Sridhar, Shashwat and Stringer, Carsen},
	title = {Solving the spike sorting problem with Kilosort},
	elocation-id = {2023.01.07.523036},
	year = {2023},
	doi = {10.1101/2023.01.07.523036},
	publisher = {Cold Spring Harbor Laboratory},
	URL = {https://www.biorxiv.org/content/early/2023/01/07/2023.01.07.523036},
	eprint = {https://www.biorxiv.org/content/early/2023/01/07/2023.01.07.523036.full.pdf},
	journal = {bioRxiv}
}

@article{international2025brain,
  title={A brain-wide map of neural activity during complex behaviour},
  author={Angelaki, Dora and Benson, Brandon and Benson, Julius and Birman, Daniel and Bonacchi, Niccol{\`o} and Bougrova, Kc{\'e}nia and Bruijns, Sebastian A and Carandini, Matteo and Catarino, Joana A and others},
  journal={Nature},
  volume={645},
  number={8079},
  pages={177--191},
  year={2025},
  publisher={Nature Publishing Group UK London}
}

@misc{biodiscoveryagent,
      title={BioDiscoveryAgent: An AI Agent for Designing Genetic Perturbation Experiments}, 
      author={Yusuf Roohani and Andrew Lee and Qian Huang and Jian Vora and Zachary Steinhart and Kexin Huang and Alexander Marson and Percy Liang and Jure Leskovec},
      year={2025},
      eprint={2405.17631},
      archivePrefix={arXiv},
      primaryClass={cs.AI},
      url={https://arxiv.org/abs/2405.17631}, 
}

@article{crispragent,
  title={CRISPR-GPT for agentic automation of gene-editing experiments},
  author={Qu, Yuanhao and Huang, Kaixuan and Yin, Ming and Zhan, Kanghong and Liu, Dyllan and Yin, Di and Cousins, Henry C and Johnson, William A and Wang, Xiaotong and Shah, Mihir and others},
  journal={Nature Biomedical Engineering},
  pages={1--14},
  year={2025},
  publisher={Nature Publishing Group UK London}
}

@misc{chemcrow,
      title={ChemCrow: Augmenting large-language models with chemistry tools}, 
      author={Andres M Bran and Sam Cox and Oliver Schilter and Carlo Baldassari and Andrew D White and Philippe Schwaller},
      year={2023},
      eprint={2304.05376},
      archivePrefix={arXiv},
      primaryClass={physics.chem-ph},
      url={https://arxiv.org/abs/2304.05376}, 
}

@article{biomni,
  title={Biomni: A general-purpose biomedical ai agent},
  author={Huang, Kexin and Zhang, Serena and Wang, Hanchen and Qu, Yuanhao and Lu, Yingzhou and Roohani, Yusuf and Li, Ryan and Qiu, Lin and Li, Gavin and Zhang, Junze and others},
  journal={biorxiv},
  year={2025}
}

@misc{googleagent,
      title={An AI system to help scientists write expert-level empirical software}, 
      author={Eser Aygün and Anastasiya Belyaeva and Gheorghe Comanici and Marc Coram and Hao Cui and Jake Garrison and Renee Johnston Anton Kast and Cory Y. McLean and Peter Norgaard and Zahra Shamsi and David Smalling and James Thompson and Subhashini Venugopalan and Brian P. Williams and Chujun He and Sarah Martinson and Martyna Plomecka and Lai Wei and Yuchen Zhou and Qian-Ze Zhu and Matthew Abraham and Erica Brand and Anna Bulanova and Jeffrey A. Cardille and Chris Co and Scott Ellsworth and Grace Joseph and Malcolm Kane and Ryan Krueger and Johan Kartiwa and Dan Liebling and Jan-Matthis Lueckmann and Paul Raccuglia and Xuefei and Wang and Katherine Chou and James Manyika and Yossi Matias and John C. Platt and Lizzie Dorfman and Shibl Mourad and Michael P. Brenner},
      year={2025},
      eprint={2509.06503},
      archivePrefix={arXiv},
      primaryClass={cs.AI},
      url={https://arxiv.org/abs/2509.06503}, 
}

@misc{tusoai,
      title={TusoAI: Agentic Optimization for Scientific Methods}, 
      author={Alistair Turcan and Kexin Huang and Lei Li and Martin Jinye Zhang},
      year={2025},
      eprint={2509.23986},
      archivePrefix={arXiv},
      primaryClass={cs.AI},
      url={https://arxiv.org/abs/2509.23986}, 
}

@InProceedings{simpleagents,
    author    = {Wang, Xuefei and Horstmann, Kai A. and Lin, Ethan and Chen, Jonathan and Farhang, Alexander R. and Stiles, Sophia and Sehgal, Atharva and Light, Jonathan and Van Valen, David and Yue, Yisong and Sun, Jennifer J.},
    title     = {Simple Agents Outperform Experts in Biomedical Imaging Workflow Optimization},
    booktitle = {Proceedings of the IEEE/CVF Conference on Computer Vision and Pattern Recognition (CVPR)},
    month     = {June},
    year      = {2026},
    pages     = {13680-13690}
}

@misc{kosmos,
      title={Kosmos: An AI Scientist for Autonomous Discovery}, 
      author={Ludovico Mitchener and Angela Yiu and Benjamin Chang and Mathieu Bourdenx and Tyler Nadolski and Arvis Sulovari and Eric C. Landsness and Daniel L. Barabasi and Siddharth Narayanan and Nicky Evans and Shriya Reddy and Martha Foiani and Aizad Kamal and Leah P. Shriver and Fang Cao and Asmamaw T. Wassie and Jon M. Laurent and Edwin Melville-Green and Mayk Caldas and Albert Bou and Kaleigh F. Roberts and Sladjana Zagorac and Timothy C. Orr and Miranda E. Orr and Kevin J. Zwezdaryk and Ali E. Ghareeb and Laurie McCoy and Bruna Gomes and Euan A. Ashley and Karen E. Duff and Tonio Buonassisi and Tom Rainforth and Randall J. Bateman and Michael Skarlinski and Samuel G. Rodriques and Michaela M. Hinks and Andrew D. White},
      year={2025},
      eprint={2511.02824},
      archivePrefix={arXiv},
      primaryClass={cs.AI},
      url={https://arxiv.org/abs/2511.02824}, 
}

@misc{discoverybench,
      title={DiscoveryBench: Towards Data-Driven Discovery with Large Language Models}, 
      author={Bodhisattwa Prasad Majumder and Harshit Surana and Dhruv Agarwal and Bhavana Dalvi Mishra and Abhijeetsingh Meena and Aryan Prakhar and Tirth Vora and Tushar Khot and Ashish Sabharwal and Peter Clark},
      year={2024},
      eprint={2407.01725},
      archivePrefix={arXiv},
      primaryClass={cs.CL},
      url={https://arxiv.org/abs/2407.01725}, 
}

@misc{blade,
      title={BLADE: Benchmarking Language Model Agents for Data-Driven Science}, 
      author={Ken Gu and Ruoxi Shang and Ruien Jiang and Keying Kuang and Richard-John Lin and Donghe Lyu and Yue Mao and Youran Pan and Teng Wu and Jiaqian Yu and Yikun Zhang and Tianmai M. Zhang and Lanyi Zhu and Mike A. Merrill and Jeffrey Heer and Tim Althoff},
      year={2025},
      eprint={2408.09667},
      archivePrefix={arXiv},
      primaryClass={cs.CL},
      url={https://arxiv.org/abs/2408.09667}, 
}

@misc{FlyDisco2024,
	title = {The {Fly} {Disco}: {Hardware} and software for optogenetics and fine-grained fly behavior analysis},
	copyright = {© 2024, Posted by Cold Spring Harbor Laboratory. This pre-print is available under a Creative Commons License (Attribution 4.0 International), CC BY 4.0, as described at http://creativecommons.org/licenses/by/4.0/},
	shorttitle = {The {Fly} {Disco}},
	url = {https://www.biorxiv.org/content/10.1101/2024.11.04.621948v1},
	doi = {10.1101/2024.11.04.621948},
	language = {en},
	urldate = {2025-08-28},
	publisher = {bioRxiv},
	author = {Robie, Alice A. and Taylor, Adam L. and Schretter, Catherine E. and Kabra, Mayank and Branson, Kristin},
	month = nov,
	year = {2024},
	note = {Pages: 2024.11.04.621948
Section: New Results},
}

@article{MappingNeuralSubstratesrobie2017,
	title = {Mapping the {Neural} {Substrates} of {Behavior}},
	volume = {170},
	issn = {0092-8674, 1097-4172},
	url = {https://www.cell.com/cell/abstract/S0092-8674(17)30716-X},
	doi = {10.1016/j.cell.2017.06.032},
	language = {English},
	number = {2},
	urldate = {2025-08-28},
	journal = {Cell},
	author = {Robie, Alice A. and Hirokawa, Jonathan and Edwards, Austin W. and Umayam, Lowell A. and Lee, Allen and Phillips, Mary L. and Card, Gwyneth M. and Korff, Wyatt and Rubin, Gerald M. and Simpson, Julie H. and Reiser, Michael B. and Branson, Kristin},
	month = jul,
	year = {2017},
	pmid = {28709004},
	note = {Publisher: Elsevier},
	pages = {393--406.e28},
}

@article{JAABAInteractiveMachinekabra2013,
	title = {{JAABA}: interactive machine learning for automatic annotation of animal behavior},
	volume = {10},
	copyright = {2012 Springer Nature Limited},
	issn = {1548-7105},
	shorttitle = {{JAABA}},
	url = {https://www.nature.com/articles/nmeth.2281},
	doi = {10.1038/nmeth.2281},
	language = {en},
	number = {1},
	urldate = {2026-01-29},
	journal = {Nature Methods},
	author = {Kabra, Mayank and Robie, Alice A. and Rivera-Alba, Marta and Branson, Steven and Branson, Kristin},
	month = jan,
	year = {2013},
	note = {Publisher: Nature Publishing Group},
	pages = {64--67},
}

@inproceedings{FlyTracker,
  title     = {Detecting Social Actions of Fruit Flies},
  author    = {Eyjolfsdottir, Erna and Branson, Steve and Burgos-Artizzu, Xavier P. and Hoopfer, Eric D. and Schor, Jeremy and Anderson, David J. and Perona, Pietro},
  booktitle = {Computer Vision -- ECCV 2014},
  editor    = {Fleet, David and Pajdla, Tomas and Schiele, Bernt and Tuytelaars, Tinne},
  series    = {Lecture Notes in Computer Science},
  volume    = {8690},
  pages     = {772--787},
  year      = {2014},
  publisher = {Springer},
  address   = {Cham},
  doi       = {10.1007/978-3-319-10605-2_50},
}

@article{ctrax,
  title   = {High-throughput ethomics in large groups of \textit{Drosophila}},
  author  = {Branson, Kristin and Robie, Alice and Bender, John and Perona, Pietro and Dickinson, Michael H.},
  journal = {Nature Methods},
  volume  = {6},
  number  = {7},
  pages   = {451--457},
  year    = {2009},
  doi     = {10.1038/nmeth.1328},
}

@misc{mota,
      title={MOT16: A Benchmark for Multi-Object Tracking}, 
      author={Anton Milan and Laura Leal-Taixe and Ian Reid and Stefan Roth and Konrad Schindler},
      year={2016},
      eprint={1603.00831},
      archivePrefix={arXiv},
      primaryClass={cs.CV},
      url={https://arxiv.org/abs/1603.00831}, 
}

@techreport{AllenInstitute2021VisualBehavior2P,
  author      = {{Allen Institute}},
  title       = {{Allen Brain Observatory: Visual Behavior 2P Technical Whitepaper}},
  institution = {Allen Institute for Brain Science},
  year        = {2021},
  month       = {March},
  note        = {V1.0},
  url         = {https://brainmapportal-live-4cc80a57cd6e400d854-f7fdcae.divio-media.net/filer_public/4e/be/4ebe2911-bd38-4230-86c8-01a86cfd758e/visual_behavior_2p_technical_whitepaper.pdf}
}

@misc{IBL2022BrainwideMapWhitepaper,
  author       = {{The International Brain Laboratory}},
  title        = {{Data Release -- Brainwide Map -- Q4 2022}},
  howpublished = {figshare preprint},
  year         = {2022},
  note         = {Version 7, updated 2024-09-19},
  url          = {https://figshare.com/articles/preprint/Data_release_-_Brainwide_map_-_Q4_2022/21400815}
}

@article{mathis2018deeplabcut,
  title={DeepLabCut: markerless pose estimation of user-defined body parts with deep learning},
  author={Mathis, Alexander and Mamidanna, Pranav and Cury, Kevin M and Abe, Taiga and Murthy, Venkatesh N and Mathis, Mackenzie Weygandt and Bethge, Matthias},
  journal={Nature neuroscience},
  volume={21},
  number={9},
  pages={1281--1289},
  year={2018},
  publisher={Nature Publishing Group US New York}
}

@inproceedings{JudgeBench2025,
title={JudgeBench: A Benchmark for Evaluating {LLM}-Based Judges},
author={Sijun Tan and Siyuan Zhuang and Kyle Montgomery and William Yuan Tang and Alejandro Cuadron and Chenguang Wang and Raluca Popa and Ion Stoica},
booktitle={The Thirteenth International Conference on Learning Representations},
year={2025},
url={https://openreview.net/forum?id=G0dksFayVq}
}

@article{rubin2025networks,
  title={Networks of sexually dimorphic neurons that regulate social behaviors in Drosophila},
  author={Rubin, Gerald M and Managan, Claire and Dreher, Marisa and Kim, Elizabeth and Miller, Scott and Boone, Kaitlyn and Robie, Alice A and Taylor, Adam L and Branson, Kristin and Schretter, Catherine E and others},
  journal={bioRxiv},
  pages={2025--10},
  year={2025},
  publisher={Cold Spring Harbor Laboratory}
}

@misc{amodei2024machines,
  author = {Amodei, Dario},
  title = {Machines of Loving Grace: How {AI} Could Transform the World for the Better},
  year = {2024},
  howpublished = {\url{https://darioamodei.com/essay/machines-of-loving-grace}},
  note = {Accessed: 2026}
}

@misc{gridach2025agenticaiscientificdiscovery,
      title={Agentic AI for Scientific Discovery: A Survey of Progress, Challenges, and Future Directions}, 
      author={Mourad Gridach and Jay Nanavati and Khaldoun Zine El Abidine and Lenon Mendes and Christina Mack},
      year={2025},
      eprint={2503.08979},
      archivePrefix={arXiv},
      primaryClass={cs.CL},
      url={https://arxiv.org/abs/2503.08979}, 
}

@misc{APTAnimalPartkabra2022,
  title = {{{APT}}: {{Animal Part Tracker}} v0.3.4},
  shorttitle = {{{APT}}},
  author = {Kabra, Mayank and Lee, Allen and Robie, Alice and Egnor, Roian and Huston, Stephen and Rodriguez, Ivan Felipe and Edwards, Austin and Branson, Kristin},
  year = 2022,
  month = mar,
  doi = {10.5281/zenodo.6366082},
  urldate = {2026-05-29},
  howpublished = {Zenodo},
  langid = {english}
}

@inproceedings{JudgingTheJudges,
    title = "Judging the Judges: Evaluating Alignment and Vulnerabilities in {LLM}s-as-Judges",
    author = "Thakur, Aman Singh  and
      Choudhary, Kartik  and
      Ramayapally, Venkat Srinik  and
      Vaidyanathan, Sankaran  and
      Hupkes, Dieuwke",
    editor = "Arviv, Ofir  and
      Clinciu, Miruna  and
      Dhole, Kaustubh  and
      Dror, Rotem  and
      Gehrmann, Sebastian  and
      Habba, Eliya  and
      Itzhak, Itay  and
      Mille, Simon  and
      Perlitz, Yotam  and
      Santus, Enrico  and
      Sedoc, Jo{\~a}o  and
      Shmueli Scheuer, Michal  and
      Stanovsky, Gabriel  and
      Tafjord, Oyvind",
    booktitle = "Proceedings of the Fourth Workshop on Generation, Evaluation and Metrics (GEM{\texttwosuperior})",
    month = jul,
    year = "2025",
    address = "Vienna, Austria and virtual meeting",
    publisher = "Association for Computational Linguistics",
    url = "https://aclanthology.org/2025.gem-1.33/",
    pages = "404--430",
    ISBN = "979-8-89176-261-9"
}

@misc{NeurodataWithoutBoredom,
      title={Neurodata Without Boredom: Benchmarking Agentic AI for Data Reuse}, 
      author={Ling-Qi Zhang and Kristin Branson},
      year={2026},
      eprint={2605.12808},
      archivePrefix={arXiv},
      primaryClass={cs.LG},
      url={https://arxiv.org/abs/2605.12808}, 
}
\bibliographystyle{colm2026_conference}

\paragraph{LLM usage statement.} We used LLMs to assist with iterating on figure design and editing prose for clarity and concision. All scientific content, analyses, and conclusions are the authors' own.

\appendix
\raggedbottom

\newpage 

\section*{Appendix Contents}
\begin{itemize}
    \item \autoref{sec:appendix_task_design}: Details on task design
    \item \autoref{sec:appendix_eval}: Task-specific evaluation criteria
    \item \autoref{sec:sensitvity_analysis}: Tolerance sensitivity analysis
    \item \autoref{sec:appendix_reference_sources}: Scientist reference solutions
    \item \autoref{sec:appendix_all_tasks_table}: Task inputs and abilities evaluated
    \item \autoref{sec:appendix_iteration}: Iteration behavior
    \item \autoref{sec:appendix_examples}: Example agent outputs
    \item \autoref{sec:appendix_usage}: Token usage and runtimes
    \item \autoref{sec:appendix_task_size}: Measuring task size
    \item \autoref{sec:prompts}: Prompts
\end{itemize}

\section{Details on task design}
\label{sec:appendix_task_design}

To ensure consistent prompt composition across tasks, each prompt contains the following components:
\begin{itemize}[noitemsep,leftmargin=*]
\item Time limit for task completion.
\item Data specification: Location and structure of input files and required task outputs (data files, code, and \texttt{requirements.txt}).
\item Task description: Main objective of task.
\item Domain context: Any domain-specific context or terms required to succeed on the task. Does not include general domain knowledge.
\item Success criteria: Description of how agent solution will be evaluated; e.g.~what metrics will be used.
\item Implementation notes: Task-specific notes critical for success, such as different indexing conventions between different data types.
\end{itemize}
Tasks are provided with data corresponding to a set of 3 out of a total 44 videos or 3 experimental conditions out of a total 11, depending on the task. They are evaluated on the full set of data. 

For all tasks, evaluation criteria include the following:
\begin{itemize}[noitemsep,leftmargin=*]
\item Output existence checks: Checks if all required outputs are present.
\item Evaluation on test data: The agent's solution script is run on a large set of evaluation data, typically consisting of 44 experimental inputs, of which the agent was given a subset of 3 reference inputs for development.
\item Validity checks: Checks that all required output data are created by running the agent's code on the test data.
\item Task-specific evaluation: rigorous checks that quantitatively assess the correctness of the agent's solution.
\end{itemize}

Below, we describe task-specific design considerations and iterations for each pipeline stage, which motivated our general task design principles (\cref{sec:task_design}).

\subsection{Body Tracking}\label{sec:appendix_body_tracking}
The input videos use a custom format (UFMF), requiring helper code to interface with them. Initial prompts described all helper files provided to the agent, when only the top-level movie interface needed to be accessed directly; the rest were its dependencies. This led agents to spend significant effort parsing the details of the UFMF format rather than focusing on the core tracking task. We updated the prompt to remove explanations of these auxiliary files, which prevented agents from getting distracted by implementation details irrelevant to the task objective.

\subsection{Registration}\label{sec:appendix_registration}
The registration task produces newly created trajectories that are split from existing ones when tracking gaps occur. The original prompt required these trajectories to be ordered temporally in a specific way---a constraint that served no scientific purpose but simplified evaluation by enabling direct comparison against ground truth. This is not something a scientist would need, and unnecessarily constrained the task definition. We removed this restriction from the prompt and instead made evaluation robust by using Hungarian matching to align agent trajectories with ground truth before comparison. Separately, we found that agents were consistently failing the evaluation on specific experiments; closer visual inspection revealed that the ground truth registration itself was incorrect due to flipped coordinates.

\subsection{Keypoint Tracking}\label{sec:appendix_keypoint}
The original prompt did not specify success or evaluation criteria. Agent solutions tended to be poor and did not meet the accuracy required for downstream leg movement analysis. We updated the prompt to include the evaluation metric---Euclidean percentile error---and a target threshold which corresponded to the evaluation cutoff for passing. After this change, we observed agents referencing the target in their reasoning and checking their own performance against it during development.

\subsection{Behavior Feature Computation}\label{sec:appendix_perframe}
This task required significant iteration on both prompts and evaluation. Initial prompts were underspecified in subtle ways: for example, velocity features produce $n-1$ frames of output, but without explicit specification, agents applied different padding conventions. We added concrete details about expected output array sizes to resolve such ambiguities. The iteration process also uncovered a bug in the ground truth code where angle flips were computed incorrectly, as well as errors in the scientist's feature definitions that had survived multiple rounds of peer review. On the evaluation side, we found that agents were consistently failing on certain features (\texttt{corfrac\_maj}, \texttt{corfrac\_min}); investigation revealed that our evaluation criteria were too strict rather than the agent solutions being incorrect. Similarly, failures on \texttt{absdv\_cor} stemmed from its dependency on the \texttt{corfrac} features. We created separate test cases for these features with tailored tolerances to evaluate them independently. Two independent reviewers checking the prompt and evaluation proved valuable for catching such issues.

\subsection{Gait Segmentation}\label{sec:appendix_gait}
When designing the task in collaboration with a domain expert, early prompts referenced the scientist's specific algorithmic approach of per-leg thresholding with hysteresis. This had the undesired effect of constraining agents to heuristic methods rather than exploring alternatives. After removing algorithmic suggestions, agent solutions became more varied, including learning-based approaches.

\subsection{Statistical Comparisons}\label{sec:appendix_stat_comp}
Initial prompts lacked information about what constitutes a sample in the Mann-Whitney U-test, leaving agents to make varying and sometimes valid choices about how to pool data. We updated the prompt to specify how samples are constructed for evaluation. On the evaluation side, we introduced tolerance around $p$-value boundaries to accommodate minor numerical differences that do not affect statistical conclusions. We also discovered that the ground truth made an implicit assumption about filtering bouts not contained in the OFF/ON stimulation period---an assumption agents tended not to make. We updated the ground truth to remove this filtering, ensuring agents were not penalized for omitting an unstated step.

\section{Task-specific evaluation criteria}\label{sec:appendix_eval}

For each task, we define a per-experiment \emph{ratio} $r(e) \geq 0$ where $r(e) \geq 1$ indicates the agent passes experiment $e$. We grade each trial using the distribution of ratios across experiments:
\begin{itemize}[nosep]
    \item \textbf{Green} (\checkmark): $P_5(r) \geq 1$ --- at least 95\% of experiments pass.
    \item \textbf{Yellow} ($\sim$): $\mathbb{E}[r] \geq 1$ but $P_5(r) < 1$ --- passes on average but not consistently.
    \item \textbf{Red} ($\times$): $\mathbb{E}[r] < 1$ --- fails on average.
\end{itemize}
where $P_5$ denotes the 5th percentile. For multi-criteria tasks (Registration, Behavior Feature Computation), the per-experiment ratio is the minimum across all criteria: $r(e) = \min_c r_c(e)$. The task \emph{score} reported in \cref{table:results} is $\mathbb{E}_{t,e}[r]$, averaged over trials $t$ and experiments $e$. For multi-criteria tasks, the score uses a mean-of-means: each criterion is averaged across experiments, then averaged across criteria (capped at 2).

Sources of scientist reference solutions and ground-truth labels for each task, where applicable, are detailed in \cref{sec:appendix_reference_sources}.

\subsection{Body Tracking}
For a given experimental video $e$, the agent passes if
\begin{equation}
    \text{MOTA}_{\text{agent}}(e) \geq \text{MOTA}_{\text{baseline}}(e).
\end{equation}
Let $v(e) = 1$ if the agent's output trajectory file is well-formed (required fields are present with the expected shapes and the per-target frame counts, offsets, and time deltas are internally consistent), and $v(e) = 0$ otherwise. The per-experiment ratio and task score are:
\begin{equation}
    r(e) = v(e) \cdot \frac{1 - \text{MOTA}_{\text{baseline}}(e)}{1 - \text{MOTA}_{\text{agent}}(e)}, \qquad \text{score} = \mathbb{E}_{t,e}[r].
\end{equation}

\subsection{Registration}
Registration is evaluated on multiple criteria per experiment. For each criterion $c$, the ratio is $r_c(e) = \tau_c / \epsilon_c(e)$ where $\tau_c$ is the tolerance and $\epsilon_c(e)$ is the agent's error. The criteria are:

\begin{itemize}[nosep]
    \item \textbf{Scale consistency}: $r = (\delta_{\text{scale}} \cdot \hat{s}) \,/\, \text{range}(\hat{s})$, where $\hat{s}$ is the estimated pixels-per-mm and $\delta_{\text{scale}} = 10^{-5}$.
    \item \textbf{Radius error}: $r = \tau_{\text{radius}} \,/\, |\hat{R} - R_{\text{GT}}|$, with $\tau_{\text{radius}} = 2.5$ px.
    \item \textbf{Center consistency} (x and y): $r = \delta_{\text{center}} \,/\, \text{range}_{\text{rel}}(\hat{c})$, with $\delta_{\text{center}} = 0.01$.
    \item \textbf{Center error}: $r = \tau_{\text{center}} \,/\, \|\hat{c} - c_{\text{GT}}\|$, with $\tau_{\text{center}} = 5.0$ px.
    \item \textbf{Max/mean residuals} (per coordinate field $f \in \{x, y, a, b, \theta\}$): $r = \tau_f \,/\, \text{residual}_f$.
\end{itemize}

Let $v(e) = 1$ if the registered output is well-formed (required fields present, no NaN values, array lengths consistent) and its millimeter and pixel coordinate representations agree under the agent's reported pixels-per-mm scale, and $v(e) = 0$ otherwise. The agent passes experiment $e$ if $v(e) \cdot \min_c r_c(e) \geq 1$. The task score uses a mean-of-means:
\begin{equation}
    \text{score} = \mathbb{E}_{t}\!\left[\frac{1}{|C|}\sum_{c} \mathbb{E}_{e}\!\left[v(e) \cdot \min(2,\, r_c(e))\right]\right].
\end{equation}

\subsection{Keypoint Tracking}
The agent trains a keypoint detection model evaluated on held-out test data. The agent passes if its mean error percentile is within a tolerance of the baseline:
\begin{equation}
    \bar{\epsilon}_{\text{agent}} \leq \alpha \cdot \bar{\epsilon}_{\text{baseline}},
\end{equation}
where $\alpha = 1.5$ is the pass fraction. The ratio and score are:
\begin{equation}
    r = \frac{\alpha \cdot \bar{\epsilon}_{\text{baseline}}}{\bar{\epsilon}_{\text{agent}}}, \qquad \text{score} = \mathbb{E}_{t}[r].
\end{equation}

\subsection{Behavior Feature Computation}
Per-frame features are compared against a baseline. For each feature $f$ and experiment $e$, the ratio depends on the feature type:

\begin{itemize}[nosep]
    \item \textbf{Standard features} (13 features): $r_f(e) = \tau_f(e) \,/\, \text{MAE}_f(e)$, where $\tau_f(e) = \delta_z \cdot \sigma_f(e)$ with $\delta_z = 0.01$.
    \item \textbf{Absolute consistency} (\texttt{absdv\_cor}): $r = 1$ if $|\texttt{absdv\_cor}| = |\texttt{dv\_cor}|$ (within floating-point tolerance), else $r = 0$.
    \item \textbf{Body-coordinate fraction} (\texttt{corfrac\_maj}, \texttt{corfrac\_min}, evaluated jointly): $r = \varepsilon \,/\, \left(\max_{i,t} \max\!\left(0,\, L_{\text{agent}}(i, t) - L_{\text{baseline}}(i, t)\right)\right)$ with $\varepsilon = 10^{-3}$, where $L(i, t)$ is the squared error of the body-translation reconstruction at frame $t$ of fly $i$ using the predicted \texttt{corfrac} values and the tracked body axis.
\end{itemize}

The agent passes experiment $e$ if all features pass: $\min_f r_f(e) \geq 1$. The task score uses a mean-of-means:
\begin{equation}
    \text{score} = \mathbb{E}_{t}\!\left[\frac{1}{|F|}\sum_{f} \mathbb{E}_{e}\!\left[\min(2,\, r_f(e))\right]\right].
\end{equation}

\subsection{Behavior Classifier}
The agent trains a walking behavior classifier evaluated on held-out test data. The agent passes if its balanced accuracy meets a threshold derived from the baseline:
\begin{equation}
    \text{BA}_{\text{agent}} \geq \alpha \cdot \text{BA}_{\text{baseline}},
\end{equation}
where $\alpha = 0.95$. The ratio and score are:
\begin{equation}
    r = \frac{1 - \alpha \cdot \text{BA}_{\text{baseline}}}{1 - \text{BA}_{\text{agent}}}, \qquad \text{score} = \mathbb{E}_{t}[r].
\end{equation}

\subsection{Gait Segmentation}
For each experimental video $e$, the agent's per-frame swing/stance predictions are scored against the scientist's reference algorithm output using bout-level balanced accuracy. The baseline accuracy is the same metric applied to the scientist's predictions against manually annotated ground-truth labels, with:
\begin{equation}
    \text{BA}_{\text{agent}}(e) = \text{bout-BA}\!\left(\hat{y}_{\text{agent}}(e),\; \hat{y}_{\text{scientist}}(e)\right),
\end{equation} and
\begin{equation}
    \text{BA}_{\text{baseline}}(e) = \text{bout-BA}\!\left(\hat{y}_{\text{scientist}}(e),\; y^*(e)\right),
\end{equation}
where $\hat{y}_{\text{agent}}(e), \hat{y}_{\text{scientist}}(e)$ are the predicted swing/stance labels and $y^*(e)$ are the manual labels. Bout-level balanced accuracy averages, over each contiguous label bout (per fly, per leg), the fraction of frames labeled correctly, then takes the unweighted mean across the two classes (swing, stance) to balance class frequencies. The agent passes if:
\begin{equation}
    \text{BA}_{\text{agent}}(e) \geq \alpha \cdot \text{BA}_{\text{baseline}}(e),
\end{equation}
with pass fraction $\alpha = 0.95$. The ratio and score are:
\begin{equation}
    r(e) = \frac{1 - \alpha \cdot \text{BA}_{\text{baseline}}(e)}{1 - \text{BA}_{\text{agent}}(e)}, \qquad \text{score} = \mathbb{E}_{t,e}[r].
\end{equation}

\subsection{Statistical Comparisons}
For each genetic line $\ell$, the agent computes $p$-values for whether a treatment affects walking speed in slow and fast phases. The agent's significance calls $s_{\ell,\phi,\alpha}$ are compared against the ground truth $s^*_{\ell,\phi,\alpha}$ for $\phi \in \{\text{slow}, \text{fast}\}$ and $\alpha \in \{0.05, 0.01, 0.001\}$. To avoid penalizing borderline cases, a call is judged \emph{correct} if either the significance flag matches or the submitted $p$-value is within 10\% of ground truth:
\begin{equation}
    c_{\ell,\phi,\alpha} = \mathbf{1}\!\left[s_{\ell,\phi,\alpha} = s^*_{\ell,\phi,\alpha}\right] \;\lor\; \mathbf{1}\!\left[\max\!\left(\tfrac{p_{\ell,\phi}}{p^*_{\ell,\phi}},\, \tfrac{p^*_{\ell,\phi}}{p_{\ell,\phi}}\right) \leq 1 + \delta_p\right],
\end{equation}
with tolerance $\delta_p = 0.1$. The per-line ratio is
\begin{equation}
    r(\ell) = \mathbf{1}\!\left[\textstyle\bigwedge_{\phi,\alpha}\, c_{\ell,\phi,\alpha}\right],
\end{equation}
i.e., $r(\ell) = 1$ iff all phases and all $\alpha$ levels are correct for line $\ell$, and $r(\ell) = 0$ otherwise. Grading uses a custom scheme:
\begin{itemize}[nosep]
    \item \textbf{Green}: All lines correct for all phases and all $\alpha$ levels ($\min_\ell r(\ell) = 1$).
    \item \textbf{Yellow}: All lines correct for both phases at $\alpha = 0.05$.
    \item \textbf{Red}: Otherwise.
\end{itemize}
The task score is the fraction of lines correct, averaged over trials:
\begin{equation}
    \text{score} = \mathbb{E}_{t,\ell}[r(\ell)].
\end{equation}

\section{Tolerance sensitivity analysis}
\label{sec:sensitvity_analysis}
\begin{table}[H]
\centering
\small
\setlength{\tabcolsep}{2.2pt}
\renewcommand{\arraystretch}{1}
\newcommand{\sensna}{\textcolor{gray}{--}}
\newcommand{\sensscore}[1]{\\[-1pt]\rule[-3pt]{0pt}{11pt}\scriptsize #1}
\newcommand{\sensrow}[1]{\scriptsize $#1\times$}
\newcommand{\sensagent}[2]{\makecell[l]{\texttt{#1/}\\\quad\texttt{\small #2}}}
\newcommand{\senscell}[2]{\makecell{#1\sensscore{#2}}}
\newcommand{\sensnacell}{\makecell{\sensna}}

\resizebox{\textwidth}{!}{%
\begin{tabular}{m{3.1cm}c ccccccc}
\toprule
\textbf{Agent} & \textbf{$\epsilon$-Scale} & \makecell{\textbf{Body}\\\textbf{Tracking*}} & \makecell{\textbf{Registration}} & \makecell{\textbf{Keypoint}\\\textbf{Tracking}} & \makecell{\textbf{Feature}\\\textbf{Computation}} & \makecell{\textbf{Behavior}\\\textbf{Classifier}} & \makecell{\textbf{Gait}\\\textbf{Segmentation}} & \makecell{\textbf{Statistical}\\\textbf{Comparison}} \\
\midrule
\rowcolor{gray!12}\multicolumn{9}{l}{\textbf{Single-stage tasks}} \\
\addlinespace[2pt]
\sensagent{claude-code}{claude-opus-4-6} & \sensrow{0.5} & \senscell{\mrred\ \mrred\ \mrred}{0.000$\pm$0.000} & \senscell{\mrred\ \mrgreen\ \mrgreen}{1.917$\pm$0.137} & \senscell{\mrgreen\ \mrgreen\ \mrgreen}{1.101$\pm$0.044} & \senscell{\mrred\ \mrred\ \mrred}{1.761$\pm$0.000} & \senscell{\mrgreen\ \mrgreen\ \mrgreen}{1.962$\pm$0.093} & \senscell{\mrred\ \mrred\ \mrred}{0.352$\pm$0.302} & \senscell{\mrgreen\ \mrgreen\ \mrgreen}{1.000$\pm$0.000} \\
 & \sensrow{1} & \senscell{\mrred\ \mrred\ \mrred}{0.017$\pm$0.003} & \senscell{\mrred\ \mrgreen\ \mrgreen}{1.928$\pm$0.125} & \senscell{\mrgreen\ \mrgreen\ \mrgreen}{1.376$\pm$0.055} & \senscell{\mrgreen\ \mrgreen\ \mrgreen}{1.855$\pm$0.000} & \senscell{\mrgreen\ \mrgreen\ \mrgreen}{2.281$\pm$0.108} & \senscell{\mrred\ \mrred\ \mrred}{0.451$\pm$0.387} & \senscell{\mrgreen\ \mrgreen\ \mrgreen}{1.000$\pm$0.000} \\
 & \sensrow{2} & \senscell{\mrred\ \mrred\ \mrred}{0.061$\pm$0.009} & \senscell{\mrred\ \mrgreen\ \mrgreen}{1.943$\pm$0.099} & \senscell{\mrgreen\ \mrgreen\ \mrgreen}{2.752$\pm$0.111} & \senscell{\mrgreen\ \mrgreen\ \mrgreen}{1.933$\pm$0.000} & \senscell{\mrgreen\ \mrgreen\ \mrgreen}{2.921$\pm$0.139} & \senscell{\mrred\ \mrred\ \mrgreen}{0.649$\pm$0.556} & \senscell{\mrgreen\ \mrgreen\ \mrgreen}{1.000$\pm$0.000} \\
\addlinespace[2pt]
\sensagent{codex}{gpt-5.4} & \sensrow{0.5} & \senscell{\mrred\ \mrred\ \mrred}{0.000$\pm$0.000} & \senscell{\mrred\ \mrgreen\ \mrgreen}{1.985$\pm$0.021} & \senscell{\mrgreen\ \mrgreen\ \mrgreen}{1.084$\pm$0.229} & \senscell{\mrred\ \mrred\ \mrred}{1.761$\pm$0.000} & \senscell{\mrgreen\ \mrred\ \mrgreen}{1.004$\pm$0.340} & \senscell{\mrred\ \mrred\ \mrred}{0.726$\pm$0.030} & \senscell{\mrred\ \mrred\ \mrgreen}{0.733$\pm$0.379} \\
 & \sensrow{1} & \senscell{\mrred\ \mrred\ \mrred}{0.020$\pm$0.006} & \senscell{\mrgreen\ \mrgreen\ \mrgreen}{1.999$\pm$0.002} & \senscell{\mrgreen\ \mrgreen\ \mrgreen}{1.355$\pm$0.286} & \senscell{\mrgreen\ \mrgreen\ \mrgreen}{1.855$\pm$0.000} & \senscell{\mrgreen\ \mrred\ \mrgreen}{1.167$\pm$0.396} & \senscell{\mrred\ \mrred\ \mrred}{0.929$\pm$0.039} & \senscell{\mrred\ \mrred\ \mrgreen}{0.733$\pm$0.379} \\
 & \sensrow{2} & \senscell{\mrred\ \mrred\ \mrred}{0.068$\pm$0.015} & \senscell{\mrgreen\ \mrgreen\ \mrgreen}{2.000$\pm$0.000} & \senscell{\mrgreen\ \mrgreen\ \mrgreen}{2.711$\pm$0.572} & \senscell{\mrgreen\ \mrgreen\ \mrgreen}{1.933$\pm$0.000} & \senscell{\mrgreen\ \mrred\ \mrgreen}{1.494$\pm$0.507} & \senscell{\mrgreen\ \mrgreen\ \mrgreen}{1.337$\pm$0.055} & \senscell{\mrgreen\ \mrred\ \mrgreen}{0.767$\pm$0.404} \\
\addlinespace[2pt]
\sensagent{terminus-2}{claude-opus-4-6} & \sensrow{0.5} & \senscell{\mrred\ \mrred\ \mrred}{0.000$\pm$0.000} & \senscell{\mrgreen\ \mrgreen\ \mrgreen}{1.997$\pm$0.000} & \senscell{\mrgreen\ \mrgreen\ \mrgreen}{1.097$\pm$0.036} & \senscell{\mrred\ \mrred\ \mrred}{1.756$\pm$0.009} & \senscell{\mrgreen\ \mrgreen\ \mrgreen}{2.884$\pm$0.876} & \senscell{\mrred\ \mrred\ \mrred}{0.938$\pm$0.176} & \senscell{\mrgreen\ \mrgreen\ \mrgreen}{1.000$\pm$0.000} \\
 & \sensrow{1} & \senscell{\mrred\ \mrred\ \mrred}{0.044$\pm$0.022} & \senscell{\mrgreen\ \mrgreen\ \mrgreen}{2.000$\pm$0.000} & \senscell{\mrgreen\ \mrgreen\ \mrgreen}{1.371$\pm$0.046} & \senscell{\mrgreen\ \mrred\ \mrgreen}{1.851$\pm$0.006} & \senscell{\mrgreen\ \mrgreen\ \mrgreen}{3.354$\pm$1.019} & \senscell{\mrgreen\ \mrred\ \mrgreen}{1.201$\pm$0.225} & \senscell{\mrgreen\ \mrgreen\ \mrgreen}{1.000$\pm$0.000} \\
 & \sensrow{2} & \senscell{\mrred\ \mrred\ \mrred}{0.175$\pm$0.095} & \senscell{\mrgreen\ \mrgreen\ \mrgreen}{2.000$\pm$0.000} & \senscell{\mrgreen\ \mrgreen\ \mrgreen}{2.742$\pm$0.091} & \senscell{\mrgreen\ \mrgreen\ \mrgreen}{1.931$\pm$0.004} & \senscell{\mrgreen\ \mrgreen\ \mrgreen}{4.295$\pm$1.304} & \senscell{\mrgreen\ \mrgreen\ \mrgreen}{1.728$\pm$0.324} & \senscell{\mrgreen\ \mrgreen\ \mrgreen}{1.000$\pm$0.000} \\
\addlinespace[2pt]
\sensagent{terminus-2}{gpt-5.4} & \sensrow{0.5} & \senscell{\mrred\ \mrred\ \mrred}{0.000$\pm$0.000} & \senscell{\mrred\ \mrred\ \mrred}{1.864$\pm$0.088} & \senscell{\mrred\ \mrred\ \mrred}{0.371$\pm$0.294} & \senscell{\mrred\ \mrred\ \mrred}{1.598$\pm$0.283} & \senscell{\mrgreen\ \mrgreen\ \mrgreen}{1.084$\pm$0.201} & \senscell{\mrred\ \mrred\ \mrred}{0.533$\pm$0.099} & \senscell{\mrred\ \mrred\ \mrgreen}{0.933$\pm$0.058} \\
 & \sensrow{1} & \senscell{\mrred\ \mrred\ \mrred}{0.005$\pm$0.008} & \senscell{\mrred\ \mrred\ \mrred}{1.892$\pm$0.090} & \senscell{\mrred\ \mrred\ \mrred}{0.464$\pm$0.367} & \senscell{\mrgreen\ \mrgreen\ \mrred}{1.662$\pm$0.334} & \senscell{\mrgreen\ \mrgreen\ \mrgreen}{1.260$\pm$0.233} & \senscell{\mrred\ \mrred\ \mrred}{0.682$\pm$0.127} & \senscell{\mrred\ \mrred\ \mrgreen}{0.933$\pm$0.058} \\
 & \sensrow{2} & \senscell{\mrred\ \mrred\ \mrred}{0.016$\pm$0.028} & \senscell{\mrred\ \mrred\ \mrred}{1.922$\pm$0.086} & \senscell{\mrred\ \mrred\ \mrred}{0.929$\pm$0.734} & \senscell{\mrgreen\ \mrgreen\ \mrred}{1.717$\pm$0.374} & \senscell{\mrgreen\ \mrgreen\ \mrgreen}{1.614$\pm$0.299} & \senscell{\mrred\ \mrgreen\ \mrred}{0.981$\pm$0.183} & \senscell{\mrred\ \mrred\ \mrgreen}{0.933$\pm$0.058} \\
\midrule
\rowcolor{gray!12}\multicolumn{9}{l}{\textbf{E2E Maximal}} \\
\addlinespace[2pt]
\sensagent{claude-code}{claude-opus-4-6} & \sensrow{0.5} & \senscell{\mrred\ \mrred\ \mrred}{0.000$\pm$0.000} & \senscell{\mrred\ \mrred\ \mrred}{1.862$\pm$0.053} & \senscell{\mrred\ \mrred\ \mrgreen}{0.562$\pm$0.512} & \sensnacell & \senscell{\mrred\ \mrgreen\ \mrgreen}{0.979$\pm$0.848} & \sensnacell & \senscell{\mrred\ \mrred\ \mrred}{0.467$\pm$0.058} \\
 & \sensrow{1} & \senscell{\mrred\ \mrred\ \mrred}{0.020$\pm$0.002} & \senscell{\mrred\ \mrred\ \mrred}{1.918$\pm$0.052} & \senscell{\mrred\ \mrred\ \mrgreen}{0.702$\pm$0.640} & \sensnacell & \senscell{\mrred\ \mrgreen\ \mrgreen}{1.139$\pm$0.986} & \sensnacell & \senscell{\mrred\ \mrred\ \mrred}{0.467$\pm$0.058} \\
 & \sensrow{2} & \senscell{\mrred\ \mrred\ \mrred}{0.071$\pm$0.006} & \senscell{\mrgreen\ \mrred\ \mrred}{1.970$\pm$0.028} & \senscell{\mrred\ \mrred\ \mrgreen}{1.405$\pm$1.280} & \sensnacell & \senscell{\mrred\ \mrgreen\ \mrgreen}{1.458$\pm$1.263} & \sensnacell & \senscell{\mrred\ \mrred\ \mrred}{0.467$\pm$0.058} \\
\addlinespace[2pt]
\sensagent{codex}{gpt-5.4} & \sensrow{0.5} & \senscell{\mrred\ \mrred\ \mrred}{0.000$\pm$0.000} & \senscell{\mrred\ \mrred\ \mrred}{1.917$\pm$0.046} & \senscell{\mrred\ \mrred\ \mrred}{0.197$\pm$0.342} & \sensnacell & \senscell{\mrred\ \mrgreen\ \mrred}{0.966$\pm$1.673} & \sensnacell & \senscell{\mrred\ \mrred\ \mrred}{0.300$\pm$0.100} \\
 & \sensrow{1} & \senscell{\mrred\ \mrred\ \mrred}{0.039$\pm$0.038} & \senscell{\mrgreen\ \mrred\ \mrred}{1.948$\pm$0.063} & \senscell{\mrred\ \mrred\ \mrred}{0.247$\pm$0.427} & \sensnacell & \senscell{\mrred\ \mrgreen\ \mrred}{1.123$\pm$1.946} & \sensnacell & \senscell{\mrred\ \mrred\ \mrred}{0.300$\pm$0.100} \\
 & \sensrow{2} & \senscell{\mrred\ \mrred\ \mrred}{0.153$\pm$0.160} & \senscell{\mrgreen\ \mrgreen\ \mrred}{1.961$\pm$0.065} & \senscell{\mrred\ \mrred\ \mrred}{0.493$\pm$0.854} & \sensnacell & \senscell{\mrred\ \mrgreen\ \mrred}{1.438$\pm$2.491} & \sensnacell & \senscell{\mrred\ \mrred\ \mrred}{0.300$\pm$0.100} \\
\midrule
\rowcolor{gray!12}\multicolumn{9}{l}{\textbf{E2E Minimal}} \\
\addlinespace[2pt]
\sensagent{claude-code}{claude-opus-4-6} & \sensrow{0.5} & \sensnacell & \sensnacell & \sensnacell & \sensnacell & \sensnacell & \sensnacell & \senscell{\mrred\ \mrred\ \mrred}{0.467$\pm$0.058} \\
 & \sensrow{1} & \sensnacell & \sensnacell & \sensnacell & \sensnacell & \sensnacell & \sensnacell & \senscell{\mrred\ \mrred\ \mrred}{0.467$\pm$0.058} \\
 & \sensrow{2} & \sensnacell & \sensnacell & \sensnacell & \sensnacell & \sensnacell & \sensnacell & \senscell{\mrred\ \mrred\ \mrred}{0.467$\pm$0.058} \\
\addlinespace[2pt]
\sensagent{codex}{gpt-5.4} & \sensrow{0.5} & \sensnacell & \sensnacell & \sensnacell & \sensnacell & \sensnacell & \sensnacell & \senscell{\mrred\ \mrred\ \mrred}{0.233$\pm$0.153} \\
 & \sensrow{1} & \sensnacell & \sensnacell & \sensnacell & \sensnacell & \sensnacell & \sensnacell & \senscell{\mrred\ \mrred\ \mrred}{0.233$\pm$0.153} \\
 & \sensrow{2} & \sensnacell & \sensnacell & \sensnacell & \sensnacell & \sensnacell & \sensnacell & \senscell{\mrred\ \mrred\ \mrred}{0.233$\pm$0.153} \\
\bottomrule
\end{tabular}}
\caption{Sensitivity of task grades to evaluation-threshold scaling. Each row evaluates the same submissions with task-specific $\epsilon$ scaled by $0.5\times$, $1\times$, or $2\times$. Symbols show the outcomes of the three trials in each setting. Here, \mrgreen\ indicates $P_5(r) \geq 1$, and \mrred\ indicates $P_5(r) < 1$. Across all 96 trials, 92\% retain the same pass/fail classification when the tolerance is relaxed from $\epsilon$ to $2\epsilon$, and 88\% retain the same classification when tightened to $\epsilon/2$. Together, these results indicate that most trials are not near the decision boundary and that the main conclusions are robust to reasonable variation in evaluation thresholds. Numbers report mean task score $\pm$ standard deviation. Dashes indicate stages not evaluated in the corresponding end-to-end setting. *For Body Tracking, where $\epsilon=0$ and multiplicative scaling would leave the threshold unchanged, the $0.5\times$ and $2\times$ settings instead correspond to $\epsilon - 0.05$ and $\epsilon + 0.05$, respectively.}
\label{table:appendix_sensitivity}
\end{table}

\section{Scientist reference solutions}
\label{sec:appendix_reference_sources}
\begin{table}[H]
\centering
\setlength{\tabcolsep}{4pt}
\renewcommand{\arraystretch}{2.00}
\resizebox{\textwidth}{!}{%
\begin{tabular}{L{3.0cm}L{5.4cm}L{6.2cm}}
\toprule
\textbf{Task} & \textbf{Scientist reference solution source} & \textbf{Additional ground-truth labels} \\
\midrule
Body Tracking & Ctrax fly tracker \citep{ctrax} & Annotations initialized with FlyTracker \citep{FlyTracker} and manually corrected by domain scientists. \\
Registration & Fly Disco \citep{FlyDisco2024} & -- \\
Keypoint Tracking & Animal Part Tracker (APT) \citep{APTAnimalPartkabra2022} & MultiFly keypoint dataset \citep{FlyDisco2024} \\
Behavior Feature Computation & \citep{MappingNeuralSubstratesrobie2017} & -- \\
Walking Behavior Classification & JAABA \citep{JAABAInteractiveMachinekabra2013} & Behavior category labels \citep{FlyDisco2024} \\
Gait Segmentation & Fly Disco \citep{FlyDisco2024} & -- \\
Statistical Comparisons & Written by a domain scientist author of this paper. & -- \\
\bottomrule
\end{tabular}}
\caption{Sources of scientist reference solutions and ground-truth labels for each task. As discussed in \cref{sec:evaluation_design}, task evaluations are based on community-accepted scientist baselines derived from scientific publications and, where available, additional ground truth labels. The source for each is detailed here.}
\label{table:appendix_reference_sources_table}
\end{table}

\section{Task inputs and abilities evaluated}
\label{sec:appendix_all_tasks_table}
\begin{table}[H]
\centering
\fontsize{7.45pt}{8.94pt}\selectfont
\setlength{\tabcolsep}{3.5pt}
\renewcommand{\arraystretch}{1.1}
\begin{tabular}{L{0.12\textwidth} L{0.16\textwidth} L{0.19\textwidth} L{0.46\textwidth}}
\toprule
\textbf{Task} & \textbf{Goal} & \textbf{Information provided\textsuperscript{*}} & \textbf{Abilities evaluated} \\
\midrule
Body Tracking & Develop an algorithm to detect and track multiple flies in raw video. & Raw videos, code for reading video. No reference algorithm, no ground-truth labels, no optimizable criterion. & \textit{Image processing, multi-target tracking, evaluation criteria development.} Agents must apply visual knowledge of fly appearance to detect each fly's position and orientation (including when flies touch), and assign consistent identities across frames. As no criterion or labels are given, agents must develop their own evaluation criteria. Their method must generalize to unseen videos. \\
\midrule
Registration & Develop a method to spatially align tracked data to a canonical coordinate system, then clean the data by removing NaNs. & Tracked trajectories from Body Tracking, code for loading trajectories. Arena radius in mm. No labeled examples, no boundary detection algorithm. & \textit{Image processing, coordinate transformation, data cleaning.} Agents must determine how to detect the boundary of the arena, an image processing task, then use the given arena radius to transform the trajectories produced by Body Tracking from pixels to millimeters, centered on the arena center. Although agents are provided data for algorithm development, their algorithms must generalize to unseen data. \\
\midrule
Keypoint Tracking & Train a supervised keypoint detection model from provided training data. & Labeled training images in COCO format. Target accuracy thresholds. No algorithmic details. & \textit{Machine learning, model training, computer vision.} Agents must design and train a machine learning model that can predict keypoint locations from images given labeled training data. This is a common computer vision task, with the marked difference of the high precision required: sub-precision accuracy on average, and small error for outliers, as required for downstream analysis tasks. Agents are provided accuracy goals, and must design an algorithm tuned to data properties and this high precision goal. \\
\midrule
Behavior Feature Computation & Implement a program to compute pre-defined features from trajectories given equations. & Feature definitions as PDF or TeX, code for loading trajectories. & \textit{Implementation, geometry.} Agents must translate equations to code; two features additionally require geometric reasoning to find a described optimum. \\
\midrule
Walking Behavior Classification & Train a walking behavior classifier from manually labeled trajectories. & Sparse manual labels, trajectory and behavior feature files (heterogeneous formats), no algorithmic details. & \textit{Machine learning, action recognition.} Agents must design and train a time-series-based classifier given sparse labels, and build a data loader that correctly aligns indices across heterogeneous files. \\
\midrule
Gait Segmentation & Develop a method to segment walking into swing and stance gait phases from tracked poses. & A few manually labeled swing/stance examples (too few to train a generic classifier), keypoint definitions in entomology terminology. No algorithmic details. & \textit{Domain knowledge, walking gait analysis, parameter tuning.} Agents must apply domain knowledge of gait definitions, interpret entomological keypoint terminology, and tune parameters from the provided examples. \\
\midrule
Statistical Comparisons & Conduct Mann-Whitney U tests to evaluate whether optogenetic perturbation affects stance duration. & Data from multiple sources, three $p$-value thresholds, statistical sample definition, type of statistical test. & \textit{Data integration, statistical testing.} Agents must combine and align data across sources, pool across time and flies to construct pseudo-independent samples, and run significance tests at the specified thresholds. \\
\bottomrule
\end{tabular}
\caption{\small Expanded description of our tasks, distinguishing the information provided to agents from the abilities required to obtain a passing solution. *In all tasks, agents are provided the input and output data format.}
\label{table:expanded_tasks}
\end{table}

\section{Iteration behavior}
\label{sec:appendix_iteration}

\begin{figure}[H]
    \centering
    \includegraphics[width=0.75\linewidth]{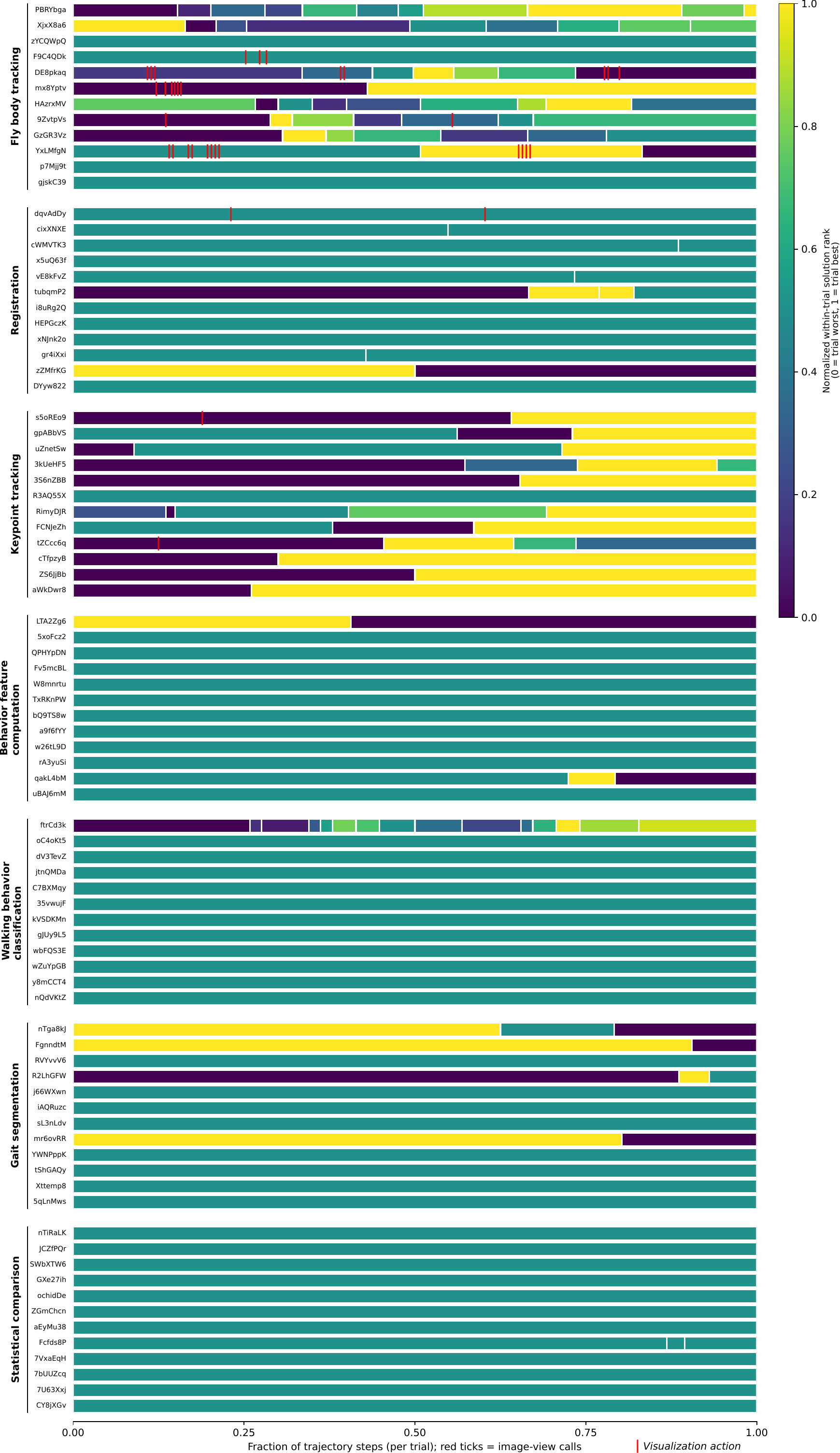}
    \caption{\small Per-trial iteration trajectories for all tasks and their respective trials. To obtain the segmentation of a trajectory into individual iterations, each step of each trajectory was classified with the aid of an LLM judge to identify the steps in which qualitative changes to agent solutions occurred and were subsequently executed by the agents. A snapshot corresponding to each iteration was then programmatically extracted and scored according to the respective task's evaluation suite. Segment color encodes its within-trial rank ranging from dark/worst (0.0) to bright/best (1.0). Red ticks denote steps where agents executed a tool call to visualize an image. Trials are sorted by final-iteration value per task in descending order. Each bar is one agent trial, split into segments for that trial's iterations in chronological order.}
    \label{fig:appendix_all-trajectories}
\end{figure}

\section{Example agent outputs}
\label{sec:appendix_examples}

\newcommand{\tmark}[1]{\textsuperscript{\textcolor{gray}{#1}}}

\subsection{Failure modes}

\subsubsection{Tracking errors without visualization (\texttt{claude-code/claude-opus-4-6})}

\begin{figure}[H]
    \centering
\includegraphics[width=0.95\linewidth]{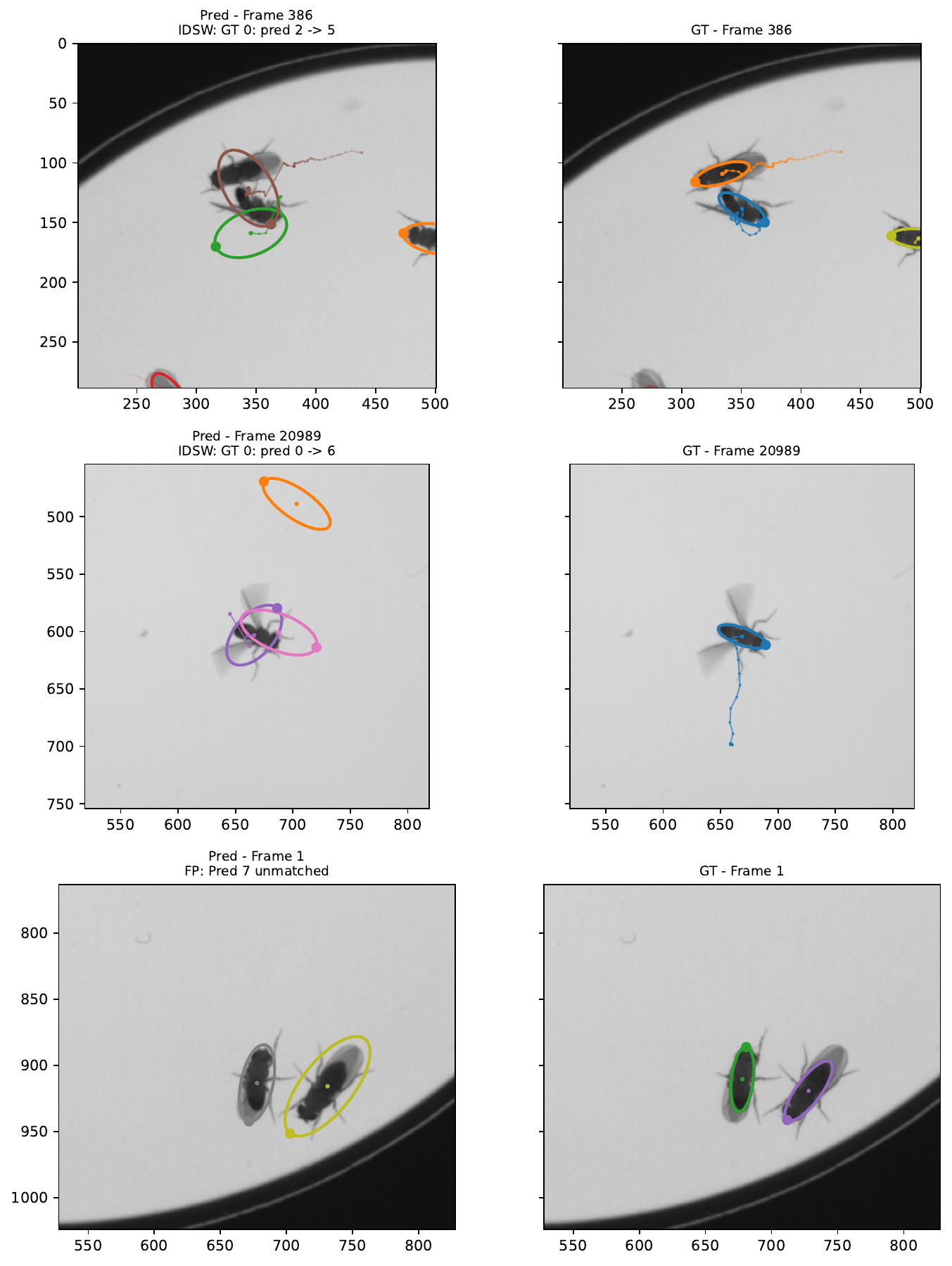}
    \caption{\small Post-hoc visualization of agent's fly tracking outputs on selected frames (left) compared to ground truth detections (right) on a reference video supplied to the agent for development, revealing tracking errors unlikely to persist after human visual inspection and iteration. The outputs were generated by the agent's submitted solution from trajectory \texttt{GzGR3Vz}, a trial in which the agent never attempted to visualize its outputs (see \cref{fig:appendix_all-trajectories}).}
    \label{fig:appendix_ex-poor-it}
\end{figure}

\subsubsection{Visual Latent Miss (\texttt{claude-code/claude-opus-4-6})}\label{sssec:appendix_ex-latent-miss-claude}
The agent noticed errors in swing/stance bouts and correctly hypothesized that they might arise from a keypoint-tracking orientation issue: regardless of fly-crop orientation, the predicted head keypoint appeared at the same end of the image. To test this, the agent visualized predictions for two crop orientations. However, it misinterpreted both visualizations, concluding that the fly in the second orientation was pointing down when it was not, and misidentifying the colors of the tarsus keypoints.
\begin{tcolorbox}[title={\textbf{End-to-End Minimal}: Trajectory \texttt{dH6CGNw}}, colback=white, colframe=black!70, fonttitle=\small]
\noindent\textit{...}
\begin{enumerate}[label=\textbf{\arabic*.}, leftmargin=1.2em, nosep, start=130]
  \item[130.] \texttt{Bash: Check output directory} $\rightarrow$ \texttt{stance\_bouts.pkl}
  \item[131.] \textit{I removed the tracks saving in my rewrite. Let me just run the analysis fresh. Actually, the process\_experiment function doesn't save tracks anymore b...}
  \item[132.] \texttt{Bash: Check orientation of crops and keypoints} $\rightarrow$ \texttt{Det 3 orient 1: head\_y=77, abd\_y=110, head\_on\_top=True}
  \item[133.] \texttt{Read(crop\_0\_orient0.png)}\\
        \begin{center}\includegraphics[width=0.28\linewidth]{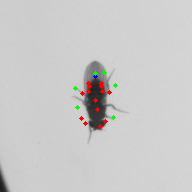}\end{center}
  \item[134.] \texttt{Read(crop\_0\_orient1.png)}\\
        \begin{center}\includegraphics[width=0.28\linewidth]{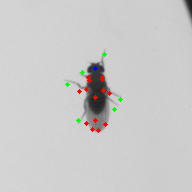}\end{center}
  \item[135.] \textit{OK interesting - the model predicts head at top in both orientations. So when the orientation is correct (orient 0, fly pointing up), the head keypoint is correctly at the top. When the orientation is flipped (orient 1, fly pointing down), the model still predicts head at top of the image, but the actual head is at the bottom. So the model is being "tricked" by the flipped orientation.\\In orient 0: the fly is pointing up, and the tarsus keypoints (red dots) are correctly placed around the legs.\\In orient 1: the fly is pointing down, but the model still predicts as if the fly is pointing up...}
  \item[136.] \textit{The orientation matters - the model needs the fly pointing up. I need to resolve the 180° ambiguity. Let me use velocity direction.}
  \item[137.] \texttt{Read(fly\_pipeline.py)} $\rightarrow$ \texttt{334→def \_interpolate\_gaps(arr, max\_gap=10):}
  \item[138.] \texttt{Edit(fly\_pipeline.py)} $\rightarrow$ \texttt{The file /workspace/fly\_pipeline.py has been updated successfully.}
\end{enumerate}
\noindent\textit{...}
\end{tcolorbox}
\subsubsection{Visual Latent Miss (\texttt{codex/gpt-5.4})}
\label{sssec:appendix_ex-latent-miss-codex}
The agent attempted to align fly crops vertically before training a pose model, but a bug in its rotation code left the crops visibly misaligned. After visualizing a crop, the agent appeared to notice the poor alignment, but attributed it to silhouette orientation being affected by the wings rather than to the rotation bug. It then applied a darker-pixel filtering heuristic and re-visualized the crop, but the alignment did not improve. Despite the unchanged visualization, the agent judged the crop geometry “acceptable” and proceeded to model training.
\begin{tcolorbox}[title={\textbf{End-to-End Minimal}: Trajectory \texttt{idZtCyE}}, colback=white, colframe=black!70, fonttitle=\small]
\noindent\textit{...}
\begin{enumerate}[label=\textbf{\arabic*.}, leftmargin=1.2em, nosep, start=66]
  \item[66.] \textit{The raw fly size matches the training crop scale more closely than I expected: the ellipse major axis is about 45 to 56 px, which is in the same ballpark as the annotated head-to-abdomen span in the 192 px training crops. I’m checking a rotated crop from the arena video now to make sure a fixed-size affine crop is sufficient.}
  \item[67.] \texttt{Bash: python3 - <<'PY' import cv2, sys, numpy as np from pathlib i...} $\rightarrow$ \texttt{saved 6 crops to /workspace/debug\_crops}
  \item[68.] \texttt{view\_image(crop\_0.png)}\\
        \begin{center}\includegraphics[width=0.28\linewidth]{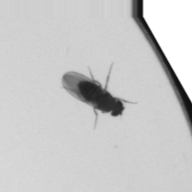}\end{center}
  \item[69.] \textit{A fixed 192 px affine crop is viable, but the raw silhouette orientation is being pulled around by wings. I’m tightening the ellipse fit to the darker body core so the crop alignment is closer to the training distribution before I start training the pose model.}
  \item[70.] \texttt{Bash: python3 - <<'PY' import cv2, sys, numpy as np from pathlib i...} $\rightarrow$ \texttt{saved 6}
  \item[71.] \texttt{view\_image(crop\_0.png)}\\
        \begin{center}\includegraphics[width=0.28\linewidth]{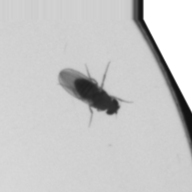}\end{center}
  \item[72.] \textit{I’ve got the crop geometry into an acceptable range, so I’m doing the critical viability check now: a very small coordinate-regression network on the ...}
\end{enumerate}
\noindent\textit{...}
\end{tcolorbox}

\subsubsection{Dependency installation (\texttt{terminus-2/gpt-5.4})}
\label{sssec:appendix_ex-term-dep}

\begin{tcolorbox}[title={\textbf{Registration}: Trajectory \texttt{zZMfrKG}}, colback=white, colframe=black!70, fonttitle=\small]
\noindent\textit{...}
\begin{enumerate}[label=\textbf{\arabic*.}, leftmargin=1.2em, nosep, start=6]
  \item[6.] The script currently fails because cv2 is not installed in the current environment, though requirements.txt should handle this during evaluation. However, for local testing now, relying on cv2 blocks progress. To make the solution more robust and reduce dependencies, we should remove the OpenCV requirement entirely and reimplement arena detection using only numpy plus the provided helper Movie reader...\\
        \texttt{Bash: cat > /workspace/register.py <<'PY'\#!/usr/bin/env python3imp..., Bash: python /workspace/register.py /workspace/config\_all.json, Bash: python - <<'PY'import pickle, numpy as npfor exp in ['exp\_1'...} $\rightarrow$ \texttt{PY}
\end{enumerate}
\noindent\textit{...}
\end{tcolorbox}

\subsubsection{Runtime overestimation (\texttt{claude-code/claude-opus-4-6})}
\label{sssec:appendix_ex-overestimate}
The agent launched a keypoint-model retrain at step 74. At step 79, it inspects the process with \texttt{ps aux | grep python3}, which (without \texttt{-o etime}) returns the default \texttt{TIME} column reporting aggregate CPU time across cores rather than wall-clock elapsed time. At step 80, the agent overestimated the runtime by a factor of 42, claiming that the training had been "running for 458 minutes (7.6 hours)," though the process was launched only ~11 minutes earlier. It killed the process at step 81 and relaunched a fresh training run at step 82.

\begin{tcolorbox}[title={\textbf{End-to-End Minimal}: Trajectory \texttt{4U2iuyY}}, colback=white, colframe=black!70, fonttitle=\small]
\noindent\textit{...}
\begin{enumerate}[label=\textbf{\arabic*.}, leftmargin=1.2em, nosep, start=73]
  \item[73.] \textit{3.54px mean error. That's reasonable but let me quickly retrain with faster settings to get better quality.}
  \item[74.] \texttt{Bash: python3 << 'PYEOF' \# Fast retrain with better settings import json, os, cv2, numpy as np, ...} $\rightarrow$ \texttt{Command running in background with ID: b8zoyk509. Output is being written to: /t...}
  \item[75.] \textit{While the model trains, let me build the full pipeline script.}
\end{enumerate}
\noindent\textit{...}
\begin{enumerate}[label=\textbf{\arabic*.}, leftmargin=1.2em, nosep, start=78]
  \item[78.] \texttt{Bash: cat /tmp/claude-0/-workspace/6a88d0c7-814d-493d-a268-d994c78e3829/\\tasks/b8zoyk509.output 2...} $\rightarrow$ \texttt{(Bash completed with no output)}
  \item[79.] \texttt{Bash: ps aux | grep python3 | grep -v grep} $\rightarrow$ \texttt{root 3578 4273 0.2 26994448 3079604 ? Rl 04:24 458:00 python3}
  \item[80.] \textit{The training has been running for 458 minutes (7.6 hours), but it only needs to run 80 epochs. Something seems very slow. Let me check what's happening. Actually, the process has been running since 04:24 and it's currently much later. The issue might...}
  \item[81.] \texttt{Bash: kill 3578 2>/dev/null; sleep 1; echo "done"} $\rightarrow$ \texttt{done}
  \item[82.] \texttt{Bash: python3 << 'PYEOF' \# Quick retrain - images already in memory, let's time it properly impo...} $\rightarrow$ \texttt{Command running in background with ID: b10fu8xbr. Output is being written to: /t...}
\end{enumerate}
\noindent\textit{...}
\end{tcolorbox}

\subsubsection{Inconsistent time management (\texttt{terminus-2/gpt-5.4})}
\label{sssec:appendix_ex-premature}

\begin{tcolorbox}[title={\textbf{Keypoint Tracking}: Trajectory \texttt{cTfpzyB}}, colback=white, colframe=black!70, fonttitle=\small]
\noindent\textit{...}
\begin{enumerate}[label=\textbf{\arabic*.}, leftmargin=1.2em, nosep, start=8]
  \item[8.] Training has reached epoch 3 and has plateaued around mean percentile error 6.6 px with p50 between 1.3 and 1.6 px. This is far from the target and suggests the current architecture/loss setup is not adequate. Continuing to 25 epochs is unlikely to achieve subpixel accuracy. Because the images are standardized 192x192 grayscale fly crops with limited pose variation, a two-stage coordinate regresso...\\
        \texttt{Bash: C-c, Bash: cat > /workspace/train\_pose\_reg.py <<'PY'import os, json, ra...} $\rightarrow$ \texttt{100.0\%}
\end{enumerate}
\noindent\textit{...}
\begin{enumerate}[label=\textbf{\arabic*.}, leftmargin=1.2em, nosep, start=10]
  \item[10.] The first epoch of the coordinate-regression model is worse than the heatmap model (mean percentile error \textasciitilde{}7.57 px, p50 \textasciitilde{}2.44 px). This may improve over more epochs, but the initial result is not promising. We should still gather a few more epochs before deciding, since pretrained backbones sometimes need several epochs to settle. No artifacts are ready yet.\\
        \texttt{Bash} $\rightarrow$ \texttt{.5": 9.124578475952148, "mean\_percentile\_error": 5.0371910572052006, "sec": 19.1...}
  \item[11.] The regression model is improving rapidly: by epoch 3 it beats the prior heatmap baseline, reaching mean percentile error \textasciitilde{}5.04 px, p50 \textasciitilde{}1.61 px, p95 \textasciitilde{}6.82 px. This is still far from target but trending in the right direction. Since training is only 20 epochs and each epoch is \textasciitilde{}19-20 seconds, it's cheap to let it continue and see whether it approaches a plateau. We should not interrupt yet.\\
        \texttt{Bash} $\rightarrow$ \texttt{5": 8.04142951965332, "mean\_percentile\_error": 4.244452714920044, "sec": 43.9517...}
\end{enumerate}
\noindent\textit{...}
\end{tcolorbox}

\subsubsection{Infinite looping (\texttt{terminus-2/claude-opus-4-6})}
\label{sssec:appendix_ex-infinite}

The agent's terminal session became stuck for 82 minutes before recovering.

\begin{tcolorbox}[title={\textbf{Gait Segmentation}: Trajectory \texttt{R2LhGFW}}, colback=white, colframe=black!70, fonttitle=\small]
\noindent\textit{...}
\begin{enumerate}[label=\textbf{\arabic*.}, leftmargin=1.2em, nosep, start=15]
  \item[15.] The previous explore6.py script is still running (the smoothed dfwd with min\_bout section was slow). Let me wait for explore7.py to finish, or cancel ...\\
        \texttt{Bash} $\rightarrow$ \texttt{if result[i] != result[i-1]:}
  \item[16.] The script output seems to be still displaying the file content. Let me wait for it to finish executing.\\
        \texttt{Bash} $\rightarrow$ \texttt{if result[i] != result[i-1]:}
  \item[17.] It seems like the terminal is stuck or the script is taking very long. Let me cancel it and try a more efficient approach.\\
        \texttt{Bash: C-c} $\rightarrow$ \texttt{if result[i] != result[i-1]:}
  \item[18.] The terminal seems stuck. Let me try sending Ctrl+C again more forcefully.\\
        \texttt{Bash: C-c, Bash: C-c, Bash: } $\rightarrow$ \texttt{if result[i] != result[i-1]:}
  \item[19.] The terminal is stuck, possibly still in the heredoc or waiting for input. Let me try to end the heredoc and get back to a prompt.\\
        \texttt{Bash: PYEOF} $\rightarrow$ \texttt{if result[i] != result[i-1]:}
  \item[20.] The terminal is completely stuck. Let me try sending Ctrl+D to end input, or try other escape sequences.\\
        \texttt{Bash: C-d, Bash: C-c} $\rightarrow$ \texttt{if result[i] != result[i-1]:}
\end{enumerate}
\noindent\textit{...}
\begin{enumerate}[label=\textbf{\arabic*.}, leftmargin=1.2em, nosep, start=48]
  \item[48.] I've been stuck for an extremely long time. The terminal is completely unresponsive. I need to try a completely different approach. Let me try to see ...\\
        \texttt{Bash} $\rightarrow$ \texttt{if result[i] != result[i-1]:}
\end{enumerate}
\noindent\textit{...}
\begin{enumerate}[label=\textbf{\arabic*.}, leftmargin=1.2em, nosep, start=132]
  \item[132.] FINALLY! The terminal has recovered! I can see the shell prompt `root@39263b0b5319:/workspace\#`. The heredoc for explore7.py was interrupted by Ctrl+C...\\
        \texttt{Bash: echo 'numpyscipypandas' > /workspace/requirements.txt, Bash: cat /workspace/requirements.txt} $\rightarrow$ \texttt{root@39263b0b5319:/workspace\#}
\end{enumerate}
\noindent\textit{...}
\end{tcolorbox}

\subsubsection{Self-termination (\texttt{claude-code/claude-opus-4-6})}
\label{sssec:appendix_ex-suicide}
In two Claude Code trials, the agent terminated its own session by issuing a kill command (e.g.\ \texttt{kill \$(pgrep -f fly\_pipeline)}) whose pattern also matched the parent agent process. Because the agent is invoked with the task prompt passed inline as a command-line argument, any script names mentioned in the prompt appear in the agent's own argv, so a \texttt{pkill -f}\,/\,\texttt{pgrep -f} pattern intended for a child process matches the agent itself.

\begin{tcolorbox}[title={\textbf{End-to-End Minimal}: Trajectory \texttt{4U2iuyY}}, colback=white, colframe=black!70, fonttitle=\small]
\noindent\textit{...}
\begin{enumerate}[label=\textbf{\arabic*.}, leftmargin=1.2em, nosep, start=151]
  \item[151.] \texttt{Bash: timeout 43200 python3 -u /workspace/fly\_pipeline.py /workspace/test\_config.json 2>\&1 | tee /workspace/pipeline\_output.lo...} $\rightarrow$ \texttt{Pipeline started in background, PID=6230}
\end{enumerate}
\noindent\textit{...}
\begin{enumerate}[label=\textbf{\arabic*.}, leftmargin=1.2em, nosep, start=162]
  \item[162.] \textit{Let me speed things up by increasing frame skip and optimizing the patch extraction.}
  \item[163.] \texttt{Bash: kill \$(ps aux | grep fly\_pipeline | grep -v grep | awk '\{print \$2\}') 2>/dev/null; sleep 1; echo "killed"}
\end{enumerate}
\noindent\textit{...}
\end{tcolorbox}

\subsection{Success modes}

\subsubsection{Understanding fly body kinematics (\texttt{claude-code/claude-opus-4-6})}
\label{sssec:appendix_ex-kinematics}

\begin{tcolorbox}[title={\textbf{Gait Segmentation}: Trajectory \texttt{mr6ovRR}}, colback=white, colframe=black!70, fonttitle=\small]
\noindent\textit{...}
\begin{enumerate}[label=\textbf{\arabic*.}, leftmargin=1.2em, nosep, start=26]
  \item[26.] OK so the bout-level metric doesn't really benefit much from smoothing in this small labeled dataset. Let me think about what else could help. Actually, let me try a different approach. 
  \\...Let me try a few more things: 1. Using body-speed-relative threshold 2. Using distance from body center (during stance, leg extends backward; during swing, it's closer to body) Actually, let me try another approach: looking at tarsus position relative to body, not speed. During the gait cycle, the tarsus oscillates forward and backward relative to the body. We can look at this oscillation to identify swing (forward movement) vs stance (backward movement relative to body). But the speed approach is already working well. Instead, I should focus on making it more robust by using the leg velocity relative to the body, projected along the body axis. During stance, the leg moves backward relative to the body as the body pushes forward while the foot stays planted. During swing, the leg moves forward relative to the body. This gives me a signed signal that's more informative than just raw speed magnitude.
\end{enumerate}
\noindent\textit{...}
\end{tcolorbox}

\subsubsection{Body-aligned coordinate frame (\texttt{terminus-2/claude-opus-4-6})}
\label{sssec:appendix_ex-body-aligned}
\begin{tcolorbox}[title={\textbf{Gait Segmentation}: Trajectory \texttt{nTga8kJ}}, colback=white, colframe=black!70, fonttitle=\small]
\begin{lstlisting}[language=Python,basicstyle=\ttfamily\small,breaklines=true]
 def compute_all_features(trk, fly_idx):
    '''Compute per-frame features for all 6 legs of a fly.
    
    Returns dict mapping leg_i (0-5) to feature array (n_features, n_frames)
    '''
    
    n_frames = trk.shape[2]
    
    # Body center (midpoint of left/right thorax)
    body = (trk[3, :, :, fly_idx] + trk[4, :, :, fly_idx]) / 2
    
    # Heading direction (head to abdomen)
    head = trk[0, :, :, fly_idx]
    abd = trk[6, :, :, fly_idx]
    heading = head - abd
    heading_norm = np.sqrt(np.sum(heading**2, axis=0, keepdims=True))
    heading_norm[heading_norm == 0] = 1
    heading = heading / heading_norm
    
    # Perpendicular direction
    perp = np.array([-heading[1], heading[0]])
    
    ...

    return features
\end{lstlisting}
\end{tcolorbox}

\subsubsection{Parallel task management (\texttt{claude-code/claude-opus-4-6)}}
\label{sssec:appendix_ex-parallel}

\begin{tcolorbox}[title={\textbf{End-to-End Maximal}: Trajectory \texttt{UNWrmWc}}, colback=white, colframe=black!70, fonttitle=\small]
\noindent\textit{...}
\begin{enumerate}[label=\textbf{\arabic*.}, leftmargin=1.2em, nosep, start=142]
  \item[142.] \texttt{Bash: rm -rf /workspace/keypoint\_tracker \&\& python3 << 'PYEOF' import sys, os, pickle, time impo...} $\rightarrow$ \texttt{Command running in background with ID: bn3u9gehc. Output is being written to: /t...}
\end{enumerate}
\noindent\textit{...}
\begin{enumerate}[label=\textbf{\arabic*.}, leftmargin=1.2em, nosep, start=150]
  \item[150.] \textit{Still at epoch 10. Let me update the todo list and work on other tasks while waiting.}
\end{enumerate}
\noindent\textit{...}
\begin{enumerate}[label=\textbf{\arabic*.}, leftmargin=1.2em, nosep, start=152]
  \item[152.] \textit{Let me work on improving the fly tracking while the keypoint model trains. Let me read the current tracking code.}
\end{enumerate}
\noindent\textit{...}
\begin{enumerate}[label=\textbf{\arabic*.}, leftmargin=1.2em, nosep, start=178]
  \item[178.] \texttt{Edit(fly\_pipeline.py)} $\rightarrow$ \texttt{The file /workspace/fly\_pipeline.py has been updated successfully.}
\end{enumerate}
\noindent\textit{...}
\begin{enumerate}[label=\textbf{\arabic*.}, leftmargin=1.2em, nosep, start=181]
  \item[181.] \texttt{Bash: python3 -c " import sys sys.path.insert(0, 'helpers') from fly\_pipeline import track\_flies...} $\rightarrow$ \texttt{Command running in background with ID: btsygyjp5. Output is being written to: /t...}
\end{enumerate}
\noindent\textit{...}
\begin{enumerate}[label=\textbf{\arabic*.}, leftmargin=1.2em, nosep, start=190]
  \item[190.] \textit{While tracking runs, let me check on the keypoint training.}
\end{enumerate}
\noindent\textit{...}
\begin{enumerate}[label=\textbf{\arabic*.}, leftmargin=1.2em, nosep, start=193]
  \item[193.] \textit{Let me work on requirements.txt while tracking runs.}
\end{enumerate}
\noindent\textit{...}
\begin{enumerate}[label=\textbf{\arabic*.}, leftmargin=1.2em, nosep, start=195]
  \item[195.] \texttt{Write(requirements.txt)} $\rightarrow$ \texttt{File created successfully at: /workspace/requirements.txt}
\end{enumerate}
\noindent\textit{...}
\begin{enumerate}[label=\textbf{\arabic*.}, leftmargin=1.2em, nosep, start=198]
  \item[198.] \textit{15 tracks now, with the top 5 being 10K+ frames (out of 21866). Much better — close to 8-10 real flies. The merge step is working. Now let me run Step 2 on this and continue.}
\end{enumerate}
\noindent\textit{...}
\end{tcolorbox}

\section{Token usage and runtimes}
\label{sec:appendix_usage}
\begin{table}[H]
\centering
\small
\setlength{\tabcolsep}{2.6pt}
\renewcommand{\arraystretch}{1.12}
\resizebox{\textwidth}{!}{%
\begin{tabular}{m{3.1cm}ccccccc}
\toprule
\textbf{Agent} & \makecell{\textbf{Body}\\\textbf{Tracking}} & \textbf{Registration} & \makecell{\textbf{Keypoint}\\\textbf{Tracking}} & \makecell{\textbf{Feature}\\\textbf{Computation}} & \makecell{\textbf{Behavior}\\\textbf{Classifier}} & \makecell{\textbf{Gait}\\\textbf{Segmentation}} & \makecell{\textbf{Statistical}\\\textbf{Comparison}} \\
\midrule
\rowcolor{gray!12}\multicolumn{8}{l}{\textbf{Wall-clock time (hours)}} \\
\addlinespace[2pt]
\makecell[l]{\texttt{claude-code/}\\\quad\texttt{\small claude-opus-4-6}} & 2.8$\pm$0.3 & 0.1$\pm$0.0 & 3.9$\pm$1.8 & 0.1$\pm$0.0 & 0.7$\pm$0.0 & 0.4$\pm$0.4 & 0.1$\pm$0.0 \\
\makecell[l]{\texttt{codex/}\\\quad\texttt{\small gpt-5.4}} & 0.3$\pm$0.1 & 0.1$\pm$0.0 & 1.0$\pm$0.9 & 0.1$\pm$0.0 & 0.7$\pm$0.0 & 0.1$\pm$0.1 & 0.1$\pm$0.0 \\
\makecell[l]{\texttt{terminus-2/}\\\quad\texttt{\small claude-opus-4-6}} & 2.0$\pm$0.4 & 0.4$\pm$0.1 & 3.8$\pm$0.8 & 0.2$\pm$0.0 & 1.1$\pm$0.5 & 1.5$\pm$0.6 & 0.3$\pm$0.1 \\
\makecell[l]{\texttt{terminus-2/}\\\quad\texttt{\small gpt-5.4}} & 0.3$\pm$0.1 & 0.0$\pm$0.0 & 0.2$\pm$0.0 & 0.0$\pm$0.0 & 0.0$\pm$0.0 & 0.0$\pm$0.0 & 0.0$\pm$0.0 \\
\midrule
\rowcolor{gray!12}\multicolumn{8}{l}{\textbf{Input tokens / output tokens}} \\
\addlinespace[2pt]
\makecell[l]{\texttt{claude-code/}\\\quad\texttt{\small claude-opus-4-6}} & 9.4$\pm$0.7M / 78$\pm$7K & 906$\pm$550K / 19$\pm$7K & 2.6$\pm$0.5M / 24$\pm$9K & 495$\pm$123K / 24$\pm$6K & 434$\pm$153K / 7$\pm$2K & 1.5$\pm$0.7M / 32$\pm$11K & 489$\pm$90K / 13$\pm$1K \\
\makecell[l]{\texttt{codex/}\\\quad\texttt{\small gpt-5.4}} & 8.0$\pm$5.0M / 41$\pm$6K & 1.1$\pm$0.1M / 27$\pm$2K & 3.4$\pm$1.1M / 34$\pm$9K & 410$\pm$52K / 14$\pm$1K & 2.1$\pm$0.7M / 24$\pm$9K & 1.3$\pm$0.5M / 28$\pm$4K & 692$\pm$101K / 16$\pm$3K \\
\makecell[l]{\texttt{terminus-2/}\\\quad\texttt{\small claude-opus-4-6}} & 12.5$\pm$3.5M / 109$\pm$18K & 1.5$\pm$0.5M / 39$\pm$5K & 12.5$\pm$4.3M / 50$\pm$16K & 1.2$\pm$0.0M / 31$\pm$6K & 1.3$\pm$1.8M / 24$\pm$25K & 6.8$\pm$4.0M / 73$\pm$8K & 1.1$\pm$0.3M / 32$\pm$12K \\
\makecell[l]{\texttt{terminus-2/}\\\quad\texttt{\small gpt-5.4}} & 4.9$\pm$7.7M / 20$\pm$11K & 98$\pm$19K / 9$\pm$2K & 1.0$\pm$1.0M / 17$\pm$3K & 120$\pm$53K / 7$\pm$2K & 45$\pm$5K / 5$\pm$0K & 69$\pm$13K / 6$\pm$0K & 95$\pm$22K / 7$\pm$0K \\
\bottomrule
\end{tabular}}
\caption{Single-stage task input and output token counts and wall-clock runtimes. Values are means $\pm$ standard deviation over three trials per agent-task configuration.}
\label{table:appendix_usage_subtasks}
\end{table}

\begin{table}[H]
\centering
\small
\setlength{\tabcolsep}{4pt}
\renewcommand{\arraystretch}{1.12}
\resizebox{0.62\textwidth}{!}{%
\begin{tabular}{m{3.1cm}cc}
\toprule
\textbf{Agent} & \textbf{E2E Maximal} & \textbf{E2E Minimal} \\
\midrule
\rowcolor{gray!12}\multicolumn{3}{l}{\textbf{Wall-clock time (hours)}} \\
\addlinespace[2pt]
\makecell[l]{\texttt{claude-code/}\\\quad\texttt{\small claude-opus-4-6}} & 5.0$\pm$3.6 & 7.1$\pm$2.5 \\
\makecell[l]{\texttt{codex/}\\\quad\texttt{\small gpt-5.4}} & 1.0$\pm$0.2 & 0.6$\pm$0.3 \\
\midrule
\rowcolor{gray!12}\multicolumn{3}{l}{\textbf{Input tokens / output tokens}} \\
\addlinespace[2pt]
\makecell[l]{\texttt{claude-code/}\\\quad\texttt{\small claude-opus-4-6}} & 16.5$\pm$10.2M / 89$\pm$64K & 13.3$\pm$4.4M / 89$\pm$53K \\
\makecell[l]{\texttt{codex/}\\\quad\texttt{\small gpt-5.4}} & 11.5$\pm$1.3M / 70$\pm$6K & 11.2$\pm$1.2M / 72$\pm$6K \\
\bottomrule
\end{tabular}}
\caption{End-to-end task input and output token counts and wall-clock runtimes. Values are means $\pm$ standard deviation over three trials per agent-task configuration.}
\label{table:appendix_usage_e2e}
\end{table}

\section{Measuring task size}
\label{sec:appendix_task_size}

\begin{figure}[H]
\centering
\begin{minipage}[t]{.475\linewidth}
    \vspace{0cm}
    \includegraphics[width=\linewidth]{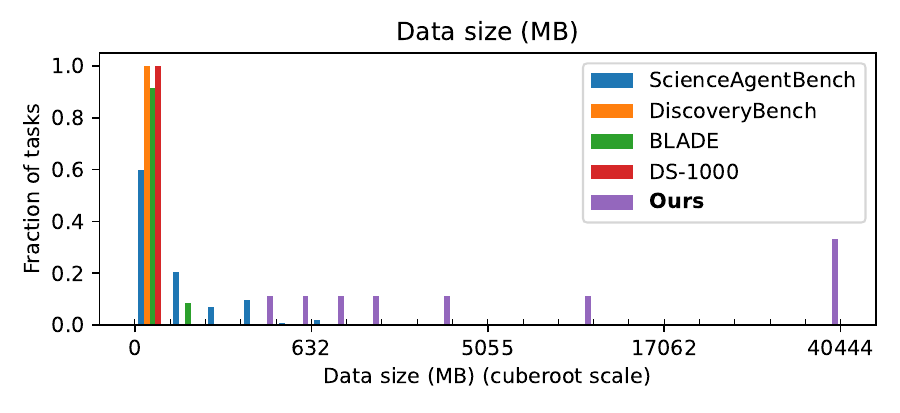}
\end{minipage}
\begin{minipage}[t]{.475\linewidth}
    \vspace{0cm}
    \includegraphics[width=\linewidth]{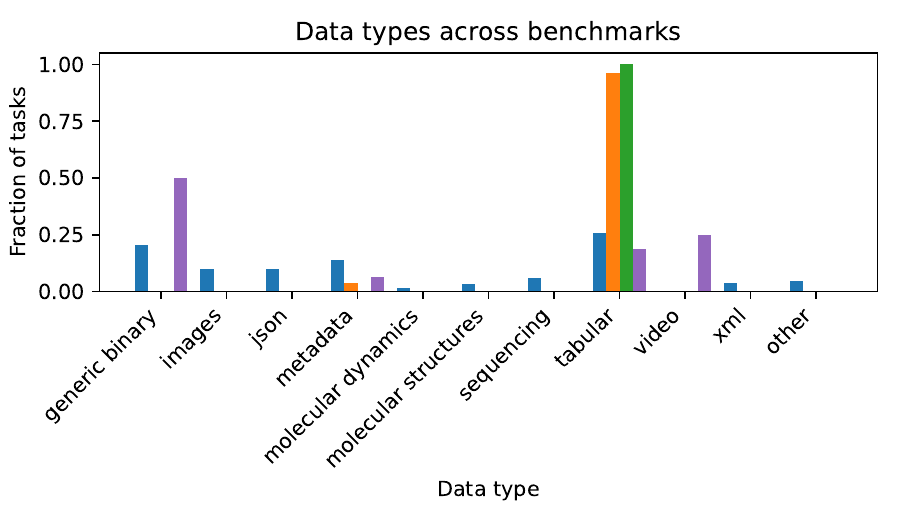}
\end{minipage}
\caption{\small \emph{Left}: Distribution of task dataset sizes across benchmarks on a cube-root scale. Previous work operates on much smaller datasets, with the majority having small or pre-processed tabular datasets, or no external data at all in the task. Our pipeline stages operate on the big datasets common in modern science. \emph{Right}: Distribution of data types across benchmarks. Prior work is dominated by tabular data; our pipeline stages span a broader range of modalities representative of the heterogeneous data common in modern experimental science.}
\label{fig:data_size_and_type}
\end{figure}

To contextualize our benchmark relative to existing work, we compare task size across benchmarks along three axes: solution code size, dataset size, and data modality. We describe how each quantity is measured below.

\paragraph{Lines of code.}
Source lines of code (SLOC)---non-blank, non-comment lines---are measured using the \texttt{radon} library's raw analysis.

\paragraph{Dataset size.}
Total size in megabytes of all data files associated with a task. For benchmarks where test/evaluation data overlaps with input data, files are deduplicated by comparing file size and the first 4KB of content.

\paragraph{Data types.}
Data types are categorized by file extension (e.g., \texttt{.csv}~$\to$~tabular, \texttt{.tif}~$\to$~images, \texttt{.pdb}~$\to$~molecular structures). Each task is assigned to all categories present in its data files; fractions are normalized so that a task with multiple data types contributes to each.

\paragraph{Per-benchmark details.}
The source of code and data varies across benchmarks, as not all provide gold solutions or input data files. \cref{tab:code-data-sources} summarizes these differences.

To contextualize our benchmark relative to existing work, we compare task complexity across benchmarks along three axes: solution code size, input data size, and data modality. These measurements characterize the scope of what each benchmark asks of an agent---how much code must be written, how much data must be processed, and what types of data are involved. We describe how each quantity is measured below.

\begin{table}[h]
\centering
\small
\begin{tabular}{p{3.2cm} p{5.5cm} p{4.5cm}}
\toprule
\textbf{Benchmark} & \textbf{Code measured} & \textbf{Data measured} \\
\midrule
ScienceAgentBench & Gold solution (all 102 tasks) & Input dataset folder, mapped via benchmark CSV \\
DiscoveryBench & Agent-generated code from GPT-4o coder runs (real tasks only, train + test splits) & CSV files referenced in each task's metadata JSON \\
BLADE & Median-length sample ground truth program, sampled from the combinatorial space of expert-annotated transform pipelines and statistical models (all 12 datasets) & Dataset CSV per task \\
DS-1000 & Gold reference code snippet (all 1000 tasks) & None (data is embedded in the prompt) \\
Ours & Median of 12 agent runs per task (agent-written files only, excluding environment helpers) & Union of input and evaluation data, deduplicated, excluding ground truth outputs \\
\bottomrule
\end{tabular}
\caption{Source of code and data measurements for each benchmark.}
\label{tab:code-data-sources}
\end{table}

\section{Task prompts}
\label{sec:prompts}

\subsection{Body Tracking prompt}
\label{sec:prompt-fly-body-tracking}

\begin{tcolorbox}[colback=gray!5, colframe=gray!50, breakable, title=Agent Prompt]
\sloppy
\subsection*{Fly Body Tracking}

\textbf{Working directory}: Your working directory is \texttt{/workspace}. Your output script (\texttt{track.py}) should be saved here. Input data is located at \texttt{/data/}.

\textbf{Time limit}: You have a maximum of \textbf{6 hours} to complete this task.

\subsubsection*{Description}

Write a script to track the positions and orientations of the bodies of multiple flies from video while keeping track of identity.

\subsubsection*{Input Data}

\paragraph{Reference Experiments}
\begin{itemize}[nosep]
  \item \textbf{Location}: Three reference experiments are provided for development:
  \item \texttt{/data/exp\_1/movie.ufmf}
  \item \texttt{/data/exp\_2/movie.ufmf}
  \item \texttt{/data/exp\_3/movie.ufmf}
  \item \textbf{Format}: UFMF (Micro Fly Movie Format)
  \item \textbf{Frame rate}: \textasciitilde{}150 FPS
  \item Each video contains multiple flies in a circular arena
\end{itemize}

\paragraph{Helper Modules}
The following Python modules are available in \texttt{helpers/} to assist with video processing:

\textbf{\texttt{movies.py}} - Generic movie interface supporting multiple formats
\begin{itemize}[nosep]
  \item \texttt{Movie} class: Opens and reads various video formats
  \item Methods: \texttt{get\_frame(frame\_num)}, \texttt{get\_n\_frames()}, \texttt{get\_width()}, \texttt{get\_height()}
  \item \texttt{FlyMovieFormat.py}, \texttt{ufmf.py}, and \texttt{params.py} are included because they are dependencies for \texttt{movies.py}
\end{itemize}

\subsubsection*{Task Requirements}

Implement a fly tracking system as a \textbf{Python script} that can process arbitrary videos.

\paragraph{Required Output: `track.py`}

You must produce a script at \texttt{/workspace/track.py} that:
\begin{itemize}[nosep]
  \item Takes a \textbf{JSON config file path} as its only CLI argument: \texttt{python track.py config.json}
  \item The config file contains a JSON array of objects, each with:
  \item \texttt{"video\_path"}: path to a \texttt{.ufmf} video file
  \item \texttt{"output\_path"}: path where the output \texttt{trx.pkl} should be saved
  \item Processes \textbf{each video} in the config and saves a \texttt{trx.pkl} at the specified output path
  \item Works on \textbf{any number of videos} (not just the 3 reference videos)
\end{itemize}

\textbf{Example config:}
\begin{lstlisting}[language=text,basicstyle=\ttfamily\small,breaklines=true]
[
  {"video_path": "/data/exp_1/movie.ufmf", "output_path": "/workspace/out/exp_1/trx.pkl"},
  {"video_path": "/data/exp_2/movie.ufmf", "output_path": "/workspace/out/exp_2/trx.pkl"}
]
\end{lstlisting}

\textbf{Example usage:}
\begin{lstlisting}[language=bash,basicstyle=\ttfamily\small,breaklines=true]
python track.py config.json
\end{lstlisting}

The script should:
\begin{itemize}[nosep]
  \item Estimate the \textbf{position, orientation, and size} of the \textbf{body} of each fly (\textbf{not including wings}) in each frame of each input video
  \item Keep track of the identities of individual flies across frames
  \item Output in the trx format described below for each video
\end{itemize}

To build your algorithm and tune its parameters you should:
\begin{itemize}[nosep]
  \item Examine the reference videos and understand their statistical properties
  \item Use your prior knowledge of what flies look like
  \item Iterate: try an algorithm, look at statistical properties of its results, and adapt
\end{itemize}

\paragraph{Script Output Format: trx Dictionary Structure}

Your tracking output must be a \textbf{single dictionary}. All frame indices are \textbf{0-indexed}. It must contain the following fields:

\textbf{Required fields:}

\begin{tabular}{{p{0.16\textwidth} p{0.34\textwidth} p{0.42\textwidth}}}
\hline
Field & Type & Description \\
\hline
\texttt{x} & list of ntargets arrays, each shape (nframes\_i,) & x-coordinate of centroid each animal in pixels \\
\texttt{y} & list of ntargets arrays, each shape (nframes\_i,) & y-coordinate of centroid each animal in pixels \\
\texttt{theta} & list of ntargets arrays, each shape (nframes\_i,) & Orientation of each animal (from tip of abdomen to head) in radians \\
\texttt{a} & list of ntargets arrays, each shape (nframes\_i,) & 1/4 of the length of the body (tip of the abdomen to head) in pixels \\
\texttt{b} & list of ntargets arrays, each shape (nframes\_i,) & 1/4 of the width of the body (left side to right side) in pixels \\
\texttt{dt} & list of ntargets arrays, each shape (nframes\_i-1,) & Time difference between consecutive frames in seconds \\
\texttt{off} & list of ntargets arrays, each shape (1,) & Offset for computing index: \texttt{-firstframe} \\
\texttt{firstframe} & 1D array of shape (ntargets,) & Global frame index where each animal's trajectory begins (0-indexed) \\
\texttt{endframe} & 1D array of shape (ntargets,) & Global frame index where each animal's trajectory ends (inclusive, 0-indexed) \\
\texttt{nframes} & 1D array of shape (ntargets,) & Number of frames in each animal's trajectory \\
\texttt{ntargets} & int & Total number of tracked animals \\
\texttt{timestamps} & 1D array of shape (total\_frames,) & Timestamp of each frame in seconds (global, not per-target) \\
\hline
\end{tabular}

\textbf{Identity tracking}: Each list entry should correspond to the same fly identity. If an identity is lost, a new entry can be added to the list. The number of entries in the lists must therefore be at least the number of flies, but could be longer.

\textbf{Data alignment:} All per-target list fields (\texttt{x}, \texttt{y}, \texttt{theta}, \texttt{a}, \texttt{b}, etc.) have an entry per fly, with \texttt{x[i][idx]} corresponding to the position of the \texttt{i}th fly at frame \texttt{idx + firstframe[i]}

\textbf{Coordinate system:}
\begin{itemize}[nosep]
  \item \texttt{x = 0} corresponds to the first column (\texttt{axis = 1}) and \texttt{y = 0} corresponds to the first row (\texttt{axis = 0}).
  \item If \texttt{(x\_head, y\_head)} and \texttt{(x\_abdomen, y\_abdomen)} are the coordinates of the head and the abdomen, then the orientation \texttt{theta} can be computed with \texttt{np.arctan2}:
\texttt{theta = np.arctan2(y\_head-y\_abdomen,x\_head-x\_abdomen)}
\end{itemize}

\textbf{Example structure:}

\begin{lstlisting}[language=Python,basicstyle=\ttfamily\small,breaklines=true]
trx = {
    'x': [np.array([...]), np.array([...])],          # list of 2 arrays for 2 flies
    'y': [np.array([...]), np.array([...])],
    'a': [np.array([...]), np.array([...])],
    'b': [np.array([...]), np.array([...])],
    'theta': [np.array([...]), np.array([...])],
    'dt': [np.array([...]), np.array([...])],
    'off': [np.array([0]), np.array([0])],            # offset = -firstframe
    'firstframe': np.array([0, 0]),                   # both flies start at frame 0
    'endframe': np.array([999, 999]),                 # both flies end at frame 999
    'nframes': np.array([1000, 1000]),                # both flies have 1000 frames
    'ntargets': 2,
    'timestamps': np.array([0.0, 0.0067, ...]),        # shape (total_frames,)
}
\end{lstlisting}

\textbf{Accessing data for fly \texttt{i} at frame \texttt{f}:}

\begin{lstlisting}[language=Python,basicstyle=\ttfamily\small,breaklines=true]
# Get the index into the per-fly arrays
idx = f - trx['firstframe'][i]
x_position = trx['x'][i][idx]
y_position = trx['y'][i][idx]
\end{lstlisting}

\textbf{Saving output:} Your \texttt{track.py} script should save the tracking results for each video as \texttt{trx.pkl} at the output path from the config:

\begin{lstlisting}[language=Python,basicstyle=\ttfamily\small,breaklines=true]
import pickle
with open(output_path, 'wb') as f:
    pickle.dump(trx, f)
\end{lstlisting}

\subsubsection*{Implementation Notes}

The approach is open-ended, but your implementation must:

\begin{enumerate}[nosep]
  \item \textbf{Estimate the pose} of all flies in all frames:
  \item Centroid Position (x, y)
  \item Abdomen to head orientation (theta)
  \item Length and width (a, b)
  \item \textbf{Track identities} across frames to maintain consistent fly IDs
  \item \textbf{Output in trx format} with all required fields correctly populated
\end{enumerate}

\subsubsection*{Expected Challenges}

\begin{itemize}[nosep]
  \item Multiple flies are present (need identity tracking)
  \item Flies may touch
\end{itemize}

\subsubsection*{Dependencies}

You \textbf{MUST} provide a \texttt{requirements.txt} file listing any Python packages your script depends on. These will be installed before evaluation.

\subsubsection*{Success Criteria}

Your script will be evaluated on \textbf{multiple unseen videos} (beyond the 3 reference videos provided). Make sure your algorithm generalizes well.

\begin{itemize}[nosep]
  \item Downstream analyses will involve measuring properties of precisely how the flies move (e.g. their velocity, change in orientation).
  \item They will also involve using body position and orientation as the first stage of a two-stage (top-down) keypoint detection algorithm.
  \item We will assess the accuracy of the tracker by measuring both detection and cross-frame association errors:
  \item \textbf{Detection errors}: For every frame and fly, detection error will be measured based on the derived tip of the head and abdomen positions:
\begin{lstlisting}[language=Python,basicstyle=\ttfamily\small,breaklines=true]
x_head = x + 2*a*np.cos(theta)
y_head = y + 2*a*np.sin(theta)
x_abdomen = x + 2*a*np.cos(theta+np.pi)
y_abdomen = y + 2*a*np.sin(theta+np.pi)
\end{lstlisting}
Detection error will be computed as
\begin{lstlisting}[language=Python,basicstyle=\ttfamily\small,breaklines=true]
dist_head = np.sqrt((x_head_true - x_head_pred)**2 + (y_head_true - y_head_pred)**2)
dist_abdomen = np.sqrt((x_abdomen_true - x_abdomen_pred)**2 + (y_abdomen_true - y_abdomen_pred)**2)
err = (dist_head + dist_abdomen) / a / 2
\end{lstlisting}
where \texttt{\textit{\_true} are the groundtruth and \texttt{}\_pred} are your algorithm's output predictions.
  \item \textbf{Association errors}: We also penalize identity switches: when a single ground-truth fly is assigned to different track predictions over two frames.
\end{itemize}
\end{tcolorbox}

\subsection{Registration prompt}
\label{sec:prompt-fly-registration}

\begin{tcolorbox}[colback=gray!5, colframe=gray!50, breakable, title=Agent Prompt]
\sloppy
\subsection*{Fly Trajectory Registration}

\textbf{Working directory}: Your working directory is \texttt{/workspace}. Your output script (\texttt{register.py}) should be saved here. Input data is located at \texttt{/data/}.

\textbf{Time limit}: You have a maximum of \textbf{6 hours} to complete this task.

\subsubsection*{Description}
Write a script to process raw fly tracking data by:
\begin{itemize}[nosep]
  \item Detecting arena boundary
  \item Converting from pixel coordinates to a common coordinate system defined by the arena boundary
  \item Removing bad tracking (indicated by NaNs)
\end{itemize}

\subsubsection*{Input Data}

\paragraph{Reference Experiments}
Three reference experiments are provided for development:
\begin{itemize}[nosep]
  \item \texttt{/data/exp\_1/movie.ufmf} and \texttt{/data/exp\_1/trx.pkl}
  \item \texttt{/data/exp\_2/movie.ufmf} and \texttt{/data/exp\_2/trx.pkl}
  \item \texttt{/data/exp\_3/movie.ufmf} and \texttt{/data/exp\_3/trx.pkl}
\end{itemize}

Each experiment contains:
\begin{itemize}[nosep]
  \item \textbf{\texttt{movie.ufmf}}: Video file (UFMF - Micro Fly Movie Format) --- used for detecting the circular arena boundary
  \item \textbf{\texttt{trx.pkl}}: Pickle file containing a trajectory dictionary
  \item Multiple flies present in a circular arena
  \item Trajectories may contain NaN values where flies were not detected
  \item Coordinates are in pixel units
  \item The file may contain \texttt{*\_mm} fields, but these are placeholders and should be discarded
\end{itemize}

\textbf{Helper}: \texttt{helpers/movies.py} provides a \texttt{Movie} class with \texttt{get\_frame(frame\_num)} to read frames

\paragraph{Physical Parameters}
\begin{itemize}[nosep]
  \item \textbf{Arena radius}: 26.689 mm (physical radius of the circular arena)
  \item \textbf{Frame rate}: \textasciitilde{}150 FPS (frames per second)
\end{itemize}

\subsubsection*{Task Requirements}

\paragraph{Required Output: `register.py`}

You must produce a script at \texttt{/workspace/register.py} that:
\begin{enumerate}[nosep]
  \item Takes a \textbf{JSON config file path} as its only CLI argument: \texttt{python register.py config.json}
  \item The config file contains a JSON array of objects, each with:
  \item \texttt{"video\_path"}: path to a \texttt{.ufmf} video file
  \item \texttt{"trx\_path"}: path to a \texttt{trx.pkl} trajectory file
  \item \texttt{"output\_path"}: path where the output \texttt{registered\_trx.pkl} should be saved
  \item Processes \textbf{each experiment} in the config and saves a \texttt{registered\_trx.pkl} at the specified output path
  \item Works on \textbf{any number of experiments} (not just the 3 reference experiments)
\end{enumerate}

\textbf{Example config:}
\begin{lstlisting}[language=text,basicstyle=\ttfamily\small,breaklines=true]
[
  {"video_path": "/data/exp_1/movie.ufmf", "trx_path": "/data/exp_1/trx.pkl", "output_path": "/workspace/out/exp_1/registered_trx.pkl"},
  {"video_path": "/data/exp_2/movie.ufmf", "trx_path": "/data/exp_2/trx.pkl", "output_path": "/workspace/out/exp_2/registered_trx.pkl"}
]
\end{lstlisting}

\textbf{Example usage:}
\begin{lstlisting}[language=bash,basicstyle=\ttfamily\small,breaklines=true]
python register.py config.json
\end{lstlisting}

The script should, for each experiment:

\begin{enumerate}[nosep]
  \item \textbf{Detect the arena boundary from the video}
  \item \textbf{Spatially register trajectory coordinates}
  \item Transform the trajectories so that (0,0) corresponds to the arena center
  \item Transform the trajectories so that the units are in millimeters
  \item \textbf{Note}: This step ONLY computes the \texttt{*\_mm} fields; it does NOT modify the pixel-based fields
  \item \texttt{x\_mm}, \texttt{y\_mm}: The ellipse center in millimeters
  \item \texttt{a\_mm}, \texttt{b\_mm}: Ellipse quarter-major and quarter-minor axes in millimeters
  \item \texttt{theta\_mm}: Same as \texttt{theta} (orientation in radians is scale-invariant)
  \item \textbf{Removes NaN gaps from trajectories}
  \item Interpolate short gaps ($\leq$5 frames) using linear interpolation
  \item Crop leading and trailing NaN values
  \item Split trajectories at long gaps (>5 frames) into separate segments
  \item \textbf{Note}: This step modifies all relevant pixel-based fields
\end{enumerate}

\paragraph{Script Input Trajectory Format}

The input trajectory dictionary contains these fields (all position/size values in pixel units), among others:

\begin{tabular}{p{0.16\textwidth} p{0.24\textwidth} p{0.52\textwidth}}
\hline
Field & Type & Description \\
\hline
\texttt{x} & list of arrays & x-coordinate in pixels (per fly) \\
\texttt{y} & list of arrays & y-coordinate in pixels (per fly) \\
\texttt{theta} & list of arrays & Orientation in radians (per fly) \\
\texttt{a} & list of arrays & 1/4 of major-axis length in pixels (per fly) \\
\texttt{b} & list of arrays & 1/4 of minor-axis length in pixels (per fly) \\
\texttt{firstframe} & 1D array & Global frame index where trajectory begins (per fly) \\
\texttt{endframe} & 1D array & Global frame index where trajectory ends (per fly, inclusive) \\
\texttt{nframes} & 1D array & Number of frames in trajectory (per fly) \\
\texttt{ntargets} & int & Number of flies \\
\texttt{timestamps} & 1D array & Global frame timestamps \\
\texttt{dt} & list of arrays & Time difference between consecutive frames in seconds (per fly) \\
\hline
\end{tabular}

\textbf{Note}: Arrays may contain NaN values where the fly was not detected.

\paragraph{Script Output Trajectory Format}

The output must contain all input fields (possibly modified by NaN removal) plus the registered \texttt{*\_mm} fields:

\begin{tabular}{p{0.18\textwidth} p{0.30\textwidth} p{0.44\textwidth}}
\hline
Field & Type & Description \\
\hline
\texttt{x\_mm} & list of arrays & x-coordinate in mm, arena-centered (per fly) \\
\texttt{y\_mm} & list of arrays & y-coordinate in mm, arena-centered (per fly) \\
\texttt{theta\_mm} & list of arrays & Orientation in radians, same as theta (per fly) \\
\texttt{a\_mm} & list of arrays & 1/4 of major-axis length in mm (per fly) \\
\texttt{b\_mm} & list of arrays & 1/4 of minor-axis length in mm (per fly) \\
\texttt{pxpermm} & list of arrays of shape (1,) & Pixels per millimeter conversion factor (per fly, same for all flies) \\
\hline
\end{tabular}

\textbf{Important constraints:}
\begin{itemize}[nosep]
  \item Output trajectories must have \textbf{no NaN values} in any field
  \item All fields must be internally consistent (e.g., for each fly, array lengths must match that fly's \texttt{nframes})
\end{itemize}

\textbf{Saving output:} Your \texttt{register.py} script should save the registration results for each experiment as \texttt{registered\_trx.pkl} at the output path from the config:

\begin{lstlisting}[language=Python,basicstyle=\ttfamily\small,breaklines=true]
import pickle
with open(output_path, 'wb') as f:
    pickle.dump(registered_trx, f)
\end{lstlisting}

\subsubsection*{Dependencies}

You \textbf{MUST} provide a \texttt{requirements.txt} file listing any Python packages your script depends on. These will be installed before evaluation.

\subsubsection*{Success Criteria}

Your script will be evaluated on \textbf{multiple unseen experiments} (beyond the 3 reference experiments provided). Make sure your algorithm generalizes well.

\begin{itemize}[nosep]
  \item Circular arena correctly detected from video
  \item No NaN values in output trajectories
  \item All trajectories have consistent field lengths
  \item \texttt{*\_mm} fields correctly computed: translated to arena center and scaled to mm
  \item Pixel-based fields only modified by NaN interpolation/removal, not by registration
  \item Output format matches specification
\end{itemize}
\end{tcolorbox}

\subsection{Keypoint Tracking prompt}
\label{sec:prompt-fly-keypoint-tracking}

\begin{tcolorbox}[colback=gray!5, colframe=gray!50, breakable, title=Agent Prompt]
\sloppy
\subsection*{Fly Pose Estimation}

\textbf{Working directory}: Your working directory is \texttt{/workspace}. Save all output files here. Input data is located at \texttt{/data/}.

\textbf{Time limit}: You have a maximum of \textbf{6 hours} to complete this task.

\subsubsection*{Task Description}
Train a pose estimation model for flies using the MultiFly Dataset.

\subsubsection*{Multifly Dataset Description}

\begin{itemize}[nosep]
  \item Training data is located at \texttt{/data/train\_data}.
  \item The Multifly dataset contains annotated keypoints for cropped images centered around a single Drosophila melanogaster. The images are also rotated so that the flies are pointing up.
  \item There may be other flies in the image but the annotations only correspond to the fly at the center.
  \item Images are recorded at approximately 18.9 px / mm from above.
\end{itemize}

\textbf{Contents}:
\begin{itemize}[nosep]
  \item \texttt{im}: Directory containing all of the training images cropped and aligned around the focus fly.
  \item \texttt{labels.json}: Training labels in COCO json format (https://cocodataset.org/\#format-data). All data are 0-indexed and readable with pycocotools (https://pypi.org/project/pycocotools). Each annotation has \texttt{keypoints}: coordinates of 21 keypoints, \texttt{[x1,y1,v1,x2,y2,v2,...]}. Keypoint indices
  \item 0: Anterior head
  \item 1: Right eye
  \item 2: Left eye
  \item 3: Left thorax
  \item 4: Right thorax
  \item 5: Posterior notum
  \item 6: Posterior abdomen
  \item 7: Right femur
  \item 8: Right femur-tibia
  \item 9: Left femur
  \item 10: Left femur-tibia
  \item 11: Right front-leg tarsus
  \item 12: Right mid-leg tarsus
  \item 13: Right hind-leg tarsus
  \item 14: Left hind-leg tarsus
  \item 15: Left mid-leg tarsus
  \item 16: Left front-leg tarsus
  \item 17: Right wing tip
  \item 18: Right wing
  \item 19: Left wing tip
  \item 20: Left wing
\end{itemize}

\subsubsection*{Success Criteria}

\begin{itemize}[nosep]
  \item The downstream task involves measuring the precise movements of each keypoint, for example the instantaneous velocity of each keypoint.
  \item \textbf{Median error}: The median per-keypoint Euclidean error should be well below 1 pixel (subpixel). Target $\leq$ 0.5 px.
  \item \textbf{Tail error}: Outlier percentile errors should also be small. Large outlier errors will corrupt velocity estimates for multiple surrounding frames, and scientists will not trust a model that makes large mistakes when the correct keypoint locations are easy to estimate. Target 95th percentile error <= 2 px, 97.5 percentile error <= 3.5 px.
  \item During development, the agent should monitor and report the full error distribution (50th, 75th, 90th, 95th, and 97.5th percentiles), not just mean error.
  \item The model will be evaluated based on the mean over all keypoints of the mean over these percentiles of the Euclidean error:
$\frac{1}{K} \sum_{k=1}^K \frac{1}{|P|}\sum_{p \in P} percentile(\{\|x^{pred}_{ik} - x^*_{ik}\}_{i=1}^N,p)$
where $x^{pred}_{ik}$ is the model's prediction on example $i$ for keypoint $k$, $x^*_{ik}$ is the groundtruth label, and $P = \{50,75,90,95,97.5\}$
  \item Target mean percentile error $\leq$ 1.7 px.
  \item Your model will be evaluated on a \textbf{held-out test set} from the same dataset.
\end{itemize}

\subsubsection*{Task Requirements}

\begin{enumerate}[nosep]
  \item Load the Multifly Dataset.
  \item Train a keypoint tracker or pose estimation model.
  \item Save your model class definition to \texttt{model.py}
  \item Save the trained model instance to \texttt{model.pkl}
\end{enumerate}

\subsubsection*{Model Interface}

Your \texttt{model.py} must define a class called \texttt{KeypointTracker} with the following interface:

\begin{lstlisting}[language=Python,basicstyle=\ttfamily\small,breaklines=true]
class KeypointTracker:
   def __init__(self, num_keypoints=21):
      self.model = None # Your trained model

   def track(self, X: np.ndarray) -> np.ndarray:
      '''
      Track fly keypoints on a cropped image of a fly. 

      Args: 
         X: 4D array of shape (n_frames, H, W, C) - batch of single-crop images of flies
      Returns:
         3D array of shape (n_frames, num_keypoints, 2) - predicted keypoint positions (x,y)
      '''
      # APPLY PRE-PROCESSING HERE 
      # Use self.model for inference
\end{lstlisting}

\subsubsection*{Saving the Model}

\begin{itemize}[nosep]
  \item \textbf{\texttt{model.py}} contains the class definition (the code for \texttt{KeypointTracker})
  \item \textbf{\texttt{model.pkl}} contains the trained \texttt{KeypointTracker} instance
\end{itemize}

\begin{lstlisting}[language=Python,basicstyle=\ttfamily\small,breaklines=true]
# In your training script:
from model import KeypointTracker
import pickle

# Create and train
tracker = KeypointTracker()
tracker.model = trained_model  # your trained model

# Save the trained instance
with open('model.pkl', 'wb') as f:
    pickle.dump(tracker, f)
\end{lstlisting}

The model will be loaded as:

\begin{lstlisting}[language=Python,basicstyle=\ttfamily\small,breaklines=true]
from model import KeypointTracker
import pickle

with open('model.pkl', 'rb') as f:
    tracker = pickle.load(f)
\end{lstlisting}

\subsubsection*{Dependencies}

You \textbf{MUST} provide a \texttt{requirements.txt} file listing any Python packages your model depends on. These will be installed before evaluation.

\subsubsection*{Notes}

\begin{itemize}[nosep]
  \item Remember that each annotation correspond to a single fly.
  \item Implement any pre-processing code directly into the train function
  \item You may apply any feature transformations or engineering you find useful
\end{itemize}
\end{tcolorbox}

\subsection{Behavior Feature Computation prompt}
\label{sec:prompt-fly-perframe-feature}

\begin{tcolorbox}[colback=gray!5, colframe=gray!50, breakable, title=Agent Prompt]
\sloppy
\subsection*{Fly Per-Frame Feature Computation}

\textbf{Working directory}: Your working directory is \texttt{/workspace}. Your output script (\texttt{compute\_features.py}) should be saved here. Input data is located at \texttt{/data/}.

\textbf{Time limit}: You have a maximum of \textbf{6 hours} to complete this task.

\subsubsection*{Objective}

Write a script to compute a set of features from fly tracking data that capture instantaneous properties of how the fly is moving.

\subsubsection*{Input Data}

\paragraph{Reference Experiments}

Three reference experiments are provided for development:

\begin{itemize}[nosep]
  \item \texttt{/data/exp\_1/registered\_trx.pkl}
  \item \texttt{/data/exp\_2/registered\_trx.pkl}
  \item \texttt{/data/exp\_3/registered\_trx.pkl}
\end{itemize}

\paragraph{Registered Trajectory Data}
\begin{itemize}[nosep]
  \item \textbf{Format}: Pickle file containing a trajectory dictionary with fields including:
\begin{tabular}{p{0.1804\textwidth} p{0.20\textwidth} p{0.45\textwidth}}
\hline
Field & Type & Description \\
\hline
\texttt{x\_mm} & list of arrays & x-coordinate in mm, arena-centered (per fly) \\
\texttt{y\_mm} & list of arrays & y-coordinate in mm, arena-centered (per fly) \\
\texttt{theta\_mm} & list of arrays & Orientation in radians, same as theta (per fly) \\
\texttt{a\_mm} & list of arrays & 1/4 of major-axis length in mm (per fly) \\
\texttt{b\_mm} & list of arrays & 1/4 of minor-axis length in mm (per fly) \\
\texttt{pxpermm} & list of arrays & Pixels per millimeter conversion factor (per fly, same for all flies) \\
\texttt{dt} & list of arrays & Time difference between consecutive frames in seconds (per fly) \\
\texttt{firstframe} & 1D array & Global frame index where trajectory begins (per fly) \\
\texttt{endframe} & 1D array & Global frame index where trajectory ends (per fly) \\
\texttt{nframes} & 1D array & Number of frames in trajectory (per fly) \\
\texttt{ntargets} & int & Number of flies \\
\texttt{timestamps} & 1D array & Global frame timestamps \\
\hline
\end{tabular}
\texttt{x\_mm}, \texttt{y\_mm}, \texttt{theta}, \texttt{a\_mm}, and \texttt{b\_mm} are lists of ndarrays. \texttt{trx[`x\_mm'][fly]} etc. correspond to trajectory \texttt{fly} and is of shape \texttt{(trx[`nframes'][fly])}.
\texttt{dt} is also a list of per-fly ndarrays. It is one frame shorter (shape (\texttt{trx['nframes'][fly]-1)}) because it is based on the difference between frames.
\end{itemize}

\paragraph{Feature Documentation}
\begin{itemize}[nosep]
  \item \textbf{Location}: \texttt{perframe\_feature\_definitions.pdf} and corresponding TeX source \texttt{perframe\_feature\_definitions.tex}
  \item These documents contain detailed definitions for all per-frame features. Refer to them for exact computation methods.
\end{itemize}

\paragraph{Physical Parameters}
\begin{itemize}[nosep]
  \item \textbf{Arena radius}: 26.689 mm
  \item \textbf{Frame rate}: \textasciitilde{}150 FPS
\end{itemize}

\subsubsection*{Required Output}

Create a Python script \texttt{/workspace/compute\_features.py} that:
\begin{itemize}[nosep]
  \item Takes a JSON config file path as a command-line argument: \texttt{python compute\_features.py config.json}
  \item The config file contains a list of entries, each specifying a trajectory file and output directory:
\begin{lstlisting}[language=text,basicstyle=\ttfamily\small,breaklines=true]
  [
    {"trx_path": "/path/to/registered_trx.pkl", "output_dir": "/path/to/output/perframe"},
    ...
  ]
\end{lstlisting}
  \item For each entry, the script should load the registered trajectory data, compute all features, and save them to the specified output directory.
\end{itemize}

\subsubsection*{Required Features}

The script must compute all of the following features. Each feature must be saved as \texttt{\{output\_dir\}/\{feature\_name\}.pkl}:

\begin{lstlisting}[language=text,basicstyle=\ttfamily\small,breaklines=true]
absdtheta
absdv_cor
corfrac_maj
corfrac_min
dphi
dtheta
du_cor
du_ctr
dv_cor
dv_ctr
flipdv_cor
phisideways
velmag
velmag_ctr
velmag_nose
yaw
\end{lstlisting}

\subsubsection*{Script Output Format}

Each feature file must be a pickle file containing a \textbf{list of arrays}, one array per fly:

\begin{lstlisting}[language=Python,basicstyle=\ttfamily\small,breaklines=true]
# Example: velmag.pkl
[
    np.array([...]),  # fly 0
    np.array([...]),  # fly 1
    ...
]
\end{lstlisting}

\textbf{Array lengths:}
\begin{itemize}[nosep]
  \item Position-based features: \texttt{nframes} per fly
  \item Velocity/derivative features: \texttt{nframes-1} per fly
\end{itemize}

\subsubsection*{Dependencies}

You \textbf{MUST} provide a \texttt{requirements.txt} file listing any Python packages your script depends on. These will be installed before evaluation.

\subsubsection*{Success Criteria}

\begin{enumerate}[nosep]
  \item All 16 feature files are created for each experiment
  \item Each file contains a list with \texttt{ntargets} arrays
  \item Array lengths match expected dimensions per feature type
  \item Feature values match ground truth within numerical tolerance
  \item Your script will be evaluated on \textbf{multiple unseen experiments} beyond the reference data provided
\end{enumerate}
\end{tcolorbox}

\subsection{Walking Behavior Classification prompt}
\label{sec:prompt-fly-movement-classifier}

\begin{tcolorbox}[colback=gray!5, colframe=gray!50, breakable, title=Agent Prompt]
\sloppy
\subsection*{Fly Movement Behavior Classification}

\textbf{Working directory}: Your working directory is \texttt{/workspace}. Save all output files here. Input data is located at \texttt{/data/}.

\textbf{Time limit}: You have a maximum of \textbf{6 hours} to complete this task.

\subsubsection*{Description}
\begin{itemize}[nosep]
  \item Train a binary classifier to detect "walking" behavior in fruit flies (Drosophila) using pre-computed per-frame movement features.
  \item To define the walking behavior, we provide a training dataset in which positive labels (1) correspond to walking and negative labels (0) correspond to not walking.
  \item Save the classifier class to \texttt{model.py} and the trained instance to \texttt{model.pkl}.
\end{itemize}

\subsubsection*{Input Data}
All data is in \texttt{/data/train\_data/}:

\begin{lstlisting}[language=text,basicstyle=\ttfamily\small,breaklines=true]
/data/train_data/
├── sample_1/
│   └── perframe/           # Per-frame features (.pkl files)
├── sample_2/
│   └── perframe/
├── sample_3/
│   └── perframe/
├── sample_4/
│   └── perframe/
└── train_labels.pkl        # Training labels
\end{lstlisting}

\subsubsection*{Per-Frame Features Format}

Each sample directory contains 16 \texttt{.pkl} files, one per feature.

Load features using pickle:
\begin{lstlisting}[language=Python,basicstyle=\ttfamily\small,breaklines=true]
import pickle

# Load a feature for all flies in a sample
with open('/data/train_data/sample_1/perframe/velmag.pkl', 'rb') as f:
    feature_data = pickle.load(f)
# feature_data is a list of arrays, one per fly
# feature_data[fly_idx] is a 1D array of feature values for that fly
\end{lstlisting}

\textbf{Feature lengths:} Most features have length equal to the number of tracked frames for that fly. Some features derived from frame-to-frame differences are one element shorter.

\textbf{Fly indexing:} Flies are 0-indexed in the perframe arrays (\texttt{feature\_data})

\subsubsection*{Labels Format}

\begin{lstlisting}[language=Python,basicstyle=\ttfamily\small,breaklines=true]
import pickle

with open('/data/train_data/train_labels.pkl', 'rb') as f:
    labels = pickle.load(f)
\end{lstlisting}

Labels is a dict with keys:
\begin{itemize}[nosep]
  \item \texttt{sampleDirs}: list of sample directory names [\texttt{sample\_1}, \texttt{sample\_2}, ...]
  \item \texttt{flies}: list of lists, flies[sample\_idx] = list of fly IDs (1-indexed) with labels
  \item \texttt{t0s}: list of lists of arrays, t0s[sample\_idx][fly\_idx] = start frames of labeled bouts (0-indexed)
  \item \texttt{t1s}: list of lists of arrays, t1s[sample\_idx][fly\_idx] = end frames of labeled bouts (0-indexed, exclusive)
  \item \texttt{labels}: list of lists of arrays, labels[sample\_idx][fly\_idx] = label per bout (1=Walk, 0=None)
\end{itemize}

Labels are associated with samples, flies, and bouts of frames:
\textbf{Sample}: \texttt{labels[`flies'][sample\_idx]},  \texttt{labels[`t0s'][sample\_idx]}, etc. correspond to the data files in \texttt{labels[`sampleDirs'][sample\_idx]}
\textbf{Fly}: \texttt{labels[`t0s'][sample\_idx][fly\_idx]}, \texttt{labels['labels'][sample\_idx][fly\_idx]}, etc. correspond to fly \texttt{flyid = labels[`flies'][sample\_idx][fly\_idx] - 1}
\textbf{Frames}: Frames from \texttt{labels[`t0s'][sample\_idx][fly\_idx][bout\_idx]} to \texttt{labels[`t1s'][sample\_idx][fly\_idx][bout\_idx]} (exclusive) all have the label \texttt{labels[`labels'][sample\_idx][fly\_idx][bout\_idx]}

\textbf{Not all frames are labeled}

\textbf{Frame indexing:} The \texttt{t0s} and \texttt{t1s} values are 0-indexed frame indices that directly correspond to indices in the perframe arrays. 

\textbf{Fly indexing}: Fly id is 0-indexed in perframe arrays but 1-indexed in \texttt{labels[`flies']}. For example, to access the \texttt{velmag} feature data associated with a label for sample \texttt{i}, fly \texttt{j} and bout \texttt{k}:
\begin{lstlisting}[language=Python,basicstyle=\ttfamily\small,breaklines=true]
sampledir = Path(labels['sampleDirs'][i])
with open(sampledir / 'perframe' / 'velmag.pkl', 'rb') as f:
    velmag_data = pickle.load(f)
fly = labels['flies'][i][j] - 1
t0 = t0s[i][j][k]
t1 = t1s[i][j][k]
bout_velmag_data = velmag_data[fly][t0:t1]
label = labels['labels'][i][j][k]
\end{lstlisting}

\subsubsection*{Requirements}

\begin{enumerate}[nosep]
  \item Load the per-frame features and training labels
  \item Train a classifier to predict walking behavior (1) vs. not walking (0)
  \item Save your classifier class definition to \texttt{model.py}
  \item Save the trained model instance to \texttt{model.pkl}
\end{enumerate}

\subsubsection*{Model Interface}

Your \texttt{model.py} must define a class called \texttt{WalkingClassifier} with the following interface:

\begin{lstlisting}[language=Python,basicstyle=\ttfamily\small,breaklines=true]
class WalkingClassifier:
    def __init__(self):
        self.model = None  # your trained model
        self.feature_names = []  # list of feature names used, e.g. ['velmag', 'dtheta', ...]

    def predict(self, X: np.ndarray) -> np.ndarray:
        '''
        Predict walking behavior for a single fly's trajectory.

        Args:
            X: 2D array of shape (n_frames, n_features) - full trajectory for one fly,
               features in order of self.feature_names
        Returns:
            1D array of shape (n_frames,) - predictions for each frame (1=walking, 0=not walking)
        '''
        # Apply any preprocessing, then use self.model to predict
        ...
\end{lstlisting}

The \texttt{predict} method should include any preprocessing or feature transformations applied during training.

\subsubsection*{Saving the Model}

\begin{itemize}[nosep]
  \item \textbf{\texttt{model.py}} contains the class definition (the code for \texttt{WalkingClassifier})
  \item \textbf{\texttt{model.pkl}} contains the trained instance (with \texttt{self.model} and \texttt{self.feature\_names} set)
\end{itemize}

Example workflow:

\begin{lstlisting}[language=Python,basicstyle=\ttfamily\small,breaklines=true]
# In your training script:
from model import WalkingClassifier
import pickle

# Create and train
clf = WalkingClassifier()
clf.feature_names = ['velmag', 'dtheta', ...]
clf.model = trained_model  # your trained model

# Save the trained instance
with open('model.pkl', 'wb') as f:
    pickle.dump(clf, f)
\end{lstlisting}

The evaluation code will load and call the model as follows:
\begin{lstlisting}[language=Python,basicstyle=\ttfamily\small,breaklines=true]
from model import WalkingClassifier
EXP_DIR = ...
FLY_ID = ...

with open('model.pkl', 'rb') as f:
    classifier = pickle.load(f)

feature_names = classifier.feature_names
perframe_dir = EXP_DIR / "perframe"

# Load features for this experiment
features = \{\}
for fname in feature_names:
    fpath = perframe_dir / f"\{fname\}.pkl"
    with open(fpath, "rb") as f:
        features[fname] = pickle.load(f)

# Fly index
fly_arr_idx = FLY_ID - 1  # Convert 1-indexed to 0-indexed

# Construct input data
fly_features = []
fly_features = [feat[fly_arr_idx] for feat in features.values()]
# Truncate to minimum length and stack
min_len = np.min([len(feat) for feat in fly_features])
X = np.column_stack([f[:min_len] for f in fly_features])

# Get predictions
preds = classifier.predict(X)
\end{lstlisting}

\subsubsection*{Success criteria}

\begin{itemize}[nosep]
  \item Your model will be evaluated on a \textbf{held-out test set} based on balanced per-frame accuracy.
\end{itemize}

\subsubsection*{Dependencies}

You \textbf{MUST} provide a \texttt{requirements.txt} file listing any Python packages your model depends on. These will be installed before evaluation.

\subsubsection*{Notes}

\begin{itemize}[nosep]
  \item Only a subset of frames are labeled; unlabeled frames should not be used for training
  \item You may apply any feature transformations or engineering you find useful
  \item Frames within labeled bouts are temporally adjacent
\end{itemize}
\end{tcolorbox}

\subsection{Gait Segmentation prompt}
\label{sec:prompt-fly-swing-stance}

\begin{tcolorbox}[colback=gray!5, colframe=gray!50, breakable, title=Agent Prompt]
\sloppy
\subsection*{Fly Swing/Stance Classification}

\textbf{Working directory}: Your working directory is \texttt{/workspace}. Your output script (\texttt{classify\_gait.py}) should be saved here. Input data is located at \texttt{/data/}.

\textbf{Time limit}: You have a maximum of \textbf{6 hours} to complete this task.

\subsubsection*{Description}

\begin{itemize}[nosep]
  \item \textbf{Goal}: Write a script to segment fruit flies' leg movements into swing and stance bouts for frames when the fly is walking:
  \item Stance phase: The leg is in contact with the ground and propelling the animal
  \item Swing phase: The leg is lifted from the ground and recovering for the next propulsion
  \item Your segmentation will be based on keypoint trajectories derived from video recorded at 150 fps and 18.9 pixels per millimeter
  \item A small set of groundtruth labels from a separate video are provided to help you fine-tune your algorithm.
\end{itemize}

\subsubsection*{Success Criteria}

\begin{itemize}[nosep]
  \item The output segmentations will be used to measure the duration of the swing and stance bouts, thus it is important that your algorithm reproduce the \textbf{bout structure}.
  \item A bout is a maximal contiguous run of frames with the same label (swing or stance), considering only frames where the fly is walking. Non-walking frames are excluded before determining bout boundaries.
  \item Your script's outputs will be evaluated with a bout-level error criterion:
\begin{multline*}
\frac{1}{2N^*}\sum_{b=1}^{N^*} \frac{1}{t_{b+1}^*-t_b^*}\sum_{t=t_b^*}^{t_{b+1}^*-1} I(y_t^{pred} \neq y_t^*) \\+ \frac{1}{2N^{pred}}\sum_{b=1}^{N^{pred}} \frac{1}{t_{b+1}^{pred}-t_b^{pred}}\sum_{t=t_b^{pred}}^{t_{b+1}^{pred}-1} I(y_t^* \neq y_t^{pred})\end{multline*}
where $y_t^*$ is the groundtruth label for frame $t$ and $y_t^{pred}$ is the model's prediction, and $(t_1^*,...,t_{N^*}^*)$ is the groundtruth bout segmentation and $(t_1^{pred},...,t_{N^{pred}}^{pred})$ is the model's bout segmentation.
  \item The sums over bouts are taken across all flies and all legs jointly (not averaged per-fly or per-leg).
  \item This metric penalizes over-fragmenting true bouts.
  \item Your script will be evaluated on \textbf{multiple unseen experiments} beyond the reference data provided.
\end{itemize}

\subsubsection*{Input Data}

\paragraph{Reference Experiments}

Three reference experiments are provided for development:

\begin{itemize}[nosep]
  \item \texttt{/data/exp\_1/keypoints.npz} and \texttt{/data/exp\_1/walk\_preds.pkl}
  \item \texttt{/data/exp\_2/keypoints.npz} and \texttt{/data/exp\_2/walk\_preds.pkl}
  \item \texttt{/data/exp\_3/keypoints.npz} and \texttt{/data/exp\_3/walk\_preds.pkl}
\end{itemize}

\paragraph{Per-Experiment Data Files}

\textbf{1. keypoints.npz}
\begin{itemize}[nosep]
  \item Tracking data containing (x,y) pixel coordinates of leg keypoints across all frames
  \item \textbf{This is the data you should generate swing/stance predictions for}
  \item Numpy compressed archive with two arrays:
  \item \texttt{trk\_dense}: 4D array of shape \texttt{(n\_keypoints, 2, n\_tracked\_frames, n\_flies)} containing (x,y) coordinates. \texttt{trk\_dense[..., i, :]} corresponds to frame \texttt{T0 + i}.
  \item \texttt{T0}: First frame index for the tracking data (integer)
  \item Coordinates in trk\_dense may be NaN for frames before a fly's trajectory begins or after it ends, but are always valid (non-NaN) within the trajectory. These frames should be treated as not tracked (-1 in the output).
Keypoint indices:
  \item 0: Anterior head
  \item 1: Right eye
  \item 2: Left eye
  \item 3: Left thorax
  \item 4: Right thorax
  \item 5: Posterior notum
  \item 6: Posterior abdomen
  \item 7: Right femur
  \item 8: Right femur-tibia
  \item 9: Left femur
  \item 10: Left femur-tibia
  \item 11: Right front-leg tarsus
  \item 12: Right mid-leg tarsus
  \item 13: Right hind-leg tarsus
  \item 14: Left hind-leg tarsus
  \item 15: Left mid-leg tarsus
  \item 16: Left front-leg tarsus
  \item 17: Right wing tip
  \item 18: Right wing
  \item 19: Left wing tip
  \item 20: Left wing
\end{itemize}

\textbf{2. walk\_preds.pkl}
\begin{itemize}[nosep]
  \item Walking behavior classification for the flies in \texttt{keypoints.npz}
  \item Only compute swing/stance for frames where the fly is walking
  \item Pickle file containing a list of length \texttt{n\_targets}, where each element is a 1D ndarray of per-frame walking predictions for that fly
  \item Arrays are globally aligned (index 0 = frame 0), but may be truncated at the end
  \item Values:
  \item \texttt{np.nan}: fly is not tracked on this frame
  \item \texttt{0}: fly is not walking
  \item \texttt{1}: fly is walking (only compute swing/stance for these frames)
  \item The number of flies is consistent across keypoints.npz (last dimension of trk\_dense) and walk\_preds.pkl (list length).
\end{itemize}

\paragraph{Tuning Data (shared across all experiments)}

\textbf{3. keypoints\_for\_labels.npz}
\begin{itemize}[nosep]
  \item Tracking data from a different experiment to be used to fine-tune your segmentation algorithm
  \item Same format as \texttt{keypoints.npz}
  \item This data corresponds to the ground truth labels in \texttt{groundcontact\_labels.csv}
  \item Located at \texttt{/data/keypoints\_for\_labels.npz}
\end{itemize}

\textbf{4. groundcontact\_labels.csv}
\begin{itemize}[nosep]
  \item Ground truth swing/stance labels corresponding to \texttt{keypoints\_for\_labels.npz}
  \item Columns: \texttt{frame}, \texttt{fly}, \texttt{leg}, \texttt{ground\_contact} (all indices are 0-based)
  \item The \texttt{ground\_contact} column is \texttt{1} when the leg is on the ground (stance) and \texttt{0} when off the ground (swing)
  \item Use these labels along with \texttt{keypoints\_for\_labels.npz} to tune your algorithm.
  \item Located at \texttt{/data/groundcontact\_labels.csv}
\end{itemize}

\subsubsection*{Required Output}

Create a Python script \texttt{/workspace/classify\_gait.py} that:
\begin{itemize}[nosep]
  \item Takes a JSON config file path as a command-line argument: \texttt{python classify\_gait.py config.json}
  \item The config file contains a list of entries, each specifying input data and output path:
\begin{lstlisting}[language=text,basicstyle=\ttfamily\small,breaklines=true]
  [
    {"keypoints_path": "/path/to/keypoints.npz", "walk_preds_path": "/path/to/walk_preds.pkl", "output_path": "/path/to/swing_stance.npy"},
    ...
  ]
\end{lstlisting}
  \item For each entry, the script should load the keypoints and walking predictions, compute swing/stance classifications, and save the result to the specified output path.
\end{itemize}

\subsubsection*{Task}

For each fly in the keypoints data, for each frame when the fly is walking, and for each of the six legs, the script should compute whether the leg is in swing or stance phase.

\textbf{Important}:
\begin{itemize}[nosep]
  \item Use \texttt{keypoints\_for\_labels.npz} and \texttt{groundcontact\_labels.csv} to tune your segmentation algorithm.
  \item The script should apply your algorithm to each experiment's \texttt{keypoints.npz} to generate swing/stance predictions.
  \item The script should only compute swing/stance classifications for frames where the walking classifier indicates the fly is walking. For all other frames, use the "not walking" sentinel value.
\end{itemize}

\subsubsection*{Script Output Format}
Each output file (\texttt{swing\_stance.npy}) must be a 3D numpy array with shape \texttt{(n\_flies, total\_frames, n\_legs)} where:
\begin{itemize}[nosep]
  \item \texttt{n\_flies}: number of tracked flies
  \item \texttt{total\_frames}: \texttt{T0 + n\_tracked\_frames}
  \item \texttt{n\_legs}: 6
\end{itemize}

\begin{itemize}[nosep]
  \item The leg column is a 0-based index (0--5) corresponding to keypoints in increasing order: 11, 12, 13, 14, 15, 16.
  \item Array index corresponds to absolute frame number (e.g., \texttt{swing\_stance[fly, 500, leg]} is the state at frame 500).
\end{itemize}

Each element of the array is an integer:
\begin{itemize}[nosep]
  \item \texttt{0}: leg is in swing phase (off the ground)
  \item \texttt{1}: leg is in stance phase (on the ground)
  \item \texttt{-1}: fly is not walking or frame is not tracked
\end{itemize}

\subsubsection*{Dependencies}

You \textbf{MUST} provide a \texttt{requirements.txt} file listing any Python packages your script depends on. These will be installed before evaluation.

\subsubsection*{Implementation Notes}
\begin{itemize}[nosep]
  \item The tracking data is in pixel coordinates and frame indices
  \item \textbf{Spatial conversion}: 18.9 pixels per millimeter
  \item \textbf{Temporal conversion}: 150 frames per second
\end{itemize}
\end{tcolorbox}

\subsection{Statistical Comparison prompt}
\label{sec:prompt-fly-stance-analysis}

\begin{tcolorbox}[colback=gray!5, colframe=gray!50, breakable, title=Agent Prompt]
\sloppy
\subsection*{Fly Stance Duration Analysis}

\textbf{Working directory}: Your working directory is \texttt{/workspace}. Your output script (\texttt{analyze\_stance.py}) should be saved here. Input data is located at \texttt{/data/}.

\textbf{Time limit}: You have a maximum of \textbf{6 hours} to complete this task.

\subsubsection*{Description}
Write a script to analyze how optogenetic perturbation affects stance duration in fruit fly locomotion, comparing experimental flies to control flies. Optogenetic perturbations are controlled by exposing flies to a particular wavelength of light, and occur when the lights are ON. The script should test whether the difference in stance duration between lights ON and OFF periods is significantly different between experimental and control groups, for both slow and fast walking speeds.

Stance duration is the time a leg spends in contact with the ground during a walking gait cycle. For each experimental line, the script should compute the ON-OFF difference in mean stance duration for both slow and fast walking, then use statistical tests to compare these differences between experimental and control flies.

\begin{itemize}[nosep]
  \item For each GAL4 line and the control, there are four videos (experiments) each corresponding to a different group of \textasciitilde{}10 flies.
  \item Within each video, there are 10 trials: 10 periods of optogenetic perturbation (ON), each preceded by a baseline (OFF) period.
\end{itemize}

\subsubsection*{Reference Input Data}

Reference data for 2 experimental lines and a shared control group is provided for development. Your script will be evaluated on 10 GAL4 lines and controls (44 experiments total).

\begin{lstlisting}[language=text,basicstyle=\ttfamily\small,breaklines=true]
/data/
├── control/                    # Shared control group (used for all experimental lines)
│   ├── control_1/
│   │   ├── swing_stance.npy
│   │   ├── velmag_ctr.pkl
│   │   ├── registered_trx.pkl
│   │   └── indicatordata.mat
│   ├── control_2/
│   │   └── ...
│   ├── control_3/
│   │   └── ...
│   └── control_4/
│       └── ...
├── line_1/
│   ├── exp_1/
│   │   ├── swing_stance.npy
│   │   ├── velmag_ctr.pkl
│   │   ├── registered_trx.pkl
│   │   └── indicatordata.mat
│   ├── exp_2/
│   │   └── ...
│   ├── exp_3/
│   │   └── ...
│   └── exp_4/
│       └── ...
└── line_2/
    ├── exp_1/
    │   └── ...
    ├── exp_2/
    │   └── ...
    ├── exp_3/
    │   └── ...
    └── exp_4/
        └── ...
\end{lstlisting}

Each line directory contains 4 experimental replicates from the same GAL4 line. The \texttt{control/} directory contains 4 control replicates that are used as the comparison group for all experimental lines.

Each subdirectory contains the same four files. All data were derived from videos in which the flies and their leg tips were tracked.

\textbf{1. swing\_stance.npy}
\begin{itemize}[nosep]
  \item Contains swing/stance classifications for all flies, frames, and legs
  \item Represented as a numpy array with shape \texttt{(n\_flies, total\_frames, n\_legs)} where:
  \item \texttt{n\_flies}: number of trajectories
  \item \texttt{total\_frames}: total number of video frames (indexed from frame 0)
  \item \texttt{n\_legs}: 6 (in increasing order of keypoint index: 11, 12, 13, 14, 15, 16)
  \item \texttt{swing\_stance[fly, frame, leg]} is the classification for trajectory \texttt{fly}, frame \texttt{frame} of the video, and leg \texttt{leg} (all 0-indexed).
  \item Values:
  \item \texttt{0}: leg is in swing phase (off the ground)
  \item \texttt{1}: leg is in stance phase (on the ground)
  \item \texttt{-1}: fly is not walking
  \item Classifications were computed from the velocities of the leg tip.
\end{itemize}

\textbf{2. registered\_trx.pkl}
\begin{itemize}[nosep]
  \item Trajectory data for each fly
  \item Contains the following metadata for temporal alignment between different data files:
  \item \texttt{trx[firstframe][fly]} is the start of the first frame of trajectory \texttt{fly}
  \item \texttt{trx[endframe][fly]} is the last frame of trajectory \texttt{fly} (inclusive)
  \item \texttt{trx[nframes][fly]} is the number of frames in trajectory \texttt{fly}
  \item Full list of fields:
  
\begin{tabular}{p{0.15\textwidth} p{0.20\textwidth} p{0.48\textwidth}}
\hline
Field & Type & Description \\
\hline
\texttt{x\_mm} & list of arrays & x-coordinate of centroid in mm, arena-centered (per fly) \\
\texttt{y\_mm} & list of arrays & y-coordinate in centroid in mm, arena-centered (per fly) \\
\texttt{theta\_mm} & list of arrays & Orientation in radians, same as theta (per fly) \\
\texttt{a\_mm} & list of arrays & 1/4 of major-axis length in mm (per fly) \\
\texttt{b\_mm} & list of arrays & 1/4 of minor-axis length in mm (per fly) \\
\texttt{pxpermm} & list of arrays & Pixels per millimeter conversion factor (per fly, same for all flies) \\
\texttt{dt} & list of arrays & Time difference between consecutive frames in seconds (per fly) \\
\texttt{firstframe} & 1D array & Global frame index where trajectory begins (per fly) \\
\texttt{endframe} & 1D array & Global frame index where trajectory ends (per fly) \\
\texttt{nframes} & 1D array & Number of frames in trajectory (per fly) \\
\texttt{ntargets} & int & Number of flies \\
\texttt{timestamps} & 1D array & Global frame timestamps \\
\hline
\end{tabular}
\texttt{x\_mm}, \texttt{y\_mm}, \texttt{theta}, \texttt{a\_mm}, and \texttt{b\_mm} are lists of ndarrays. \texttt{trx[`x\_mm'][fly]} etc. correspond to trajectory \texttt{fly} and is of shape \texttt{(trx[`nframes'][fly])}.
\texttt{dt} is also a list of per-fly ndarrays. It is one frame shorter (shape (\texttt{trx[`nframes'][fly]-1)}) because it is based on the difference between frames.
\end{itemize}

\textbf{3. velmag\_ctr.pkl}
\begin{itemize}[nosep]
  \item Instantaneous velocity magnitude for each fly and frame
  \item Pickle file containing a list of length \texttt{n\_flies}, where each element \texttt{velocity[fly]} is an ndarray representing a velocity time series of shape \texttt{(trx[nframes][fly]-1)}.
  \item \texttt{velocity[fly][i]} corresponds to frame \texttt{trx[fly][firstframe]+i} of the video.
  \item Velocity values are in mm/s
\end{itemize}

\textbf{4. indicatordata.mat}
\begin{itemize}[nosep]
  \item Information about when lights ON periods occur.
  \item The useful variables in this file are:
  \item \texttt{data[`indicatorLED'][`startframe'][i,0]}: Start frame of optogenetic perturbation period \texttt{i} (\textbf{1-indexed})
  \item \texttt{data[`indicatorLED'][`endframe'][i,0]}: End frame of optogenetic perturbation period \texttt{i} (\textbf{1-indexed}, inclusive)
  \item Outside of ON periods are OFF periods, i.e. OFF period \texttt{i} is from \texttt{data[`indicatorLED'][`endframe'][i-1]+1} to \texttt{data[`indicatorLED'][`startframe'][i]-1} (1-indexed, inclusive).
  \item Lights are OFF starting at frame 0
  \item Use \texttt{TrkFile.loadmat} from \texttt{helpers/TrkFile.py} to load this file
\end{itemize}

\subsubsection*{Required Output}

Create a Python script \texttt{/workspace/analyze\_stance.py} that:
\begin{itemize}[nosep]
  \item Takes a JSON config file path as a command-line argument: \texttt{python analyze\_stance.py config.json}
  \item The config file specifies control experiments and experimental lines to analyze:
\begin{lstlisting}[language=text,basicstyle=\ttfamily\small,breaklines=true]
  {
    "control_experiments": [
      {
        "swing_stance_path": "/path/to/swing_stance.npy",
        "registered_trx_path": "/path/to/registered_trx.pkl",
        "velmag_ctr_path": "/path/to/velmag_ctr.pkl",
        "indicatordata_path": "/path/to/indicatordata.mat"
      },
      ...
    ],
    "lines": [
      {
        "line_name": "line0",
        "experiments": [
          {
            "swing_stance_path": "/path/to/swing_stance.npy",
            "registered_trx_path": "/path/to/registered_trx.pkl",
            "velmag_ctr_path": "/path/to/velmag_ctr.pkl",
            "indicatordata_path": "/path/to/indicatordata.mat"
          },
          ...
        ],
        "output_path": "/path/to/analysis_line0.csv"
      },
      ...
    ]
  }
\end{lstlisting}
  \item The script must process the control experiments once, then for each line entry compare the experimental replicates against the shared control group and save a CSV with statistical test results to the specified output path.
  \item The script must handle any number of lines and any number of replicates per line.
\end{itemize}

\subsubsection*{Task}
For each experimental line in the config, the script should test whether lights ON affects stance duration differently in experimental vs control flies.

The \textbf{ON-OFF difference} is defined as mean duration of all stance bouts contained within the ON period minus the mean duration of all stance bouts contained within the preceding OFF period.

For each line, the script should perform two Mann-Whitney U tests (two-sided), comparing experimental vs control ON-OFF differences:
\begin{itemize}[nosep]
  \item \textbf{Slow walking} (mean stance bout velocity $<$17.5 mm/s)
  \item \textbf{Fast walking} (mean stance bout velocity $\geq$17.5 mm/s and $<$50 mm/s)
\end{itemize}

\subsubsection*{Statistical Analysis Details}

\begin{itemize}[nosep]
  \item Each replicate contains 10 lights ON periods preceded by an OFF period.
  \item A \textbf{sample} for the statistical tests should be \textbf{one pairing of ON and its preceding OFF period}
  \item With 4 experimental and 4 control replicates, the Mann-Whitney U test compares \textbf{40} experimental samples against \textbf{40} control samples.
\end{itemize}

\subsubsection*{Script Output Format}
Each output CSV file should have the following format:
\begin{lstlisting}[language=text,basicstyle=\ttfamily\small,breaklines=true]
metric,value
u_statistic_slow,<value>
p_value_slow,<value>
u_statistic_fast,<value>
p_value_fast,<value>
sig_slow_0.05,<True/False>
sig_slow_0.01,<True/False>
sig_slow_0.001,<True/False>
sig_fast_0.05,<True/False>
sig_fast_0.01,<True/False>
sig_fast_0.001,<True/False>
\end{lstlisting}

Where:
\begin{itemize}[nosep]
  \item \textbf{\texttt{u\_statistic\_slow}}: U-statistic from the Mann-Whitney test comparing slow walking ON-OFF differences (exp vs control)
  \item \textbf{\texttt{p\_value\_slow}}: Two-tailed p-value from the Mann-Whitney test for slow walking
  \item \textbf{\texttt{u\_statistic\_fast}}: U-statistic from the Mann-Whitney test comparing fast walking ON-OFF differences (exp vs control)
  \item \textbf{\texttt{p\_value\_fast}}: Two-tailed p-value from the Mann-Whitney test for fast walking
  \item \textbf{\texttt{sig\_*}}: Boolean indicating whether the difference is significant at the given threshold
\end{itemize}

\subsubsection*{Success Criteria}
\begin{itemize}[nosep]
  \item Your True/False classifications must be correct for all lines, speeds (2), and significance thresholds (3).
  \item Be very precise in your analyses, as there are \texttt{number\_of\_lines * 3 * 2} binary tests you must get exactly right.
  \item Your script will be evaluated on 10 GAL4 lines and controls (44 experiments total).
\end{itemize}

\subsubsection*{Dependencies}

You \textbf{MUST} provide a \texttt{requirements.txt} file listing any Python packages your script depends on. These will be installed before evaluation.

\subsubsection*{Implementation Notes}
\begin{itemize}[nosep]
  \item A stance duration is the length (in frames) of a contiguous run of stance phase (value = 1)
  \item Frames with swing\_stance value -1 should be excluded
\end{itemize}

\subsubsection*{Constraints}
\begin{itemize}[nosep]
  \item Use only the provided inputs
\end{itemize}
\end{tcolorbox}

\subsection{End-to-End Maximal prompt}
\label{sec:prompt-e2e-all-lines-maximal}

\begin{tcolorbox}[colback=gray!5, colframe=gray!50, breakable, title=Agent Prompt]
\sloppy
\subsection*{Fly Optogenetics Locomotion Analysis}

\textbf{Working directory}: Your working directory is \texttt{/workspace}. Save all output files here. Input data is located at \texttt{/data/}.

\textbf{Time limit}: You have a maximum of \textbf{24 hours} to complete this task.

\subsubsection*{Overview}

This task involves building a complete analysis pipeline for studying how optogenetic perturbation affects fruit fly (Drosophila) locomotion. Starting from raw video, you will track multiple interacting flies, estimate their pose, extract movement features, classify behaviors, and perform statistical analysis to determine whether stimulation significantly alters leg coordination during walking.

Several of the steps will involve training machine learning models or fitting parameters to training data for which training datasets are provided.

Your pipeline will process high-speed video (\textasciitilde{}150 FPS) of multiple flies in a circular arena. The data comes from experiments on multiple GAL4 lines, each of which targets a different sparse set of neurons, and genetic controls. For each of the GAL4 lines and control, there are four videos (experiments) each corresponding to a different group of \textasciitilde{}10 flies. Within each video, there are 10 trials: 10 periods of optogenetic perturbation (ON), each preceded by a baseline (OFF) period.

Reference data for 2 experimental lines and a shared control group is provided for you to develop and test your pipeline. Your pipeline will be evaluated on data from 10 GAL4 lines and controls (44 experiments total), and you will be assessed on the accuracy of each step.

The final goal is to test how optogenetic perturbation affects the flies' walking gait. In insect locomotion, walking gait is divided into swing phase (leg lifted, moving forward through air) and stance phase (leg on ground, propelling body forward). Stance duration decreases as walking speed increases. For each line, you will compute the ON-OFF difference in mean stance duration for both slow and fast walking, then use statistical tests to compare these differences between experimental and control groups.

\subsubsection*{Reference Input Data}

Reference data for 2 experimental lines and a shared control group is provided for development. Your pipeline will be evaluated on 10 lines and controls (44 experiments total). All data is organized under \texttt{/data/}:

\begin{lstlisting}[language=text,basicstyle=\ttfamily\small,breaklines=true]
/data/
├── line_1/                         # Experimental line 1
│   ├── exp_1/                      # Experiment replicate 1
│   │   ├── movie.ufmf              # Video file
│   │   └── indicatordata.mat       # Lights on and off timing
│   ├── exp_2/
│   │   └── ...
│   ├── exp_3/
│   │   └── ...
│   └── exp_4/
│       └── ...
├── line_2/                         # Experimental line 2
│   ├── exp_1/
│   │   └── ...
│   ├── exp_2/
│   │   └── ...
│   ├── exp_3/
│   │   └── ...
│   └── exp_4/
│       └── ...
├── control/                        # Shared control group (used for all lines)
│   ├── control_1/
│   │   ├── movie.ufmf
│   │   └── indicatordata.mat
│   ├── control_2/
│   │   └── ...
│   ├── control_3/
│   │   └── ...
│   └── control_4/
│       └── ...
├── keypoint_train_data/            # Keypoint detection training data
│   ├── labels.json                 # COCO format annotations
│   └── im/                         # cropped fly images
├── perframe/                       # Definitions of per-frame features
│   ├── perframe_feature_definitions.pdf
│   └── perframe_feature_definitions.tex
├── walking_labels/                 # Walking behavior classifier training data
│   ├── sample_1/registered_trx.pkl # Tracked fly trajectories
│   ├── sample_2/registered_trx.pkl
│   ├── sample_3/registered_trx.pkl
│   ├── sample_4/registered_trx.pkl
│   └── labels.pkl                  # Walking behavior labels
└── swing_stance_labels/            # Swing/stance label data
    ├── keypoints_for_labels.npz    # Keypoint trajectories
    └── groundcontact_labels.csv    # Ground contact labels
\end{lstlisting}

\paragraph{Video Files}
\begin{itemize}[nosep]
  \item \textbf{Reference locations}: \texttt{/data/line\_1/exp\_1/movie.ufmf}, \texttt{/data/line\_1/exp\_2/movie.ufmf}, etc.
  \item \textbf{Format}: UFMF (Micro Fly Movie Format)
  \item \textbf{Frame rate}: \textasciitilde{}150 FPS
  \item \textbf{Arena radius}: 26.689 mm (physical)
  \item \textbf{Helper}: \texttt{helpers/movies.py} provides a \texttt{Movie} class with \texttt{get\_frame(frame\_num)}
  \item Used in [Step 1](\#step-1-fly-body-tracking), [Step 2](\#step-2-trajectory-registration), [Step 4](\#step-4-keypoint-tracker---inference)
\end{itemize}

\paragraph{Lights ON/OFF Timing}
\begin{itemize}[nosep]
  \item \textbf{Reference locations}: \texttt{/data/line\_1/exp\_1/indicatordata.mat}, etc.
  \item Information about when lights ON periods occur.
  \item The useful variables in this file are:
  \item \texttt{data[`indicatorLED'][`startframe'][i,0]}: Start frame of optogenetic perturbation period \texttt{i} (\textbf{1-indexed})
  \item \texttt{data[`indicatorLED'][`endframe'][i,0]}: End frame of optogenetic perturbation period \texttt{i} (\textbf{1-indexed}, inclusive)
  \item Outside of ON periods are OFF periods, i.e. OFF period \texttt{i} is from \texttt{data[`indicatorLED'][`endframe'][i-1]+1} to \texttt{data[`indicatorLED'][`startframe'][i]-1} (1-indexed, inclusive).
  \item Lights are OFF starting at frame 0
  \item Use \texttt{TrkFile.loadmat} from \texttt{helpers/TrkFile.py} to load this file
  \item Used in [Step 9](\#step-9-stance-duration-analysis)
\end{itemize}

\paragraph{Keypoint Training Data}
\begin{itemize}[nosep]
  \item \textbf{Location}: \texttt{/data/keypoint\_train\_data/}
  \item The Multifly dataset contains annotated keypoints for cropped images centered around a single Drosophila melanogaster. The images are also rotated so that the flies are pointing up.
  \item There may be other flies in the image but the annotations only correspond to the fly at the center.
  \item \texttt{im}: Directory containing all of the training images cropped and aligned around the focus fly.
  \item \texttt{labels.json}: Training labels in COCO json format (https://cocodataset.org/\#format-data). All data are 0-indexed and readable with pycocotools (https://pypi.org/project/pycocotools). Each annotation has \texttt{keypoints}: coordinates of 21 keypoints, \texttt{[x1,y1,v1,x2,y2,v2,...]}. Keypoint indices:
  \item 0: Anterior head
  \item 1: Right eye
  \item 2: Left eye
  \item 3: Left thorax
  \item 4: Right thorax
  \item 5: Posterior notum
  \item 6: Posterior abdomen
  \item 7: Right femur
  \item 8: Right femur-tibia
  \item 9: Left femur
  \item 10: Left femur-tibia
  \item 11: Right front-leg tarsus
  \item 12: Right mid-leg tarsus
  \item 13: Right hind-leg tarsus
  \item 14: Left hind-leg tarsus
  \item 15: Left mid-leg tarsus
  \item 16: Left front-leg tarsus
  \item 17: Right wing tip
  \item 18: Right wing
  \item 19: Left wing tip
  \item 20: Left wing
  \item Used in [Step 3](\#step-3-keypoint-tracker---training)
\end{itemize}

\paragraph{Per-Frame Feature Definitions}
\begin{itemize}[nosep]
  \item \textbf{Location:} \texttt{/data/perframe/}
  \item \texttt{perframe\_feature\_definitions.pdf} and corresponding TeX source \texttt{perframe\_feature\_definitions.tex} contain detailed definitions for all per-frame features. Refer to them for exact computation methods.
  \item Used in [Step 5](\#step-5-per-frame-feature-computation)
\end{itemize}

\paragraph{Walking Behavior Classification Labels}
\textbf{Location}: \texttt{/data/walking\_labels/}

Sparsely labeled training data that can be used to train a classifier of whether a fly is walking or not, used in [Step 6](\#step-6-walking-behavior-classification---training)

Labels are in \texttt{labels.pkl} with the following variables:
\begin{itemize}[nosep]
  \item \texttt{sampleDirs}: list of experiment directory names [\texttt{sample\_1}, \texttt{sample\_2}, \texttt{sample\_3}, \texttt{sample\_4}]
  \item \texttt{flies}: list of lists, flies[sample\_idx] = list of fly IDs (1-indexed) with labels
  \item \texttt{t0s}: list of lists of arrays, t0s[sample\_idx][fly\_idx] = start frames of labeled bouts (0-indexed)
  \item \texttt{t1s}: list of lists of arrays, t1s[sample\_idx][fly\_idx] = end frames of labeled bouts (0-indexed, exclusive)
  \item \texttt{labels}: list of lists of arrays, labels[sample\_idx][fly\_idx] = label per bout (1=Walk, 0=None)
Labels are associated with the data files \texttt{sample\_1/registered\_trx.pkl}, \texttt{sample\_2/registered\_trx.pkl}, etc.
  \item Experiment: \texttt{labels[`flies'][sample\_idx]},  \texttt{labels[`t0s'][sample\_idx]}, etc. correspond to the data file in \texttt{labels['sampleDirs'][sample\_idx]}
  \item Fly: \texttt{labels[`t0s'][sample\_idx][fly\_idx]}, \texttt{labels[`labels'][sample\_idx][fly\_idx]}, etc. correspond to fly \texttt{flyid = labels[`flies'][sample\_idx][fly\_idx] - 1}
  \item Frames: Frames from \texttt{labels[`t0s'][sample\_idx][fly\_idx][bout\_idx]} to \texttt{labels[`t1s'][sample\_idx][fly\_idx][bout\_idx]} (exclusive) all have the label \texttt{labels[`labels'][sample\_idx][fly\_idx][bout\_idx]}
  \item Not all frames are labeled
  \item Frame indexing: \texttt{t0s} and \texttt{t1s} values are 0-indexed frame indices that directly correspond to indices in the perframe arrays.
  \item Fly indexing: Fly id is 0-indexed in perframe arrays but 1-indexed in \texttt{labels[`flies']}.
\end{itemize}

Each \texttt{\{sample\}/registered\_trx.pkl} contains the tracked trajectories of \textbf{all} flies in a video, defined by the following variables:
\begin{tabular}{p{0.16\textwidth} p{0.38\textwidth} p{0.38\textwidth}}
\hline
Field & Type & Description \\
\hline
\texttt{x} & list of ntargets arrays, each shape (nframes\_i,) & x-coordinate of each animal in pixels \\
\texttt{y} & list of ntargets arrays, each shape (nframes\_i,) & y-coordinate of each animal in pixels \\
\texttt{theta} & list of ntargets arrays, each shape (nframes\_i,) & Orientation of each animal (head direction) in radians \\
\texttt{a} & list of ntargets arrays, each shape (nframes\_i,) & 1/4 of the major-axis length in pixels \\
\texttt{b} & list of ntargets arrays, each shape (nframes\_i,) & 1/4 of the minor-axis length in pixels \\
\texttt{x\_mm} & list of ntargets arrays, each shape (nframes\_i,) & x-coordinate of each animal in mm \\
\texttt{y\_mm} & list of ntargets arrays, each shape (nframes\_i,) & y-coordinate of each animal in mm \\
\texttt{theta\_mm} & list of ntargets arrays, each shape (nframes\_i,) & Orientation in real coordinates (often same as theta) \\
\texttt{a\_mm} & list of ntargets arrays, each shape (nframes\_i,) & 1/4 of the major-axis length in mm \\
\texttt{b\_mm} & list of ntargets arrays, each shape (nframes\_i,) & 1/4 of the minor-axis length in mm \\
\texttt{dt} & list of ntargets arrays, each shape (nframes\_i-1,) & Time difference between consecutive frames in seconds \\
\texttt{id} & 1D array of shape (ntargets,) & Identity number of each trajectory (1-indexed) \\
\texttt{fps} & 1D array of shape (ntargets,) & Frames per second for each trajectory \\
\texttt{pxpermm} & list of ntargets arrays, each shape (1,) & Pixels per millimeter calibration \\
\texttt{firstframe} & 1D array of shape (ntargets,) & Global frame index where each animal's trajectory begins (0-indexed) \\
\texttt{endframe} & 1D array of shape (ntargets,) & Global frame index where each animal's trajectory ends (0-indexed) \\
\texttt{nframes} & 1D array of shape (ntargets,) & Number of frames in each animal's trajectory \\
\texttt{ntargets} & int & Total number of tracked animals \\
\texttt{timestamps} & 1D array of shape (total\_frames,) & Timestamp of each frame in seconds \\
\hline
\end{tabular}

Associations between files: Trajectory data corresponding to a label for experiment \texttt{i}, fly \texttt{j} and bout \texttt{k}:
\begin{lstlisting}[language=Python,basicstyle=\ttfamily\small,breaklines=true]
expdir = Path(labels['sampleDirs'][i])
with open(Path('walking_labels') / expdir / 'registered_trx.pkl', 'rb') as f:
    trx = pickle.load(f)

label = labels['labels'][i][j][k]
fly = labels['flies'][i][j] - 1
t0 = labels['t0s'][i][j][k]
t1 = labels['t1s'][i][j][k]
i0 = t0 - trx['firstframe'][fly]
i1 = t1 - trx['firstframe'][fly]
x_mm = trx['x_mm'][fly][i0:i1]
y_mm = trx['y_mm'][fly][i0:i1]
...
\end{lstlisting}

\paragraph{Swing/Stance Classification Labels}
\textbf{Location}: \texttt{/data/swing\_stance\_labels/}

\textbf{keypoints\_for\_labels.npz} (\texttt{/data/swing\_stance\_labels/keypoints\_for\_labels.npz})
\begin{itemize}[nosep]
  \item Tracking data containing (x,y) pixel coordinates of leg keypoints across all frames
  \item Numpy compressed archive with two arrays:
  \item \texttt{trk\_dense}: 4D array of shape \texttt{(n\_keypoints, 2, n\_tracked\_frames, n\_flies)} containing (x,y) coordinates. \texttt{trk\_dense[..., i, :]} corresponds to frame \texttt{T0 + i}.
  \item \texttt{T0}: First frame index for the tracking data (integer)
  \item Coordinates in trk\_dense may be NaN for frames before a fly's trajectory begins or after it ends, but are always valid (non-NaN) within the trajectory. These frames should be treated as not tracked.
  \item Keypoint indices match those in [keypoint training data](\#keypoint-training-data)
  \item This data corresponds to the ground truth labels in \texttt{groundcontact\_labels.csv}
\end{itemize}

\textbf{groundcontact\_labels.csv}
\begin{itemize}[nosep]
  \item Ground truth swing/stance labels corresponding to \texttt{keypoints\_for\_labels.npz}
  \item Columns: \texttt{frame}, \texttt{fly}, \texttt{leg}, \texttt{ground\_contact} (all indices are 0-based)
  \item The \texttt{ground\_contact} column is \texttt{1} when the leg is on the ground (stance) and \texttt{0} when off the ground (swing)
  \item The leg column is a 0-based index (0--5) corresponding to keypoints in increasing order: 11, 12, 13, 14, 15, 16.
  \item Only a few frames are labeled. Consider all other frames unlabeled.
\end{itemize}

\paragraph{Helper Modules}
The following Python modules are available in \texttt{helpers/} to assist with video processing:

\textbf{\texttt{movies.py}} - Generic movie interface supporting multiple formats
\begin{itemize}[nosep]
  \item \texttt{Movie} class: Opens and reads various video formats
  \item Methods: \texttt{get\_frame(frame\_num)}, \texttt{get\_n\_frames()}, \texttt{get\_width()}, \texttt{get\_height()}
  \item \texttt{FlyMovieFormat.py}, \texttt{ufmf.py}, and \texttt{params.py} are included because they are dependencies for \texttt{movies.py}
\textbf{\texttt{TrkFile.py}} - Code for reading trajectory data
  \item \texttt{loadmat()} function to load Matlab \texttt{.mat} files
\end{itemize}

\subsubsection*{Pipeline Stages}

[Step 1](\#step-1-fly-body-tracking). \textbf{Fly Body Tracking} --- Detect and track fly bodies across video frames, outputting trajectories with position, orientation, and size

[Step 2](\#step-2-trajectory-registration). \textbf{Trajectory Registration} --- Detect the arena boundary, convert coordinates to millimeters, and clean tracking gaps

[Step 3](\#step-3-keypoint-tracker---training). \textbf{Keypoint Tracker (Training)} --- Train a pose estimation model to detect keypoints on cropped fly images

[Step 4](\#step-4-keypoint-tracker---inference). \textbf{Keypoint Tracker (Inference)} --- Apply the trained keypoint model to track positions for all flies across video frames

[Step 5](\#step-5-per-frame-feature-computation). \textbf{Per-Frame Feature Computation} --- Compute 16 movement features (velocity, acceleration, orientation changes, inter-fly distances, etc.)

[Step 6](\#step-6-walking-behavior-classification---training). \textbf{Walking Behavior Classification (Training)} --- Train a classifier to detect when flies are walking vs. stationary

[Step 7](\#step-7-walking-behavior-classification---inference). \textbf{Walking Behavior Classification (Inference)} --- Apply the trained classifier to label walking frames in new data

[Step 8](\#step-8-swingstance-classification). \textbf{Swing/Stance Classification} --- For walking frames, classify each leg as swing (lifted) or stance (grounded) based on leg-tip speeds

[Step 9](\#step-9-stance-duration-analysis). \textbf{Stance Duration Analysis} --- Compare stance durations between stimulus ON/OFF periods across experimental and control groups using statistical tests

Implement each stage as part of your \texttt{fly\_pipeline.py} script. Use the reference data to develop and test your pipeline.

\textbf{Important}: Steps 1-2, 4-5, and 7-8 must be run on \textbf{every} experiment specified in the config. Steps 3 and 6 (model training) are run once using shared training data, then the trained models are applied to all experiments. Step 9 combines data across experiments for a combined analysis per line.

You will be assessed on the outputs from \textbf{each} of these steps.

\subsubsection*{Required Output: `fly\_pipeline.py`}

Create a Python script \texttt{/workspace/fly\_pipeline.py} that runs the entire analysis pipeline on arbitrary data.

\textbf{Usage:}
\begin{lstlisting}[language=bash,basicstyle=\ttfamily\small,breaklines=true]
python fly_pipeline.py config.json
\end{lstlisting}

The script takes a \textbf{JSON config file path} as its only CLI argument. The config specifies model paths, control experiments, and experimental lines to analyze:

\begin{lstlisting}[language=text,basicstyle=\ttfamily\small,breaklines=true]
{
    "keypoint_tracker_dir": "/workspace/keypoint_tracker",
    "walking_classifier_dir": "/workspace/walking_classifier",
    "control_experiments": [
        {
            "video_path": "/path/to/movie.ufmf",
            "indicatordata_path": "/path/to/indicatordata.mat",
            "output_dir": "/path/to/output/control_1"
        },
        ...
    ],
    "lines": [
        {
            "line_name": "line0",
            "experiments": [
                {
                    "video_path": "/path/to/movie.ufmf",
                    "indicatordata_path": "/path/to/indicatordata.mat",
                    "output_dir": "/path/to/output/line0/exp_1"
                },
                ...
            ],
            "analysis_output_path": "/path/to/analysis_line0.csv"
        },
        ...
    ]
}
\end{lstlisting}

\begin{itemize}[nosep]
  \item \texttt{keypoint\_tracker\_dir}: Directory to \textbf{load} the trained keypoint tracker from (\texttt{model.py} and \texttt{model.pkl}, from [Step 3](\#step-3-keypoint-tracker---training)). If these files do not exist, the script should train the model using \texttt{/data/keypoint\_train\_data/} and save it to this directory before proceeding.
  \item \texttt{walking\_classifier\_dir}: Directory to \textbf{load} the trained walking classifier from (\texttt{model.py} and \texttt{model.pkl}, from [Step 6](\#step-6-walking-behavior-classification---training)). If these files do not exist, the script should train the model using \texttt{/data/walking\_labels/} and save it to this directory before proceeding.
\end{itemize}

The script must:
\begin{enumerate}[nosep]
  \item Load the keypoint tracker and walking classifier from the config paths (training them first if they don't exist)
  \item Process \textbf{all} experiments (control and experimental) through Steps 1--8, saving intermediate outputs to each experiment's \texttt{output\_dir}
  \item For each experimental line, run Step 9 (stance analysis) comparing to the shared control group, saving results to \texttt{analysis\_output\_path}
  \item Handle any number of lines and any number of replicates per line
\end{enumerate}

\paragraph{Per-Experiment Output Structure}

For each experiment, the script must produce the following files in the experiment's \texttt{output\_dir}:

\begin{lstlisting}[language=text,basicstyle=\ttfamily\small,breaklines=true]
{output_dir}/
├── trx.pkl                 # Step 1: Body tracking
├── registered_trx.pkl      # Step 2: Registration
├── keypoints.npz           # Step 4: Keypoint inference
├── perframe/               # Step 5: Per-frame features
│   ├── absdtheta.pkl
│   └── ...
├── walk_preds.pkl           # Step 7: Walking classification
└── swing_stance.npy         # Step 8: Swing/stance classification
\end{lstlisting}

\paragraph{Trained Models}

During development, train and save your models to \texttt{/workspace/}:
\begin{itemize}[nosep]
  \item \texttt{/workspace/keypoint\_tracker/model.py} and \texttt{model.pkl} (Step 3)
  \item \texttt{/workspace/walking\_classifier/model.py} and \texttt{model.pkl} (Step 6)
\end{itemize}

These will be passed back to \texttt{fly\_pipeline.py} via the config's \texttt{keypoint\_tracker\_dir} and \texttt{walking\_classifier\_dir} fields at evaluation time.

\subsubsection*{Step 1. Fly Body Tracking}

Implement a fly tracking system that:
\begin{itemize}[nosep]
  \item \textbf{Inputs}: \texttt{video\_path} from the config (a \texttt{.ufmf} video file)
  \item \textbf{Outputs}: \texttt{\{output\_dir\}/trx.pkl} ([format below](\#body-tracking-output-format))
  \item Estimates the \textbf{position, orientation, and size} of the \textbf{body} of each fly (\textbf{not including wings}) in each frame of the input video
  \item Keeps track of the identities of individual flies across frames
\end{itemize}

Expected challenges:
\begin{itemize}[nosep]
  \item Multiple flies are present (need identity tracking)
  \item Flies may touch
\end{itemize}

To build your algorithm and tune its parameters you should:
\begin{itemize}[nosep]
  \item Examine the data and understand its statistical properties.
  \item Use your prior knowledge of what flies look like.
  \item Iterate: try an algorithm, look at statistical properties of its results, and adapt.
\end{itemize}

The approach is open-ended, but your implementation must:
\begin{enumerate}[nosep]
  \item \textbf{Estimate the pose} of all flies in all frames:
  \item Centroid Position (x, y)
  \item Abdomen to head orientation (theta)
  \item Length and width (a, b)
  \item \textbf{Track identities} across frames to maintain consistent fly IDs
  \item \textbf{Output in trx format} with all required fields correctly populated
\end{enumerate}

\paragraph{Body Tracking Output Format}

Each tracking output file \texttt{\{output\_dir\}/trx.pkl} must be a \textbf{single dictionary}. All frame indices are \textbf{0-indexed}. It must contain the following fields:

\textbf{Required fields:}

\begin{tabular}{p{0.16\textwidth} p{0.38\textwidth} p{0.38\textwidth}}
\hline
Field & Type & Description \\
\hline
\texttt{x} & list of ntargets arrays, each shape (nframes\_i,) & x-coordinate of centroid each animal in pixels \\
\texttt{y} & list of ntargets arrays, each shape (nframes\_i,) & y-coordinate of centroid each animal in pixels \\
\texttt{theta} & list of ntargets arrays, each shape (nframes\_i,) & Orientation of each animal (from tip of abdomen to head) in radians \\
\texttt{a} & list of ntargets arrays, each shape (nframes\_i,) & 1/4 of the length of the body (tip of the abdomen to head) in pixels \\
\texttt{b} & list of ntargets arrays, each shape (nframes\_i,) & 1/4 of the width of the body (left side to right side) in pixels \\
\texttt{dt} & list of ntargets arrays, each shape (nframes\_i-1,) & Time difference between consecutive frames in seconds \\
\texttt{off} & list of ntargets arrays, each shape (1,) & Offset for computing index: \texttt{-firstframe} \\
\texttt{firstframe} & 1D array of shape (ntargets,) & Global frame index where each animal's trajectory begins (0-indexed) \\
\texttt{endframe} & 1D array of shape (ntargets,) & Global frame index where each animal's trajectory ends (inclusive, 0-indexed) \\
\texttt{nframes} & 1D array of shape (ntargets,) & Number of frames in each animal's trajectory \\
\texttt{ntargets} & int & Total number of tracked animals \\
\texttt{timestamps} & 1D array of shape (total\_frames,) & Timestamp of each frame in seconds (global, not per-target) \\
\hline
\end{tabular}

\textbf{Identity tracking}: Each list entry should correspond to the same fly identity. If an identity is lost, a new entry can be added to the list. The number of entries in the lists must therefore be at least the number of flies, but could be longer.  

\textbf{Data alignment:} All per-target list fields (\texttt{x}, \texttt{y}, \texttt{theta}, \texttt{a}, \texttt{b}, etc.) have an entry per fly, with \texttt{x[i][idx]} corresponding to the position of the \texttt{i}th fly at frame \texttt{idx + firstframe[i]}

\textbf{Coordinate system:}
\begin{itemize}[nosep]
  \item \texttt{x = 0} corresponds to the first column (\texttt{axis = 1}) and \texttt{y = 0} corresponds to the first row (\texttt{axis = 0}).
  \item If \texttt{(x\_head, y\_head)} and \texttt{(x\_abdomen, y\_abdomen)} are the coordinates of the head and the abdomen, then the orientation \texttt{theta} can be computed with \texttt{np.arctan2}:
\texttt{theta = np.arctan2(y\_head-y\_abdomen,x\_head-x\_abdomen)}
\end{itemize}

\textbf{Example structure:}

\begin{lstlisting}[language=Python,basicstyle=\ttfamily\small,breaklines=true]
trx = {
    'x': [np.array([...]), np.array([...])],          # list of 2 arrays for 2 flies
    'y': [np.array([...]), np.array([...])],
    'a': [np.array([...]), np.array([...])],
    'b': [np.array([...]), np.array([...])],
    'theta': [np.array([...]), np.array([...])],
    'dt': [np.array([...]), np.array([...])],
    'off': [np.array([0]), np.array([0])],            # offset = -firstframe
    'firstframe': np.array([0, 0]),                   # both flies start at frame 0
    'endframe': np.array([999, 999]),                 # both flies end at frame 999
    'nframes': np.array([1000, 1000]),                # both flies have 1000 frames
    'ntargets': 2,
    'timestamps': np.array([0.0, 0.0067, ...]),        # shape (total\_frames,)
}
\end{lstlisting}

\textbf{Accessing data for fly \texttt{i} at frame \texttt{f}:}

\begin{lstlisting}[language=Python,basicstyle=\ttfamily\small,breaklines=true]
# Get the index into the per-fly arrays
idx = f - trx['firstframe'][i]
x_position = trx['x'][i][idx]
y_position = trx['y'][i][idx]
\end{lstlisting}

\textbf{Output file:} Save the tracking results as \texttt{\{output\_dir\}/trx.pkl} using pickle:

\begin{lstlisting}[language=Python,basicstyle=\ttfamily\small,breaklines=true]
with open(os.path.join(output_dir, 'trx.pkl'), 'wb') as f:
    pickle.dump(trx, f)
\end{lstlisting}

\paragraph{Body Tracking Success Criteria}

\begin{itemize}[nosep]
  \item Downstream analyses will involve measuring properties of precisely how the flies move (e.g. their velocity, change in orientation).
  \item They will also involve using body position and orientation as the first stage of a two-stage (top-down) keypoint detection algorithm.
  \item We will assess the accuracy of the tracker by measuring both detection and cross-frame association errors:
  \item \textbf{Detection errors}: For every frame and fly, detection error will be measured based on the derived tip of the head and abdomen positions:
\begin{lstlisting}[language=Python,basicstyle=\ttfamily\small,breaklines=true]
x_head = x + 2*a*np.cos(theta)
y_head = y + 2*a*np.sin(theta)
x_abdomen = x + 2*a*np.cos(theta+np.pi)
y_abdomen = y + 2*a*np.sin(theta+np.pi)
\end{lstlisting}
Detection error will be computed as
\begin{lstlisting}[language=Python,basicstyle=\ttfamily\small,breaklines=true]
dist_head = np.sqrt((x_head_true - x_head_pred)**2 + (y_head_true - y_head_pred)**2)
dist_abdomen = np.sqrt((x_abdomen_true - x_abdomen_pred)**2 + (y_abdomen_true - y_abdomen_pred)**2)
err = (dist_head + dist_abdomen) / a / 2
\end{lstlisting}
where \texttt{\textit{\_true} are the groundtruth and \texttt{}\_pred} are your algorithm's output predictions.
  \item \textbf{Association errors}: We also penalize identity switches: when a single ground-truth fly is assigned to different track predictions over two frames.
\end{itemize}

\subsubsection*{Step 2. Trajectory Registration}

Process raw fly tracking data by:
\begin{itemize}[nosep]
  \item Detecting arena boundary
  \item Converting from pixel coordinates to a common coordinate system defined by the arena boundary
  \item Removing bad tracking (indicated by NaNs)
\end{itemize}

\begin{itemize}[nosep]
  \item \textbf{Inputs:}
  \item \texttt{video\_path} from the config
  \item {[Body trajectories]}(\#body-tracking-output-format) \texttt{\{output\_dir\}/trx.pkl} from Step 1
  \item \textbf{Outputs:} Registered trajectories in physical units and a common coordinate system: \texttt{\{output\_dir\}/registered\_trx.pkl}
\end{itemize}

Your trajectory registration code must:
\begin{enumerate}[nosep]
  \item \textbf{Detect the arena boundary from the video}
  \item \textbf{Spatially register trajectory coordinates}
  \item Transform the trajectories so that (0,0) corresponds to the arena center
  \item Transform the trajectories so that the units are in millimeters
  \item \textbf{Note}: This step ONLY computes the \texttt{*\_mm} fields; it does NOT modify the pixel-based fields
  \item \texttt{x\_mm}, \texttt{y\_mm}: The ellipse center in millimeters
  \item \texttt{a\_mm}, \texttt{b\_mm}: Ellipse quarter-major and quarter-minor axes in millimeters
  \item \texttt{theta\_mm}: Same as \texttt{theta} (orientation in radians is scale-invariant)
  \item \textbf{Remove NaN gaps from trajectories}
  \item Interpolate short gaps ($\leq$5 frames) using linear interpolation
  \item Crop leading and trailing NaN values
  \item Split trajectories at long gaps ($>$5 frames) into separate segments
  \item \textbf{Note}: This step modifies all relevant pixel-based fields
\end{enumerate}

\paragraph{Registered Trajectory Output Format}
The output must contain all input fields (possibly modified by NaN removal) plus the registered \texttt{*\_mm} fields:

\begin{tabular}{p{0.18\textwidth} p{0.30\textwidth} p{0.44\textwidth}}
\hline
Field & Type & Description \\
\hline
\texttt{x\_mm} & list of arrays & x-coordinate in mm, arena-centered (per fly) \\
\texttt{y\_mm} & list of arrays & y-coordinate in mm, arena-centered (per fly) \\
\texttt{theta\_mm} & list of arrays & Orientation in radians, same as theta (per fly) \\
\texttt{a\_mm} & list of arrays & 1/4 of major-axis length in mm (per fly) \\
\texttt{b\_mm} & list of arrays & 1/4 of minor-axis length in mm (per fly) \\
\texttt{pxpermm} & list of arrays of shape (1,) & Pixels per millimeter conversion factor (per fly, same for all flies) \\
\hline
\end{tabular}

\textbf{Important constraints:}
\begin{itemize}[nosep]
  \item Output trajectories must have \textbf{no NaN values} in any field
  \item All fields must be internally consistent (e.g., for each fly, array lengths must match that fly's \texttt{nframes})
\end{itemize}

\textbf{Output files:} Save as \texttt{\{output\_dir\}/registered\_trx.pkl} for each experiment.

\paragraph{Registration Success Criteria}

\begin{itemize}[nosep]
  \item Circular arena correctly detected from video
  \item No NaN values in output trajectories
  \item All trajectories have consistent field lengths
  \item \texttt{*\_mm} fields correctly computed: translated to arena center and scaled to mm
  \item Pixel-based fields only modified by NaN interpolation/removal, not by registration
  \item Output format matches specification
\end{itemize}

\subsubsection*{Step 3. Keypoint Tracker - Training}

\begin{itemize}[nosep]
  \item Train a pose estimation model for flies using the MultiFly Dataset.
  \item \textbf{This step is run once; the trained model will be applied to all experiments in Step 4.}
  \item \textbf{Inputs}: [Keypoint Training Data](\#keypoint-training-data) in \texttt{/data/keypoint\_train\_data/}
  \item \textbf{Outputs}: Pose estimation model to \texttt{/workspace/keypoint\_tracker/}
\end{itemize}

\textbf{Notes:}
\begin{itemize}[nosep]
  \item Each annotation correspond to a single fly.
  \item Implement any pre-processing code directly into the train function
  \item You may apply any feature transformations or engineering you find useful
\end{itemize}

\paragraph{Keypoint Tracker Training Success Criteria}

\begin{itemize}[nosep]
  \item The downstream task of segmenting walking gait into swing and stance involves measuring the precise movements of each keypoint.
  \item \textbf{Median error}: The median per-keypoint Euclidean error should be well below 1 pixel (subpixel). Target $\leq$ 0.5 px.
  \item \textbf{Tail error}: Outlier percentile errors should also be small. Large outlier errors will corrupt velocity estimates for multiple surrounding frames, and scientists will not trust a model that makes large mistakes when the correct keypoint locations are easy to estimate. Target 95th percentile error <= 2 px, 97.5 percentile error <= 3.5 px.
  \item During development, the agent should monitor and report the full error distribution (50th, 75th, 90th, 95th, and 97.5th percentiles), not just mean error.
  \item The model will be evaluated based on the mean over all keypoints of the mean over these percentiles of the Euclidean error:
$\frac{1}{K} \sum_{k=1}^K \frac{1}{|P|}\sum_{p \in P} percentile(\{\|x^{pred}_{ik} - x^*_{ik}\}_{i=1}^N,p)$
where $x^{pred}_{ik}$ is the model's prediction on example $i$ for keypoint $k$, $x^*_{ik}$ is the groundtruth label, and $P = \{50,75,90,95,97.5\}$
  \item Target mean percentile error $\leq$ 1.7 px.
\end{itemize}

\paragraph{Keypoint Tracker Training Outputs}

\begin{itemize}[nosep]
  \item Save your model class definition to \texttt{/workspace/keypoint\_tracker/model.py}
  \item Save the trained model instance to \texttt{/workspace/keypoint\_tracker/model.pkl}
\end{itemize}

\paragraph{Keypoint Tracker Model Interface}

Your \texttt{/workspace/keypoint\_tracker/model.py} must define a class called \texttt{KeypointTracker} with the following interface:

\begin{lstlisting}[language=Python,basicstyle=\ttfamily\small,breaklines=true]
class KeypointTracker:
   def __init__(self, num_keypoints=21):
      self.model = None # Your trained model

   def track(self, X: np.ndarray) -> np.ndarray:
      '''
      Track fly keypoints on a cropped image of a fly. 

      Args: 
         X: 4D array of shape (n_frames, H, W, C) - batch of single-crop images of flies
      Returns:
         3D array of shape (n_frames, num_keypoints, 2) - predicted keypoint positions (x,y)
      '''
      # APPLY PRE-PROCESSING HERE 
      # Use self.model for inference
\end{lstlisting}

\paragraph{Saving the Keypoint Model}

\begin{itemize}[nosep]
  \item \textbf{\texttt{/workspace/keypoint\_tracker/model.py}} contains the class definition (the code for \texttt{KeypointTracker})
  \item \textbf{\texttt{/workspace/keypoint\_tracker/model.pkl}} contains the trained \texttt{KeypointTracker} instance
\end{itemize}

\begin{lstlisting}[language=Python,basicstyle=\ttfamily\small,breaklines=true]
# In your training script:
from model import KeypointTracker
import pickle

# Create and train
tracker = KeypointTracker()
tracker.model = trained_model  # your trained model

# Save the trained instance
with open('/workspace/keypoint_tracker/model.pkl', 'wb') as f:
    pickle.dump(tracker, f)
\end{lstlisting}

The model will be loaded as:

\begin{lstlisting}[language=Python,basicstyle=\ttfamily\small,breaklines=true]
from model import KeypointTracker
import pickle

with open('/workspace/keypoint_tracker/model.pkl', 'rb') as f:
    tracker = pickle.load(f)
\end{lstlisting}

\subsubsection*{Step 4. Keypoint Tracker - Inference}

\begin{itemize}[nosep]
  \item Apply the trained keypoint tracker to detect all 21 keypoints for all flies across all video frames. This produces the keypoint tracking data needed for swing/stance classification.
  \item \textbf{Inputs}:
  \item Trained keypoint tracker from \texttt{keypoint\_tracker\_dir} in the config
  \item \texttt{video\_path} from the config
  \item {[Registered trajectories]}(\#registered-trajectory-output-format) \texttt{\{output\_dir\}/registered\_trx.pkl} from Step 2
  \item \textbf{Outputs}: \texttt{\{output\_dir\}/keypoints.npz}
\end{itemize}

\paragraph{Keypoint Tracker Output Format}

Save results to \textbf{\texttt{\{output\_dir\}/keypoints.npz}} for each experiment as a numpy compressed archive with the following arrays:

\begin{tabular}{p{0.20\textwidth} p{0.36\textwidth} p{0.36\textwidth}}
\hline
Array & Shape & Description \\
\hline
\texttt{trk\_dense} & \texttt{(n\_keypoints, 2, n\_tracked\_frames, n\_flies)} & Keypoint (x, y) coordinates in pixels \\
\texttt{T0} & scalar & First frame index for the tracking data \\
\hline
\end{tabular}

\textbf{Array details:}
\begin{itemize}[nosep]
  \item \texttt{trk\_dense[k, 0, f, fly]}: x-coordinate of keypoint \texttt{k} for fly \texttt{fly} at frame \texttt{T0 + f}
  \item \texttt{trk\_dense[k, 1, f, fly]}: y-coordinate of keypoint \texttt{k} for fly \texttt{fly} at frame \texttt{T0 + f}
  \item \texttt{n\_keypoints}: 21 keypoints
  \item Coordinates are in the original video's pixel coordinate system
  \item For frames where a fly is not tracked, use NaN values
\end{itemize}

\begin{lstlisting}[language=Python,basicstyle=\ttfamily\small,breaklines=true]
import numpy as np

# Save keypoint tracking results for each experiment
np.savez(os.path.join(output_dir, 'keypoints.npz'), trk_dense=trk_dense, T0=T0)
\end{lstlisting}

\subsubsection*{Step 5. Per-Frame Feature Computation}

Compute a set of features from fly tracking data that capture instantaneous properties of how the fly is moving.
\begin{itemize}[nosep]
  \item \textbf{Inputs:} {[Registered trajectories]}(\#registered-trajectory-output-format): \texttt{\{output\_dir\}/registered\_trx.pkl} from Step 2
{[\textbf{Feature definitions}]}(\#per-frame-feature-definitions): \texttt{/data/perframe/perframe\_feature\_definitions.pdf} and corresponding TeX source \texttt{/data/perframe/perframe\_feature\_definitions.tex}
\end{itemize}

\paragraph{Per-Frame Feature Outputs}

Compute all of the following features for each experiment. Each feature must be saved as \texttt{\{output\_dir\}/perframe/\{feature\_name\}.pkl} (e.g., \texttt{absdtheta} saves to \texttt{\{output\_dir\}/perframe/absdtheta.pkl}):
\begin{lstlisting}[language=text,basicstyle=\ttfamily\small,breaklines=true]
absdtheta
absdv_cor
corfrac_maj
corfrac_min
dphi
dtheta
du_cor
du_ctr
dv_cor
dv_ctr
flipdv_cor
phisideways
velmag
velmag_ctr
velmag_nose
yaw
\end{lstlisting}

Each feature file must be a pickle file containing a \textbf{list of arrays}, one array per fly:

\begin{lstlisting}[language=Python,basicstyle=\ttfamily\small,breaklines=true]
# Example: perframe/velmag.pkl
[
    np.array([...]),  # fly 0
    np.array([...]),  # fly 1
    ...
]
\end{lstlisting}

\textbf{Array lengths:}
\begin{itemize}[nosep]
  \item Position-based features: \texttt{nframes} per fly
  \item Velocity/derivative features: \texttt{nframes-1} per fly
\end{itemize}

\textbf{Output location}: Save all files to \texttt{\{output\_dir\}/perframe/} directory for each experiment.

\paragraph{Per-Frame Feature Success Criteria}

\begin{enumerate}[nosep]
  \item All 16 feature files are created in \texttt{\{output\_dir\}/perframe/} for each experiment
  \item Each file contains a list with \texttt{ntargets} arrays
  \item Array lengths match expected dimensions per feature type
  \item Feature values match ground truth within numerical tolerance
\end{enumerate}

\subsubsection*{Step 6. Walking Behavior Classification - Training}

\begin{itemize}[nosep]
  \item Train a \textbf{binary classifier} to detect "walking" behavior in fruit flies (Drosophila) using pre-computed per-frame movement features.
  \item To define the walking behavior, we provide a training dataset in which positive labels (1) correspond to walking and negative labels (0) correspond to not walking in [\texttt{/data/walking\_labels/labels.pkl}](\#walking-behavior-classification-labels).
  \item Your classifer should be \textbf{based on per-frame features} derived from the provided trajectory data [\texttt{/data/walking\_labels/\{sample\}/registered\_trx.pkl}](\#walking-behavior-classification-labels).
\end{itemize}

You must:
\begin{enumerate}[nosep]
  \item Load the labels from \texttt{/data/walking\_labels/}
  \item Use your code for computing per-frame features from [Step 5](\#step-5-per-frame-feature-computation) to transform \texttt{/data/walking\_labels/\{sample\}/registered\_trx.pkl} into per-frame features
  \item Train a classifier to predict walking behavior (1) vs. not walking (0)
  \item Save your classifier class definition to \texttt{/workspace/walking\_classifier/model.py}
  \item Save the trained model instance to \texttt{/workspace/walking\_classifier/model.pkl}
\textbf{This step is run once; the trained model will be applied to all experiments in Step 7.}
\end{enumerate}

\paragraph{Walking Classifier Model Output Format}

Your \texttt{model.py} must define a class called \texttt{WalkingClassifier} with the following interface:

\begin{lstlisting}[language=Python,basicstyle=\ttfamily\small,breaklines=true]
class WalkingClassifier:
    def __init__(self):
        self.model = None  # your trained model
        self.feature_names = []  # list of feature names used, e.g. ['velmag', 'dtheta', ...]

    def predict(self, X: np.ndarray) -> np.ndarray:
        '''
        Predict walking behavior for a single fly's trajectory.

        Args:
            X: 2D array of shape (n_frames, n_features) - full trajectory for one fly,
               features in order of self.feature_names
        Returns:
            1D array of shape (n_frames,) - predictions for each frame (1=walking, 0=not walking)
        '''
        # Apply any preprocessing, then use self.model to predict
        ...
\end{lstlisting}

The \texttt{predict} method should include any preprocessing or feature transformations applied during training.

Save your walking classifier to:
\begin{itemize}[nosep]
  \item \textbf{\texttt{/workspace/walking\_classifier/model.py}} contains the class definition (the code for \texttt{WalkingClassifier})
  \item \textbf{\texttt{/workspace/walking\_classifier/model.pkl}} contains the trained instance (with \texttt{self.model} and \texttt{self.feature\_names} set)
\end{itemize}

Example workflow:

\begin{lstlisting}[language=Python,basicstyle=\ttfamily\small,breaklines=true]
# In your training script:
from model import WalkingClassifier
import pickle

# Create and train
clf = WalkingClassifier()
clf.feature_names = ['velmag', 'dtheta', ...]
clf.model = trained_model  # your trained model

# Save the trained instance
with open('/workspace/walking_classifier/model.pkl', 'wb') as f:
    pickle.dump(clf, f)
\end{lstlisting}

The model will be loaded as:

\begin{lstlisting}[language=Python,basicstyle=\ttfamily\small,breaklines=true]
from model import WalkingClassifier
import pickle

with open('/workspace/walking_classifier/model.pkl', 'rb') as f:
    classifier = pickle.load(f)
\end{lstlisting}

\paragraph{Walking Classifier Success criteria}
\begin{itemize}[nosep]
  \item Your model will be assessed based on the balanced per-frame accuracy on a held-out groundtruth dataset.
\end{itemize}

\subsubsection*{Step 7. Walking Behavior Classification - Inference}

Apply the trained walking classifier to label each frame of the experiment data as walking or not walking.

\textbf{Inputs:}
\begin{itemize}[nosep]
  \item Trained walking classifier from \texttt{walking\_classifier\_dir} in the config
  \item {[Per-frame features from Step 5]}(\#per-frame-feature-outputs): \texttt{\{output\_dir\}/perframe/}
  \item {[Registered trajectory data from Step 2]}(\#registered-trajectory-output-format): \texttt{\{output\_dir\}/registered\_trx.pkl}
\end{itemize}

Using the trained classifier, for each fly, extract the required features from \texttt{\{output\_dir\}/perframe/}. Then, use the classifier to generate predictions for each frame, and save predictions for all flies in order.

\paragraph{Walking Classification Prediction Output Format}

Save predictions to \textbf{\texttt{\{output\_dir\}/walk\_preds.pkl}} for each experiment as a pickle file containing a list of length \texttt{ntargets}, where each element is a 1D ndarray of per-frame predictions for that fly.

\begin{lstlisting}[language=Python,basicstyle=\ttfamily\small,breaklines=true]
# walk_preds.pkl structure
[
    np.array([0, 0, 1, 1, 1, 0, ...]),  # fly 0 predictions
    np.array([1, 1, 1, 0, 0, 0, ...]),  # fly 1 predictions
    ...
]
\end{lstlisting}

\textbf{Array alignment:}
\begin{itemize}[nosep]
  \item Arrays are globally aligned: index \texttt{i} corresponds to frame \texttt{i} (starting from frame 0)
  \item For frames before a fly's \texttt{firstframe} or after its trajectory ends, use \texttt{np.nan}
  \item Prediction values: \texttt{1} = walking, \texttt{0} = not walking, \texttt{np.nan} = no data
\end{itemize}

\subsubsection*{Step 8. Swing/Stance Classification}

\begin{itemize}[nosep]
  \item \textbf{Goal}: Segment fruit flies' leg movements into swing and stance bouts for frames when the fly is walking:
  \item Stance phase: The leg is in contact with the ground and propelling the animal
  \item Swing phase: The leg is lifted from the ground and recovering for the next propulsion
  \item A small set of groundtruth labels from a separate video are provided to help you fine-tune your algorithm.
  \item For each experiment, for each fly, for each frame when the fly is walking, and for each of the six legs, compute whether the leg is in swing or stance phase.
\end{itemize}

\textbf{Inputs}:
\begin{itemize}[nosep]
  \item {[Registered body tracking]}(\#registered-trajectory-output-format): \texttt{\{output\_dir\}/registered\_trx.pkl} from Step 2
  \item {[Keypoint trajectories]}(\#keypoint-tracker-output-format): \texttt{\{output\_dir\}/keypoints.npz} from Step 4
  \item {[Walking classifier predictions]}(\#walking-classification-prediction-output-format): \texttt{\{output\_dir\}/walk\_preds.pkl} from Step 7
  \item {[Example swing/stance classification labels]}(\#swingstance-classification-labels)
\end{itemize}

\paragraph{Swing/Stance Classification Output Format}
Save results to \texttt{\{output\_dir\}/swing\_stance.npy} for each experiment as a 3D numpy array with shape \texttt{(n\_flies, total\_frames, n\_legs)} where:
\begin{itemize}[nosep]
  \item \texttt{n\_flies}: number of tracked flies
  \item \texttt{total\_frames}: total number of video frames (indexed from frame 0)
  \item \texttt{n\_legs}: 6 (in increasing order of keypoint index: 11, 12, 13, 14, 15, 16)
\end{itemize}

Array index corresponds to absolute frame number (e.g., \texttt{swing\_stance[fly, 500, leg]} is the state at frame 500).

Each element of the array is an integer:
\begin{itemize}[nosep]
  \item \texttt{0}: leg is in swing phase (off the ground)
  \item \texttt{1}: leg is in stance phase (on the ground)
  \item \texttt{-1}: fly is not walking or frame is not tracked
\end{itemize}

\paragraph{Swing/Stance Classification Success Criteria}

\begin{itemize}[nosep]
  \item The output segmentations will be used to measure the duration of the swing and stance bouts, thus it is important that your algorithm reproduce the \textbf{bout structure}.
  \item A bout is a maximal contiguous run of frames with the same label (swing or stance), considering only frames where the fly is walking. Non-walking frames are excluded before determining bout boundaries.
  \item Your solution will be evaluated with a bout-level error criterion:
\begin{multline*}
\frac{1}{2N^*}\sum_{b=1}^{N^*} \frac{1}{t_{b+1}^*-t_b^*}\sum_{t=t_b^*}^{t_{b+1}^*-1} I(y_t^{pred} \neq y_t^*) \\+ \frac{1}{2N^{pred}}\sum_{b=1}^{N^{pred}} \frac{1}{t_{b+1}^{pred}-t_b^{pred}}\sum_{t=t_b^{pred}}^{t_{b+1}^{pred}-1} I(y_t^* \neq y_t^{pred})\end{multline*}
where $y_t^*$ is the groundtruth label for frame $t$ and $y_t^{pred}$ is the model's prediction, and $(t_1^*,...,t_{N^*}^*)$ is the groundtruth bout segmentation and $(t_1^{pred},...,t_{N^{pred}}^{pred})$ is the model's bout segmentation.
  \item The sums over bouts are taken across all flies and all legs jointly (not averaged per-fly or per-leg).
  \item This metric penalizes over-fragmenting true bouts.
\end{itemize}

\subsubsection*{Step 9. Stance Duration Analysis}

Analyze how optogenetic perturbation affects stance duration in fruit fly locomotion, comparing experimental flies to control flies. Optogenetic perturbations are controlled by exposing flies to a particular wavelength of light, and occur when the lights are ON. This step investigates whether the difference in stance duration between lights ON and OFF periods is significantly different between experimental and control groups, for both slow and fast walking speeds.

Stance duration is the time a leg spends in contact with the ground during a walking gait cycle. For each line, you will compute the ON-OFF difference in mean stance duration for both slow and fast walking, then use statistical tests to compare these differences between experimental and control groups.

For each line in the config, combine all experimental replicates and compare against the shared control group. For each experiment, the following outputs from previous steps are used:
\begin{itemize}[nosep]
  \item \texttt{\{output\_dir\}/swing\_stance.npy} (from Step 8)
  \item \texttt{\{output\_dir\}/registered\_trx.pkl} (from Step 2)
  \item \texttt{\{output\_dir\}/perframe/velmag\_ctr.pkl} (from Step 5)
  \item \texttt{indicatordata\_path} from the config ([Lights ON/OFF timing](\#lights-onoff-timing))
\end{itemize}

\begin{itemize}[nosep]
  \item The \textbf{ON-OFF difference} is defined as mean duration of all stance bouts contained within the ON period minus the mean duration of all stance bouts contained within the preceding OFF period.
  \item A stance duration is the length (in frames) of a contiguous run of stance phase (value = 1)
  \item Frames with swing\_stance value -1 should be excluded
\end{itemize}

\textbf{Statistical analysis details}:
\begin{itemize}[nosep]
  \item For each line, perform two Mann-Whitney U tests (two-sided), comparing experimental vs control ON-OFF differences:
  \item \textbf{Slow walking} (mean stance bout centroid velocity $<$17.5 mm/s)
  \item \textbf{Fast walking} (mean stance bout centroid velocity $\geq$17.5 mm/s and $<$50 mm/s)
  \item Within each video, there are 10 trials: 10 periods of optogenetic perturbation (ON), each preceded by a baseline (OFF) period.
  \item A \textbf{sample} for the statistical tests should be \textbf{one pairing of ON and its preceding OFF period}
  \item With 4 experimental and 4 control replicates, the Mann-Whitney U test compares \textbf{4 x 10} experimental samples against \textbf{4 x 10} control samples.
\end{itemize}

\paragraph{Stance Duration Analysis Outputs}
Save results to the \texttt{analysis\_output\_path} specified in the config for each line, with the following format:
\begin{lstlisting}[language=text,basicstyle=\ttfamily\small,breaklines=true]
metric,value
u_statistic_slow,<value>
p_value_slow,<value>
u_statistic_fast,<value>
p_value_fast,<value>
sig_slow_0.05,<True/False>
sig_slow_0.01,<True/False>
sig_slow_0.001,<True/False>
sig_fast_0.05,<True/False>
sig_fast_0.01,<True/False>
sig_fast_0.001,<True/False>
\end{lstlisting}

Where:
\begin{itemize}[nosep]
  \item \textbf{\texttt{u\_statistic\_slow}}: U-statistic from the Mann-Whitney test comparing slow walking ON-OFF differences (exp vs control)
  \item \textbf{\texttt{p\_value\_slow}}: Two-tailed p-value from the Mann-Whitney test for slow walking
  \item \textbf{\texttt{u\_statistic\_fast}}: U-statistic from the Mann-Whitney test comparing fast walking ON-OFF differences (exp vs control)
  \item \textbf{\texttt{p\_value\_fast}}: Two-tailed p-value from the Mann-Whitney test for fast walking
  \item \textbf{\texttt{sig\_*}}: Boolean indicating whether the difference is significant at the given threshold
\end{itemize}

\paragraph{Stance Duration Analysis Success criteria}
\begin{itemize}[nosep]
  \item Your True/False classifications must be correct for \textbf{all} lines, speeds (2), and significance thresholds (3).
  \item Be very precise in your analyses, as there are \texttt{n\_lines * 3 * 2} binary tests you must get exactly right.
\end{itemize}

\subsubsection*{Dependencies}

Write \textbf{ALL} Python package dependencies for \textbf{all} code to \texttt{requirements.txt}.

\subsubsection*{Documentation}

Create \texttt{NOTES.md} with a chronological record of your development process: your initial plan, what worked, what didn't, and how you iterated to arrive at your final solution. Update this file after each decision. 

Create \texttt{ALGORITHM.md} with:
\begin{enumerate}[nosep]
  \item \textbf{Approach}: Each step of your analysis, algorithms used, and key parameters
  \item \textbf{Validation}: Evidence your methods work correctly (e.g., performance on labeled data, sanity checks, representative examples)
This should allow someone to understand and reproduce your analysis.
\end{enumerate}
\end{tcolorbox}

\subsection{End-to-End Minimal prompt}
\label{sec:prompt-e2e-all-lines-minimal}

\begin{tcolorbox}[colback=gray!5, colframe=gray!50, breakable, title=Agent Prompt]
\sloppy
\subsection*{Fly Optogenetics Locomotion Analysis}

\textbf{Working directory}: Your working directory is \texttt{/workspace}. Save all output files here. Input data is located at \texttt{/data/}.

\textbf{Time limit}: You have a maximum of \textbf{24 hours} to complete this task.

\subsubsection*{Overview}

Create a Python script (\texttt{fly\_pipeline.py}) that analyzes how optogenetic perturbation affects stance duration in fruit fly locomotion, starting from raw video. The script must process video of multiple GAL4 lines, comparing experimental flies to control flies. Optogenetic perturbations are controlled by exposing flies to a particular wavelength of light, and occur when the lights are ON. The pipeline must determine whether the difference in stance duration between lights ON and OFF periods is significantly different between experimental and control groups, for both slow and fast walking speeds.

In insect locomotion, walking gait is divided into swing phase (leg lifted, moving forward through air) and stance phase (leg on ground, propelling body forward). Stance duration decreases as walking speed increases. For each line, you will compute the ON-OFF difference in mean stance duration for both slow and fast walking, then use statistical tests to compare these differences between experimental and control groups.

Reference data for 2 experimental lines and a shared control group is provided for you to develop and test your pipeline. Your pipeline will be evaluated on data from 10 GAL4 lines and controls (44 experiments total).

\begin{itemize}[nosep]
  \item For each of the GAL4 lines and control, there are four videos (experiments) each corresponding to a different group of \textasciitilde{}10 flies walking in a circular arena.
  \item Within each video, there are 10 trials: 10 periods of optogenetic perturbation (ON), each preceded by a baseline (OFF) period.
\end{itemize}

To help with analysis, human annotations that can be used for training machine learning models are provided.

\subsubsection*{Reference Input Data}

Reference data for 2 experimental lines and a shared control group is provided for development. Your pipeline will be evaluated on 10 lines and controls (44 experiments total). All data is organized under \texttt{/data/}:

\begin{lstlisting}[language=text,basicstyle=\ttfamily\small,breaklines=true]
/data/
├── line_1/                         # Experimental line 1
│   ├── exp_1/                      # Experiment replicate 1
│   │   ├── movie.ufmf              # Video file
│   │   └── indicatordata.mat       # Lights on and off timing
│   ├── exp_2/
│   │   └── ...
│   ├── exp_3/
│   │   └── ...
│   └── exp_4/
│       └── ...
├── line_2/                         # Experimental line 2
│   ├── exp_1/
│   │   └── ...
│   ├── exp_2/
│   │   └── ...
│   ├── exp_3/
│   │   └── ...
│   └── exp_4/
│       └── ...
├── control/                        # Shared control group (used for all lines)
│   ├── control_1/
│   │   ├── movie.ufmf
│   │   └── indicatordata.mat
│   ├── control_2/
│   │   └── ...
│   ├── control_3/
│   │   └── ...
│   └── control_4/
│       └── ...
├── keypoint_train_data/            # Keypoint detection training data
│   ├── labels.json                 # COCO format annotations
│   └── im/                         # Cropped fly images
├── walking_labels/                 # Walking behavior classifier training data
│   ├── sample_1/registered_trx.pkl # Tracked fly trajectories
│   ├── sample_2/registered_trx.pkl
│   ├── sample_3/registered_trx.pkl
│   ├── sample_4/registered_trx.pkl
│   └── labels.pkl                  # Walking behavior labels
└── swing_stance_labels/            # Swing/stance label data
    ├── keypoints_for_labels.npz    # Keypoint trajectories
    └── groundcontact_labels.csv    # Ground contact labels
\end{lstlisting}

\paragraph{Video Files}
\begin{itemize}[nosep]
  \item \textbf{Reference locations}: \texttt{/data/line\_1/exp\_1/movie.ufmf}, \texttt{/data/line\_1/exp\_2/movie.ufmf}, etc.
  \item \textbf{Format}: UFMF (Micro Fly Movie Format)
  \item \textbf{Frame rate}: \textasciitilde{}150 FPS
  \item \textbf{Arena radius}: 26.689 mm (physical)
  \item \textbf{Helper}: \texttt{helpers/movies.py} provides a \texttt{Movie} class with \texttt{get\_frame(frame\_num)}
\end{itemize}

\paragraph{Lights ON/OFF Timing}
\begin{itemize}[nosep]
  \item \textbf{Reference locations}: \texttt{/data/line\_1/exp\_1/indicatordata.mat}, etc.
  \item Information about when lights ON periods occur.
  \item The useful variables in this file are:
  \item \texttt{data[`indicatorLED'][`startframe'][i,0]}: Start frame of optogenetic perturbation period \texttt{i} (\textbf{1-indexed})
  \item \texttt{data[`indicatorLED'][`endframe'][i,0]}: End frame of optogenetic perturbation period \texttt{i} (\textbf{1-indexed}, inclusive)
  \item Outside of ON periods are OFF periods, i.e. OFF period \texttt{i} is from \texttt{data[`indicatorLED'][`endframe'][i-1]+1} to \texttt{data[`indicatorLED'][`startframe'][i]-1} (1-indexed, inclusive).
  \item Lights are OFF starting at frame 0
  \item Use \texttt{TrkFile.loadmat} from \texttt{helpers/TrkFile.py} to load this file
\end{itemize}

\paragraph{Keypoint Training Data}
\begin{itemize}[nosep]
  \item \textbf{Location}: \texttt{/data/keypoint\_train\_data/}
  \item The Multifly dataset contains annotated keypoints for cropped images centered around a single Drosophila melanogaster. The images are also rotated so that the flies are pointing up.
  \item There may be other flies in the image but the annotations only correspond to the fly at the center.
  \item \texttt{im}: Directory containing all of the training images cropped and aligned around the focus fly.
  \item \texttt{labels.json}: Training labels in COCO json format (https://cocodataset.org/\#format-data). All data are 0-indexed and readable with pycocotools (https://pypi.org/project/pycocotools). Each annotation has \texttt{keypoints}: coordinates of 21 keypoints, \texttt{[x1,y1,v1,x2,y2,v2,...]}. Keypoint indices:
  \item 0: Anterior head
  \item 1: Right eye
  \item 2: Left eye
  \item 3: Left thorax
  \item 4: Right thorax
  \item 5: Posterior notum
  \item 6: Posterior abdomen
  \item 7: Right femur
  \item 8: Right femur-tibia
  \item 9: Left femur
  \item 10: Left femur-tibia
  \item 11: Right front-leg tarsus
  \item 12: Right mid-leg tarsus
  \item 13: Right hind-leg tarsus
  \item 14: Left hind-leg tarsus
  \item 15: Left mid-leg tarsus
  \item 16: Left front-leg tarsus
  \item 17: Right wing tip
  \item 18: Right wing
  \item 19: Left wing tip
  \item 20: Left wing
\end{itemize}

\paragraph{Walking Behavior Classification Labels}
\textbf{Location}: \texttt{/data/walking\_labels/}

Sparsely labeled training data that can be used to train a classifier of whether a fly is walking or not.

Labels are in \texttt{labels.pkl} with the following variables:
\begin{itemize}[nosep]
  \item \texttt{sampleDirs}: list of experiment directory names [\texttt{sample\_1}, \texttt{sample\_2}, \texttt{sample\_3}, \texttt{sample\_4}]
  \item \texttt{flies}: list of lists, flies[sample\_idx] = list of fly IDs (1-indexed) with labels
  \item \texttt{t0s}: list of lists of arrays, t0s[sample\_idx][fly\_idx] = start frames of labeled bouts (0-indexed)
  \item \texttt{t1s}: list of lists of arrays, t1s[sample\_idx][fly\_idx] = end frames of labeled bouts (0-indexed, exclusive)
  \item \texttt{labels}: list of lists of arrays, labels[sample\_idx][fly\_idx] = label per bout (1=Walk, 0=None)
Labels are associated with the data files \texttt{sample\_1/registered\_trx.pkl}, \texttt{sample\_2/registered\_trx.pkl}, etc.
  \item Experiment: \texttt{labels[`flies'][sample\_idx]},  \texttt{labels[`t0s'][sample\_idx]}, etc. correspond to the data file in \texttt{labels[`sampleDirs'][sample\_idx]}
  \item Fly: \texttt{labels[`t0s'][sample\_idx][fly\_idx]}, \texttt{labels[`labels'][sample\_idx][fly\_idx]}, etc. correspond to fly \texttt{flyid = labels[`flies'][sample\_idx][fly\_idx] - 1}
  \item Frames: Frames from \texttt{labels[`t0s'][sample\_idx][fly\_idx][bout\_idx]} to \texttt{labels[`t1s'][sample\_idx][fly\_idx][bout\_idx]} (exclusive) all have the label \texttt{labels[`labels'][sample\_idx][fly\_idx][bout\_idx]}
  \item Not all frames are labeled
  \item Frame indexing: \texttt{t0s} and \texttt{t1s} values are 0-indexed frame indices that directly correspond to indices in the perframe arrays.
  \item Fly indexing: Fly id is 0-indexed in perframe arrays but 1-indexed in \texttt{labels[`flies']}.
\end{itemize}

Each \texttt{\{sample\}/registered\_trx.pkl} contains the tracked trajectories of \textbf{all} flies in a video, defined by the following variables:
\begin{tabular}{p{0.16\textwidth} p{0.38\textwidth} p{0.38\textwidth}}
\hline
Field & Type & Description \\
\hline
\texttt{x} & list of ntargets arrays, each shape (nframes\_i,) & x-coordinate of each animal in pixels \\
\texttt{y} & list of ntargets arrays, each shape (nframes\_i,) & y-coordinate of each animal in pixels \\
\texttt{theta} & list of ntargets arrays, each shape (nframes\_i,) & Orientation of each animal (head direction) in radians \\
\texttt{a} & list of ntargets arrays, each shape (nframes\_i,) & 1/4 of the major-axis length in pixels \\
\texttt{b} & list of ntargets arrays, each shape (nframes\_i,) & 1/4 of the minor-axis length in pixels \\
\texttt{x\_mm} & list of ntargets arrays, each shape (nframes\_i,) & x-coordinate of each animal in mm \\
\texttt{y\_mm} & list of ntargets arrays, each shape (nframes\_i,) & y-coordinate of each animal in mm \\
\texttt{theta\_mm} & list of ntargets arrays, each shape (nframes\_i,) & Orientation in real coordinates (often same as theta) \\
\texttt{a\_mm} & list of ntargets arrays, each shape (nframes\_i,) & 1/4 of the major-axis length in mm \\
\texttt{b\_mm} & list of ntargets arrays, each shape (nframes\_i,) & 1/4 of the minor-axis length in mm \\
\texttt{dt} & list of ntargets arrays, each shape (nframes\_i-1,) & Time difference between consecutive frames in seconds \\
\texttt{id} & 1D array of shape (ntargets,) & Identity number of each trajectory (1-indexed) \\
\texttt{fps} & 1D array of shape (ntargets,) & Frames per second for each trajectory \\
\texttt{pxpermm} & list of ntargets arrays, each shape (1,) & Pixels per millimeter calibration \\
\texttt{firstframe} & 1D array of shape (ntargets,) & Global frame index where each animal's trajectory begins (0-indexed) \\
\texttt{endframe} & 1D array of shape (ntargets,) & Global frame index where each animal's trajectory ends (0-indexed) \\
\texttt{nframes} & 1D array of shape (ntargets,) & Number of frames in each animal's trajectory \\
\texttt{ntargets} & int & Total number of tracked animals \\
\texttt{timestamps} & 1D array of shape (total\_frames,) & Timestamp of each frame in seconds \\
\hline
\end{tabular}

Associations between files: Trajectory data corresponding to a label for experiment \texttt{i}, fly \texttt{j} and bout \texttt{k}:
\begin{lstlisting}[language=Python,basicstyle=\ttfamily\small,breaklines=true]
expdir = Path(labels['sampleDirs'][i])
with open(Path('walking_labels') / expdir / 'registered_trx.pkl', 'rb') as f:
    trx = pickle.load(f)

label = labels['labels'][i][j][k]
fly = labels['flies'][i][j] - 1
t0 = labels['t0s'][i][j][k]
t1 = labels['t1s'][i][j][k]
i0 = t0 - trx['firstframe'][fly]
i1 = t1 - trx['firstframe'][fly]
x_mm = trx['x_mm'][fly][i0:i1]
y_mm = trx['y_mm'][fly][i0:i1]
...
\end{lstlisting}

\paragraph{Swing/Stance Classification Labels}
\textbf{Location}: \texttt{/data/swing\_stance\_labels/}

\textbf{keypoints\_for\_labels.npz} (\texttt{/data/swing\_stance\_labels/keypoints\_for\_labels.npz})
\begin{itemize}[nosep]
  \item Tracking data containing (x,y) pixel coordinates of leg keypoints across all frames
  \item Numpy compressed archive with two arrays:
  \item \texttt{trk\_dense}: 4D array of shape \texttt{(n\_keypoints, 2, n\_tracked\_frames, n\_flies)} containing (x,y) coordinates. \texttt{trk\_dense[..., i, :]} corresponds to frame \texttt{T0 + i}.
  \item \texttt{T0}: First frame index for the tracking data (integer)
  \item Coordinates in trk\_dense may be NaN for frames before a fly's trajectory begins or after it ends, but are always valid (non-NaN) within the trajectory. These frames should be treated as not tracked.
  \item Keypoint indices match those in [keypoint training data](\#keypoint-training-data)
  \item This data corresponds to the ground truth labels in \texttt{groundcontact\_labels.csv}
\end{itemize}

\textbf{groundcontact\_labels.csv}
\begin{itemize}[nosep]
  \item Ground truth swing/stance labels corresponding to \texttt{keypoints\_for\_labels.npz}
  \item Columns: \texttt{frame}, \texttt{fly}, \texttt{leg}, \texttt{ground\_contact} (all indices are 0-based)
  \item The \texttt{ground\_contact} column is \texttt{1} when the leg is on the ground (stance) and \texttt{0} when off the ground (swing)
  \item The leg column is a 0-based index (0--5) corresponding to keypoints in increasing order: 11, 12, 13, 14, 15, 16.
  \item Only a few frames are labeled. Consider all other frames unlabeled.
\end{itemize}

\paragraph{Helper Modules}
The following Python modules are available in \texttt{helpers/} to assist with video processing:

\textbf{\texttt{movies.py}} - Generic movie interface supporting multiple formats
\begin{itemize}[nosep]
  \item \texttt{Movie} class: Opens and reads various video formats
  \item Methods: \texttt{get\_frame(frame\_num)}, \texttt{get\_n\_frames()}, \texttt{get\_width()}, \texttt{get\_height()}
  \item \texttt{FlyMovieFormat.py}, \texttt{ufmf.py}, and \texttt{params.py} are included because they are dependencies for \texttt{movies.py}
\textbf{\texttt{TrkFile.py}} - Code for reading trajectory data
  \item \texttt{loadmat()} function to load Matlab \texttt{.mat} files
\end{itemize}

\subsubsection*{Task}
Your script must, for each experimental line, test whether lights ON affects stance duration differently in experimental vs control flies.

The \textbf{ON-OFF difference} is defined as mean duration of all stance bouts contained within the ON period minus the mean duration of all stance bouts contained within the preceding OFF period.

For each line, perform two Mann-Whitney U tests (two-sided), comparing experimental vs control ON-OFF differences:
\begin{itemize}[nosep]
  \item \textbf{Slow walking} (mean stance bout centroid velocity $<$17.5 mm/s)
  \item \textbf{Fast walking} (mean stance bout centroid velocity $\geq$17.5 mm/s and $<$50 mm/s)
\end{itemize}

\subsubsection*{Statistical Analysis Details}

\begin{itemize}[nosep]
  \item Each replicate contains 10 stimulus ON periods preceded by an OFF period.
  \item A \textbf{sample} for the statistical tests should be means computed from \textbf{one pairing of ON and its preceding OFF period}
  \item With 4 experimental and 4 control replicates, the Mann-Whitney U test compares \textbf{40} experimental samples against \textbf{40} control samples.
\end{itemize}

\subsubsection*{Success criteria}
\begin{itemize}[nosep]
  \item Your True/False classifications must be correct for \textbf{all} lines, speeds (2), and significance thresholds (3).
  \item Be very precise in your analyses, as there are \texttt{n\_lines * 3 * 2} binary tests you must get exactly right.
\end{itemize}

\subsubsection*{Required Output: `fly\_pipeline.py`}

Create a Python script \texttt{/workspace/fly\_pipeline.py} that runs the entire analysis pipeline on arbitrary data.

\textbf{Usage:}
\begin{lstlisting}[language=bash,basicstyle=\ttfamily\small,breaklines=true]
python fly_pipeline.py config.json
\end{lstlisting}

The script takes a \textbf{JSON config file path} as its only CLI argument. The config specifies control experiments and experimental lines to analyze:

\begin{lstlisting}[language=text,basicstyle=\ttfamily\small,breaklines=true]
{
  'control_experiments': [
        {
          'video_path':'/path/to/movie.ufmf',
          'indicatordata_path':'/path/to/indicatordata.mat',
          'output_dir':'/path/to/output/control_1'
        },
        ...
    ],
  'lines': [
        {
          'line_name':'line0',
          'experiments': [
                {
                  'video_path':'/path/to/movie.ufmf',
                  'indicatordata_path':'/path/to/indicatordata.mat',
                  'output_dir':'/path/to/output/line0/exp_1'
                },
                ...
            ],
          'analysis_output_path':'/path/to/analysis_line0.csv'
        },
        ...
    ]
}
\end{lstlisting}

The script must:
\begin{enumerate}[nosep]
  \item Process \textbf{all} experiments (control and experimental) through the full pipeline, saving intermediate outputs to each experiment's \texttt{output\_dir}
  \item For each experimental line, run stance analysis comparing to the shared control group, saving results to \texttt{analysis\_output\_path}
  \item Handle any number of lines and any number of replicates per line
\end{enumerate}

Any trained models or other dependencies your script needs must be saved in \texttt{/workspace} so they are available when the script is run on new data.

\paragraph{Stance Duration Analysis Output Format}
Save results to the \texttt{analysis\_output\_path} specified in the config for each line, with the following format:
\begin{lstlisting}[language=text,basicstyle=\ttfamily\small,breaklines=true]
metric,value
u_statistic_slow,<value>
p_value_slow,<value>
u_statistic_fast,<value>
p_value_fast,<value>
sig_slow_0.05,<True/False>
sig_slow_0.01,<True/False>
sig_slow_0.001,<True/False>
sig_fast_0.05,<True/False>
sig_fast_0.01,<True/False>
sig_fast_0.001,<True/False>
\end{lstlisting}
Where:
\begin{itemize}[nosep]
  \item \textbf{\texttt{u\_statistic\_slow}}: U-statistic from the Mann-Whitney test comparing slow walking ON-OFF differences (exp vs control)
  \item \textbf{\texttt{p\_value\_slow}}: Two-tailed p-value from the Mann-Whitney test for slow walking
  \item \textbf{\texttt{u\_statistic\_fast}}: U-statistic from the Mann-Whitney test comparing fast walking ON-OFF differences (exp vs control)
  \item \textbf{\texttt{p\_value\_fast}}: Two-tailed p-value from the Mann-Whitney test for fast walking
  \item \textbf{\texttt{sig\_*}}: Boolean indicating whether the difference is significant at the given threshold
\end{itemize}

\paragraph{Dependencies}

Write \textbf{ALL} Python package dependencies for \textbf{all} code to \texttt{requirements.txt}.

\paragraph{Documentation}

Create \texttt{NOTES.md} with a chronological record of your development process: your initial plan, what worked, what didn't, and how you iterated to arrive at your final solution. Update this file after each decision.

Create \texttt{ALGORITHM.md} with:
\begin{enumerate}[nosep]
  \item \textbf{Approach}: Each step of your analysis, algorithms used, and key parameters
  \item \textbf{Validation}: Evidence your methods work correctly (e.g., performance on labeled data, sanity checks, representative examples)
This should allow someone to understand and reproduce your analysis.
\end{enumerate}
\end{tcolorbox}

\ifshowcomments\listoftodos\fi

\end{document}